%% file: main.tex
\documentclass[a4paper, 10pt, conference]{ieeeconf}      
\usepackage{FG2026}

\FGfinalcopy 

\IEEEoverridecommandlockouts                              
\usepackage{graphics} 
\usepackage{epsfig} 
\usepackage{amsmath} 
\usepackage{amssymb}  

\usepackage{array} 
\usepackage{caption} 
\usepackage{booktabs} 
\usepackage{tabularx} 
\usepackage{subcaption} 

\makeatletter
\let\NAT@parse\undefined
\makeatother

\definecolor{fgblue}{rgb}{0.21,0.49,0.74}
\usepackage[pagebackref,breaklinks,colorlinks,allcolors=fgblue]{hyperref} 

\usepackage[capitalize]{cleveref}
\crefname{section}{Sec.}{Secs.}
\Crefname{section}{Section}{Sections}
\Crefname{table}{Table}{Tables}
\crefname{table}{Tab.}{Tabs.}

\def\FGPaperID{40} 

\makeatletter
\def\ps@titlepagestyle{%
  \def\@oddhead{\hfil\small 2026 International Conference on Automatic Face and Gesture Recognition (FG)\hfil}%
  \def\@evenhead{\hfil\small 2026 International Conference on Automatic Face and Gesture Recognition (FG)\hfil}%
  \def\@oddfoot{\footnotesize 979-8-3315-7231-0/26/\$31.00~\copyright 2026 IEEE\hfill}%
  \def\@evenfoot{\footnotesize 979-8-3315-7231-0/26/\$31.00~\copyright 2026 IEEE\hfill}%
}
\makeatother

\title{\LARGE \bf
NullFace: Training-Free Localized Face Anonymization
}

\author{\parbox{16cm}{\centering
  {\large Han-Wei Kung$^1$, Tuomas Varanka$^2$, Terence Sim$^3$, and Nicu Sebe$^1$}\\
  {\normalsize
  $^1$ University of Trento, Italy\\
  $^2$ University of Oulu, Finland\\
  $^3$ National University of Singapore, Singapore\\
  {\tt\small hanwei.kung@unitn.it}}}
}

\begin{document}

\ifFGfinal
\thispagestyle{empty}
\pagestyle{empty}
\else
\author{Anonymous FG2026 submission\\ Paper ID \FGPaperID \\}
\pagestyle{plain}
\fi

\newcommand{\quotes}[1]{``#1''}

\twocolumn[{
  \renewcommand\twocolumn[1][]{#1}%
  \maketitle

  \begin{center}
    \centering
    \captionsetup{type=figure}

    \resizebox{1.\textwidth}{!}{
      \begin{tabular}{
          *{1}{>{\centering\arraybackslash}m{\dimexpr.16\linewidth-2\tabcolsep-.16\arrayrulewidth}}
        }
        \multicolumn{1}{c}{Original} \\
        \midrule
        \multicolumn{1}{c}{\includegraphics[width={\dimexpr.16\linewidth}]{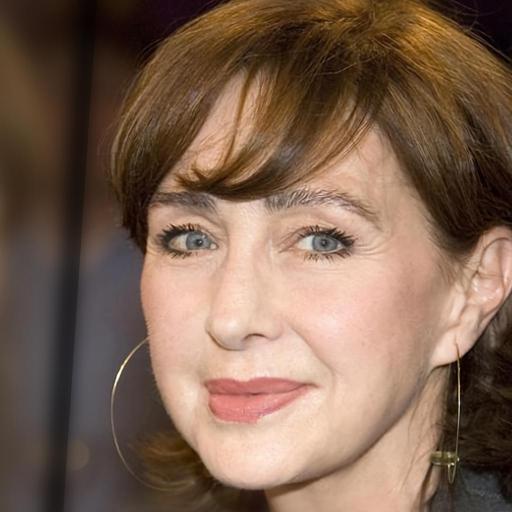}} \\
        \multicolumn{1}{c}{\includegraphics[width={\dimexpr.16\linewidth}]{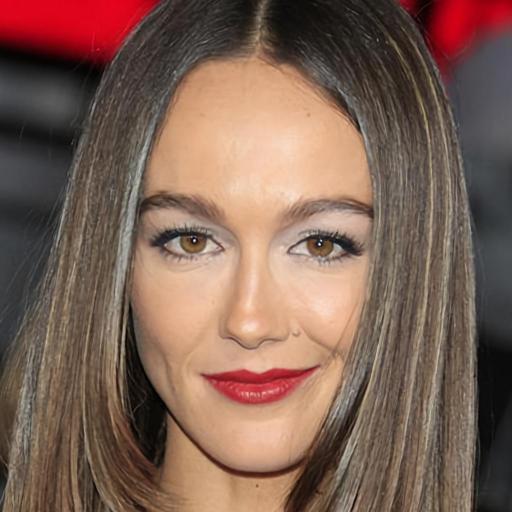}} \\
        \\
      \end{tabular}
      \quad
      \begin{tabular}{
          *{2}{>{\centering\arraybackslash}m{\dimexpr.16\linewidth-2\tabcolsep-.16\arrayrulewidth}}
        }
        \multicolumn{2}{c}{Full facial anonymization} \\
        \midrule
        \multicolumn{1}{c}{\includegraphics[width={\dimexpr.16\linewidth}]{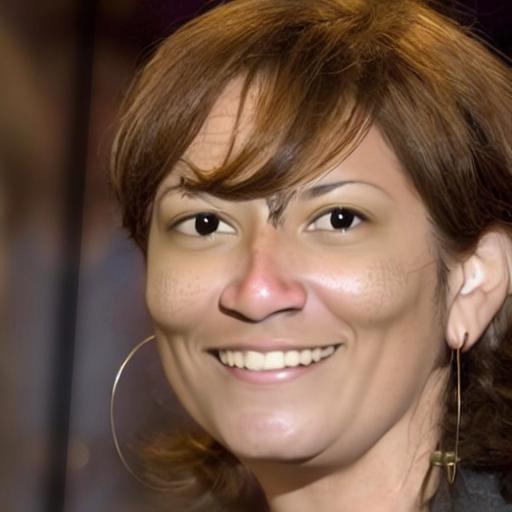}} & \multicolumn{1}{c}{\includegraphics[width={\dimexpr.16\linewidth}]{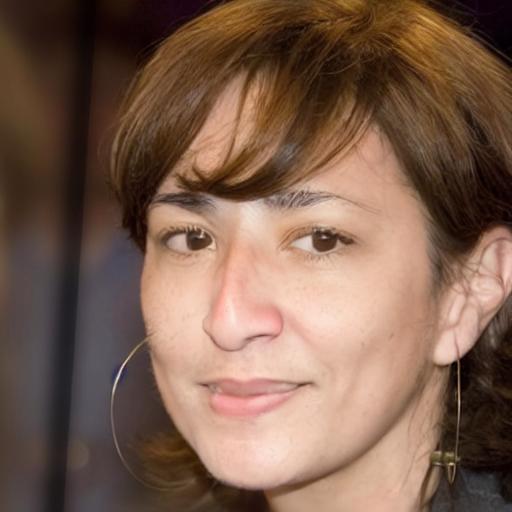}} \\
        \multicolumn{1}{c}{\includegraphics[width={\dimexpr.16\linewidth}]{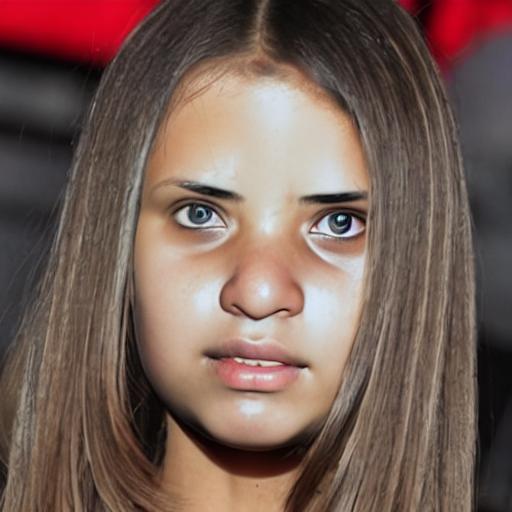}} & \multicolumn{1}{c}{\includegraphics[width={\dimexpr.16\linewidth}]{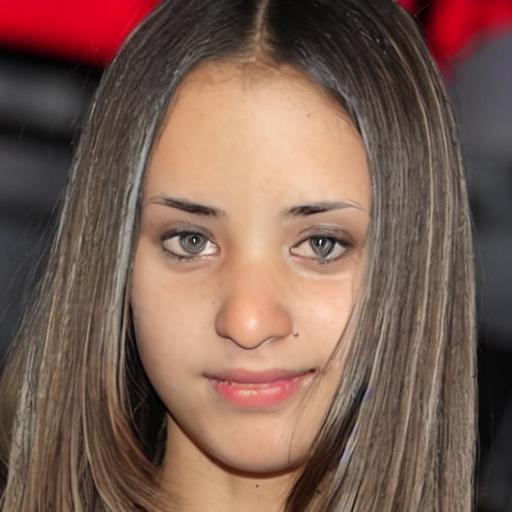}} \\
        \multicolumn{1}{c}{Inpainting~\cite{rombach2022high}} & \multicolumn{1}{c}{Ours} \\
      \end{tabular}
      \quad
      \begin{tabular}{
          *{3}{>{\centering\arraybackslash}m{\dimexpr.16\linewidth-2\tabcolsep-.16\arrayrulewidth}}
        }
        \multicolumn{3}{c}{Localized facial anonymization} \\
        \midrule
        \multicolumn{1}{c}{\includegraphics[width={\dimexpr.16\linewidth}]{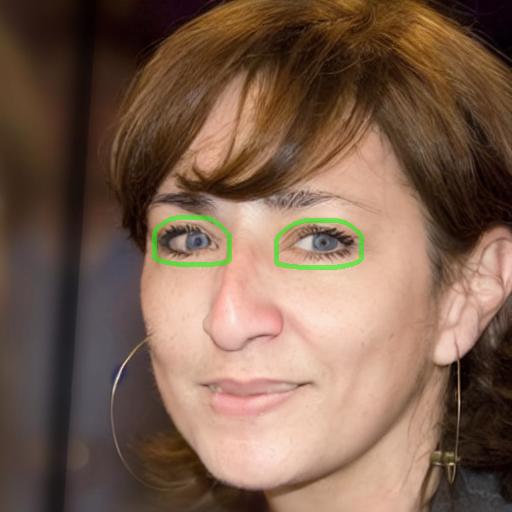}} & \multicolumn{1}{c}{\includegraphics[width={\dimexpr.16\linewidth}]{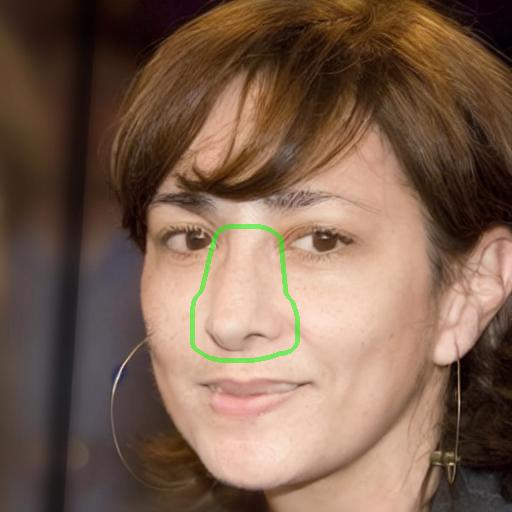}} & \multicolumn{1}{c}{\includegraphics[width={\dimexpr.16\linewidth}]{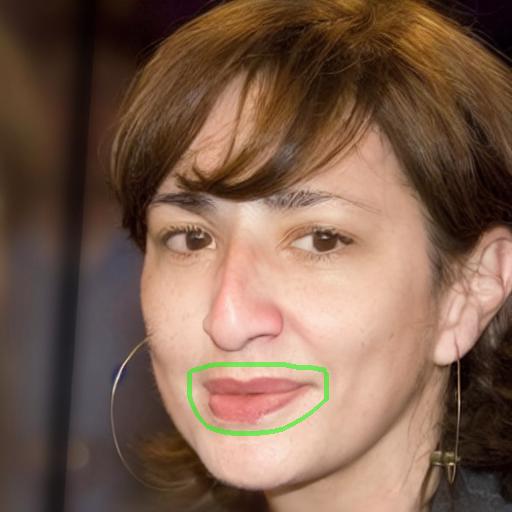}} \\
        \multicolumn{1}{c}{\includegraphics[width={\dimexpr.16\linewidth}]{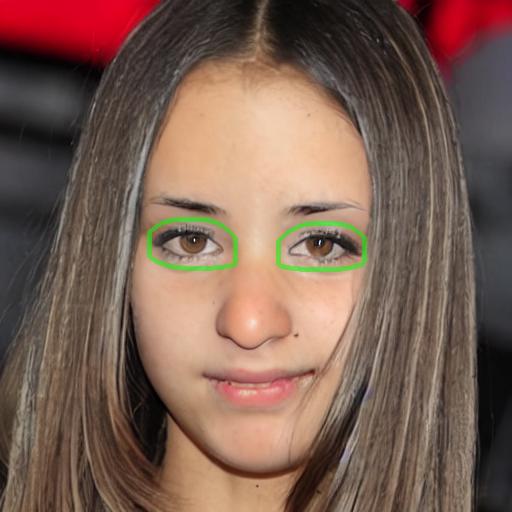}} & \multicolumn{1}{c}{\includegraphics[width={\dimexpr.16\linewidth}]{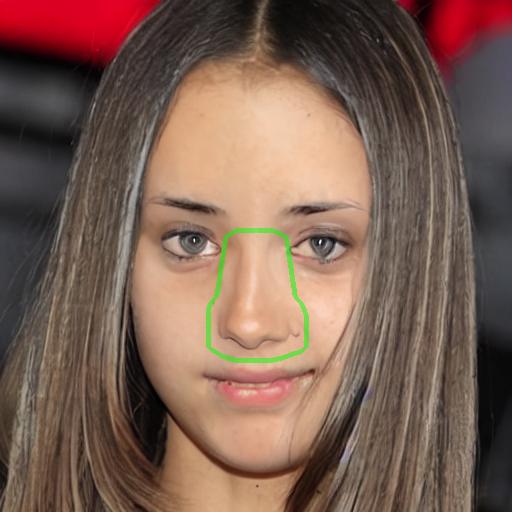}} & \multicolumn{1}{c}{\includegraphics[width={\dimexpr.16\linewidth}]{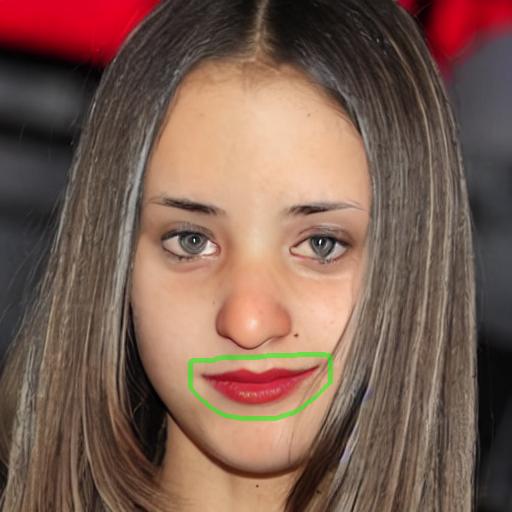}} \\
        \multicolumn{1}{c}{Keep eyes} & \multicolumn{1}{c}{Keep nose} & \multicolumn{1}{c}{Keep mouth} \\
      \end{tabular}
    }

    \captionof{figure}{Our method obscures identity while preserving attributes such as gaze, expressions, and head pose (in contrast to Stable Diffusion Inpainting~\cite{rombach2022high}) and enables selective anonymization of specific facial regions.}
  \end{center}
}]


\begin{abstract}

  Privacy concerns around ever increasing number of cameras are increasing in today's digital age. Although existing anonymization methods are able to obscure identity information, they often struggle to preserve the utility of the images. In this work, we introduce a training-free method for face anonymization that preserves key non-identity-related attributes. Our approach utilizes a pre-trained text-to-image diffusion model without requiring optimization or training. It begins by inverting the input image to recover its initial noise. The noise is then denoised through an identity-conditioned diffusion process, where modified identity embeddings ensure the anonymized face is distinct from the original identity. Our approach also supports localized anonymization, giving users control over which facial regions are anonymized or kept intact. Comprehensive evaluations against state-of-the-art methods show our approach excels in anonymization, attribute preservation, and image quality. Its flexibility, robustness, and practicality make it well-suited for real-world applications. Code and data can be found at \url{https://github.com/hanweikung/nullface}.

\end{abstract}


\section{INTRODUCTION}

As privacy regulations such as the General Data Protection Regulation (GDPR)~\cite{gdpr} in the European Union, the California Consumer Privacy Act (CCPA)~\cite{ccpa} in California, and the amended Protection of Personal Information (APPI)~\cite{appi} in Japan continue to evolve, the imperative for robust methods to protect personal data intensifies. Facial images, as biometric identifiers, represent a highly sensitive form of personal data. Organizations must adopt effective anonymization techniques to navigate these regulatory landscapes, safeguard privacy, and mitigate risks associated with cyberattacks~\cite{jain2006biometrics}.

Many methods have been developed to anonymize facial data~\cite{meden2021privacy,cao2024face}. Traditional face anonymization methods like masking, pixelation, and blurring are popular due to their simplicity, but they reduce image quality~\cite{hasan2018viewer}. Recent learning-based approaches using Generative Adversarial Networks (GANs)~\cite{goodfellow2014generative} can produce more realistic anonymous faces. However, GAN-based methods struggle to preserve non-identity-relevant attributes because they often depend on imprecise conditioning data like facial landmarks~\cite{sun2018natural,hukkelaas2019deepprivacy,maximov2020ciagan}. They also need tuning of multiple competing objectives~\cite{li2023riddle} that are difficult to balance. This limitation compromises the utility of anonymized data in applications like emotion analysis~\cite{kumar2017facial}, behavioral studies~\cite{dawson2005understanding}, and medical measurements using rPPG~\cite{tsou2020siamese}.

More recently, diffusion models~\cite{ho2020denoising,song2020denoising,rombach2022high} have gained attention for face anonymization due to their ease of training, realistic outputs, and fine-grained control. However, prior research~\cite{kung2025face} highlights a risk of overfitting, especially when training data is imbalanced across ethnicity or age groups, limiting their adaptability to diverse data scenarios.

Notably, some GAN-based~\cite{sun2018hybrid,sun2018natural,maximov2020ciagan,hukkelaas2019deepprivacy} and diffusion-based~\cite{klemp2023ldfa} methods use inpainting, where facial regions are detected, masked, and replaced with anonymous faces to maintain overall image integrity. However, this technique overwrites both identifying and non-identifying facial features, making the anonymized images less useful.

To tackle issues such as image quality degradation and difficulties in preserving non-identity-related attributes, we introduce a training-free method leveraging latent diffusion inversion with an identity-conditioned model. Our method avoids training challenges and loss balancing issues by utilizing a pre-trained diffusion model. It leverages the Denoising Diffusion Probabilistic Model (DDPM)~\cite{ho2020denoising} scheduler to invert the latent representation of an input image, retrieving its initial noise~\cite{huberman2024edit}. This inversion enables control over the anonymization process while maintaining fidelity to non-identifying facial attributes. Since the text-to-image Stable Diffusion model~\cite{rombach2022high} is solely conditioned on text prompts, we integrate an adapter~\cite{ye2023ip} to enable conditioning on identity embeddings. During denoising, we apply a controllable identity embedding to suppress the original identity, achieving effective anonymization.

Furthermore, localized anonymization, which selectively anonymizes specific facial regions, is valuable in medical settings~\cite{kaissis2020secure} where privacy is crucial, but specific clinical details must remain visible. For example, a dermatologist performing reconstructive surgery on a patient with facial burns may need to share post-surgery images in journals or conferences to highlight the treated area, such as the cheek or forehead. However, to protect the patient's privacy, the full face must remain unrecognizable. In such cases, localized anonymization can mask the rest of the face, retaining only the treated area. This approach enables doctors to share knowledge with the medical community without violating the patient's privacy. For localized control, we use segmentation maps to keep the targeted regions visible while anonymizing masked areas. Unlike Stable Diffusion Inpainting~\cite{rombach2022high}, which often changes non-identity attributes in masked areas, our method integrates segmentation maps with inversion to preserve them.

Key contributions of our method include:
\begin{itemize}
  \item \textit{Training-free anonymization.} By leveraging diffusion inversion and identity embeddings, our method anonymizes identity while preserving non-identifiable facial features, eliminating the need for training or fine-tuning.
  \item \textit{Localized anonymization.} Segmentation masks enable control over which facial regions are anonymized, providing flexibility to balance privacy and usability.
  \item \textit{Integration of inversion and segmentation maps.} Combining inversion with segmentation maps enhances the preservation of important facial attributes compared to conventional inpainting methods.
  \item \textit{State-of-the-art performance.} Our approach achieves competitive results on benchmark datasets, demonstrating its effectiveness in privacy-preserving applications without compromising data quality.
\end{itemize}


\section{RELATED WORK}

\paragraph{Face anonymization.}

Early face anonymization methods like pixelation, blurring, or mask overlays often degrade image quality. To overcome this limitation, recent deep learning advancements have employed GANs~\cite{goodfellow2014generative} to produce more realistic results~\cite{zhai2022a3gan,sun2018natural,sun2018hybrid,hukkelaas2019deepprivacy,hukkelaas2023deepprivacy2,gu2020password,dall2022graph,ciftci2023my,rosberg2023fiva,wen2023divide,helou2023vera}. Methods like~\cite{sun2018natural,sun2018hybrid,hukkelaas2019deepprivacy,hukkelaas2023deepprivacy2} use inpainting to fill masked regions with synthetic faces, while others~\cite{luo2022styleface,li2023riddle,barattin2023attribute} involve latent space manipulation with StyleGAN2~\cite{karras2020analyzing}. Techniques such as RiDDLE~\cite{li2023riddle} and FALCO~\cite{barattin2023attribute} use GAN inversion~\cite{tov2021designing} to generate high-quality anonymized faces. Some methods~\cite{li2023riddle,zhang2024facersa,he2024diff,yang2024g,gu2020password} also enable the restoration of anonymized faces through a key, similar to encryption and decryption.

More recent anonymization methods utilize diffusion models. LDFA~\cite{klemp2023ldfa} implements a two-stage approach, using a face detection model followed by a latent diffusion model to create anonymous faces through inpainting. Diff-Privacy~\cite{he2024diff} develops a custom image inversion module to generate conditional embeddings for anonymization. IDDiffuse~\cite{shaheryar2024iddiffuse} introduces a dual-conditional diffusion framework that maps synthetic identities, ensuring the anonymized version of an individual remains consistent across an entire video. FAMS~\cite{kung2025face} uses synthetic training data to achieve anonymization by modeling the process as face swapping. In contrast, our approach offers a training-free solution by utilizing diffusion model inversion to extract initial noise and conditioning anonymized identities during the denoising process.

\paragraph{Diffusion-based image editing.}

Text-guided diffusion models~\cite{ramesh2021zero,ramesh2022hierarchical,nichol2021glide,rombach2022high,saharia2022photorealistic} have excelled in image synthesis. Prompt-to-Prompt~\cite{hertz2022prompt} enables text-based editing of synthetic images, while SDEdit~\cite{meng2021sdedit} supports real image editing by adding noise to the input and denoising it using user-drawn strokes. However, maintaining fidelity between original and edited real images has been challenging. To address this, researchers developed inversion techniques to enhance editing accuracy by reconstructing a noisy input image representation~\cite{mokady2023null,huberman2024edit,tumanyan2023plug,deutch2024turboedit}. Advancements like Null-text Inversion~\cite{mokady2023null} have improved fidelity by optimizing the inversion process with a null prompt, ensuring edits match the text while preserving original features. LEDITS++~\cite{brack2024ledits++} uses a more efficient inversion technique with dpmsolver++ for faster convergence and supports simultaneous editing of multiple concepts. We adapt these inversion-based editing techniques to modify facial features while maintaining the original image's fidelity.

\paragraph{Personalized image generation using diffusion models.}

Recent advances in diffusion models have enabled identity-consistent image generation guided by minimal input, such as text or visual prompts. Textual Inversion~\cite{gal2022image} learns word embeddings from a small set of images (typically 3-5) to capture visual concepts. DreamBooth~\cite{ruiz2023dreambooth} fine-tunes diffusion models to reflect traits of individuals across diverse contexts. IP-Adapter~\cite{ye2023ip} employs a decoupled cross-attention mechanism alongside a trainable projection network to enable personalized prompts for pretrained diffusion models. InstantID~\cite{wang2024instantid} integrates an Image Adapter, which facilitates the injection of image-based prompts using cross-attention, and IdentityNet, which ensures identity preservation. PuLID~\cite{guo2024pulid} combines contrastive alignment loss and identity-specific loss to achieve high fidelity and editability, avoiding overfitting through regularization.

While these methods focus on conditioning diffusion models to generate personalized images, our approach shifts to identity anonymization. Instead of preserving or enhancing an identity, we use identity embeddings to remove defining features, repurposing identity conditioning to obscure the original identity. This leverages diffusion models' capabilities to create realistic images while ensuring the individual is unrecognizable.


\section{METHOD}

\begin{figure*}[t]
  \centering
  \includegraphics[width=0.75\linewidth]{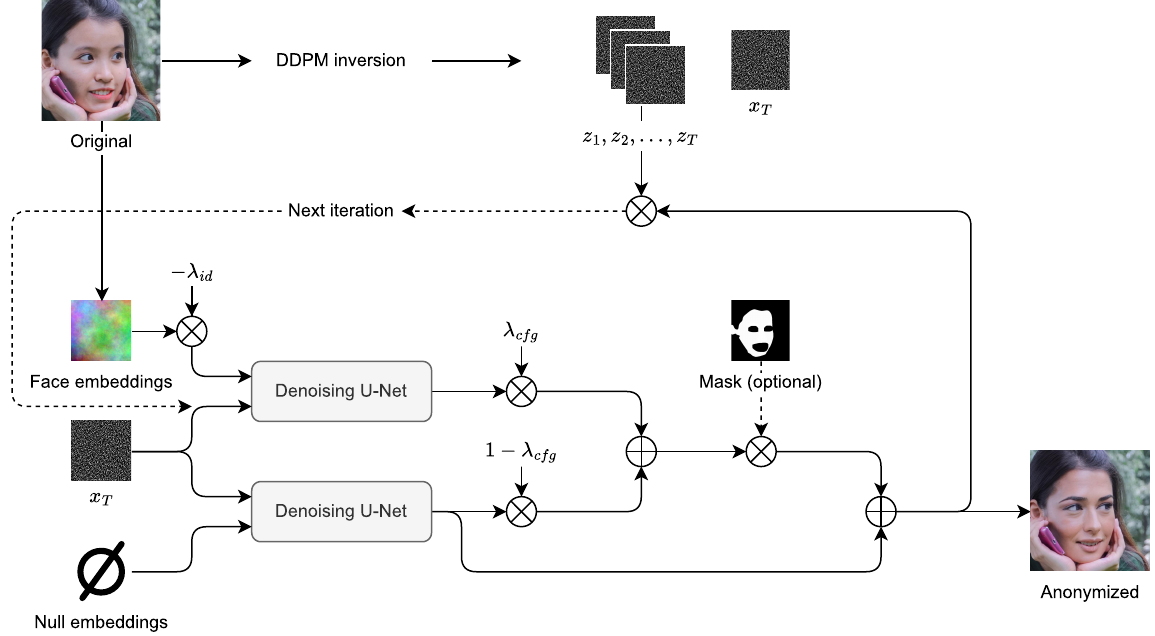}
  \caption{Face anonymization pipeline using diffusion model inversion. Given a facial image, we perform DDPM inversion~\cite{huberman2024edit} to retrieve the initial noise map $x_T$ and a sequence of noise maps $\{z_t\}$ from the diffusion process. Face embeddings are extracted using a face recognition model~\cite{deng2019arcface} and negated with a hyperparameter $\lambda_{id}$, creating negative identity guides. These guides steer the model away from reconstructing the original identity during denoising. The denoising process begins with $x_T$, combining conditional and unconditional paths. The conditional path utilizes negated identity embeddings to obscure identifiable features, while the unconditional path uses null embeddings ($\varnothing$) to preserve non-identifying attributes. Outputs from both paths are merged using a guidance scale parameter $\lambda_{cfg}$ through \cref{eq:cfg}. Lastly, masks can be applied at each iteration to control which facial features are anonymized or retained, enabling localized anonymization.}
  \label{fig:pipeline}
\end{figure*}

This section details our anonymization framework, which integrates diffusion model inversion, a dual-path denoising structure, and modified face embeddings. We also extend it for localized anonymization by integrating segmentation maps with diffusion inversion.

\subsection{Preliminary}

\paragraph{DDPM inversion.} Inversion involves determining the latent \(x_T\) that generates the latent image \(x_0\). While DDIM~\cite{song2020denoising} enables deterministic sampling and can be inverted to find \(x_T\), prior research~\cite{mokady2023null} revealed that this latent is unsuitable for editing with classifier-free guidance~\cite{ho2022classifier}. DDPM inversion~\cite{huberman2024edit} addresses this limitation by not only identifying the initial latent \(x_T\) but also computing the set of \(z_t\), representing the noise applied when transitioning from one noisy image \(x_t\) to the next \(x_{t-1}\):  
\begin{equation}
  z_t = \frac{x_{t-1} - \mu_t(x_t, c)}{\sigma_t},\quad t=T,\ldots,1.
  \label{eq:inversion}
\end{equation}
Here, \(c\) is a conditioning prompt, \(\mu_t(x_t, c)\) is the output of the denoising network, and \(\sigma_t\) is the standard deviation from the noise scheduler. Using the set \(z_t\) during editing ensures minimal changes to the image structure.

\paragraph{IP-Adapter.} IP-Adapter~\cite{ye2023ip} enhances text-to-image diffusion models with image prompt capabilities by modifying the attention mechanism. It introduces decoupled cross-attention that processes text and image features independently, preserving the integrity of the pre-trained base model. The outputs of the decoupled cross-attention are combined with the original cross-attention using a scaling factor:
\begin{equation}
  \begin{aligned}
    \mathbf{Z} & = \text{Attention}(Q_{noise}, K_{text}, V_{text}) \\
               & + \lambda_{img} \cdot \text{Attention}(Q_{noise}, K_{img}, V_{img}),
    \label{eq:ip_adapter}
  \end{aligned}
\end{equation}
where \( \lambda_{img} \) is the scaling factor, and \(\{Q, K, V\}_{source}\) represent the projection matrices---queries (\(Q\)), keys (\(K\)), and values (\(V\))---used to project various types of \textit{source} features. This approach is resource-efficient, requiring minimal parameters, yet it enables significant enhancements. Additionally, the method has been extended to include other features, such as identity embeddings, allowing text-to-image models to condition outputs on an individual's identity.

\subsection{Anonymization with Inversion}
\label{sec:method}
\paragraph{Inversion-based face editing.} Previous face anonymization methods~\cite{sun2018hybrid,sun2018natural,maximov2020ciagan,hukkelaas2019deepprivacy,kuang2024facial,klemp2023ldfa} often treat the task as an image inpainting problem, where the facial region is first erased and then replaced with an alternative identity. However, erasing the original face also removes essential non-identity features, such as gaze direction, head pose, and facial expressions, compromising attribute preservation. While some methods~\cite{sun2018hybrid,sun2018natural,maximov2020ciagan,hukkelaas2019deepprivacy,kuang2024facial} use annotations like facial landmarks to retain these attributes, such annotations can be unreliable. To overcome these limitations, we adopt a noise-level editing approach instead of inpainting, eliminating reliance on facial landmarks. The initial noise of the image is accessed through inversion techniques. In our experiments, we compared popular methods like DDIM~\cite{mokady2023null} and DDPM~\cite{huberman2024edit} inversion and found that DDPM inversion achieves higher fidelity in reconstructing the original image.

Our pipeline, illustrated in \cref{fig:pipeline}, begins with an input facial image. The image undergoes the DDPM inversion~\cite{huberman2024edit} process to derive its initial noise condition and a sequence of noise maps $\{z_t\}$, as described in \cref{eq:inversion}. In the denoising phase, we condition the generation on a controlled identity to achieve anonymization. Since the Stable Diffusion model~\cite{rombach2022high} uses text inputs rather than identity embeddings, we integrate IP-Adapter~\cite{ye2023ip} to enable conditioning on identity embeddings. However, unlike previous methods~\cite{guo2024pulid,ma2024subject,cui2024idadapter} that generate toward a specified identity, we operate in reverse to anonymize the input identity, as detailed in the \textit{Controllable Training-free Identity Variations} paragraph of this section. 

Moreover, to better preserve original facial attributes, we apply the identity embedding condition only during a portion of the denoising phase. Specifically, we introduce a parameter, \( T_{\text{skip}} \), to determine how many of the total \( T \) denoising steps to skip before applying the identity condition. Skipping more steps generates an image that aligns more closely with the original.

\paragraph{Conditional and unconditional denoising paths.} To modify the identity, the denoising process follows two parallel paths: a conditional path guided by identity embeddings \( c_{id} \) and an unconditional path that preserves non-identity-related facial attributes. The outputs of these paths are combined using a guidance scale parameter \( \lambda_{cfg} \), following the classifier-free guidance technique~\cite{ho2022classifier}, as defined by the following equation:
\begin{equation}
  \begin{aligned}
    \hat{\epsilon}_\theta(x_t, c_{id}, \varnothing) & = \lambda_{cfg} \cdot \epsilon_\theta(x_t, c_{id}) \\
                                                    & + (1-\lambda_{cfg}) \cdot \epsilon_\theta(x_t, \varnothing).
  \end{aligned}
  \label{eq:cfg}
\end{equation}
Here, \( \lambda_{cfg} \) represents the guidance scale, which determines the relative contributions of the two paths. $\epsilon_\theta$ denotes the network, $x_t$ is the noisy version of the input $x$ at timestep $t$, where $t=1,\ldots,T$, $c_{id}$ refers to the identity embeddings, and $\varnothing$ refers to the null identity embeddings.

This dual-path framework balances anonymization with the preservation of original non-identity-related features. The conditional path focuses on incorporating the alternative identity representations, while the unconditional path ensures the retention of attributes like gaze direction and facial expressions. Together, these paths enable the generation of anonymized faces that conceal the original identity while preserving important non-identity-related attributes.

\paragraph{Controllable training-free identity variations.} The conditional path modifies the original identity to an alternative one, but what should this alternative identity be for effective anonymization? Previous methods~\cite{hukkelaas2019deepprivacy,maximov2020ciagan} train generative models to create synthetic identities, which are constrained by the scope of the training data. To avoid such limitations, we adopt a training-free approach by using a pre-trained face recognition model~\cite{deng2019arcface} to extract face embeddings from the input image. This leverages the face recognition model's discriminative ability to capture diverse identity variations without additional training.

We also aim for a mathematically interpretable anonymization process, where the degree of deviation from the original identity can be controlled. To this end, we introduce an anonymization parameter \( \lambda_{id} \), a positive scalar that scales the extracted face embeddings. The scaled embeddings are then negated~\cite{rosberg2023fiva} and used as the alternative identity. By increasing \( \lambda_{id} \), the alternative identity becomes increasingly distinct from the original. During the denoising process, the model is conditioned on these negated embeddings, ensuring the generated identity diverges from the original, effectively anonymizing it.

In summary, we formulate the anonymization task as the following optimization problem. Given an input facial image \( x \), and a generation network \( \epsilon \) that produces an anonymized image conditioned on parameters \( T_{\text{skip}} \), \( \lambda_{\text{id}} \), and \( \lambda_{\text{cfg}} \), the goal is to maximize identity distance while minimizing non-identity-related attribute distortion:

{\normalsize
\[
	\begin{aligned}
		\max_{T_{\text{skip}},\, \lambda_{\text{id}},\, \lambda_{\text{cfg}}} \quad & D_{\text{id}}\left( \epsilon(x, T_{\text{skip}}, \lambda_{\text{id}}, \lambda_{\text{cfg}}), x \right), \\
		\min_{T_{\text{skip}},\, \lambda_{\text{id}},\, \lambda_{\text{cfg}}} \quad & D_{\text{attr}}\left( \epsilon(x, T_{\text{skip}}, \lambda_{\text{id}}, \lambda_{\text{cfg}}), x \right),
	\end{aligned}
\]
}

\noindent where \( D_{\text{id}}(\cdot, \cdot) \) measures identity distance (to be maximized), \( D_{\text{attr}}(\cdot, \cdot) \) measures the distance in non-identity-related attributes---such as expression, gaze, and pose (to be minimized), and \( \epsilon(x, T_{\text{skip}}, \lambda_{\text{id}}, \lambda_{\text{cfg}}) \) denotes the generated anonymized image.

\subsection{Localized Anonymization}

Localized facial anonymization can balance patient privacy with clinical utility, enabling medical practitioners to showcase detailed views of specific areas while protecting patient identity. To achieve this, we use segmentation masks to selectively anonymize facial regions. However, unlike standard inpainting~\cite{rombach2022high}, which \textit{paints} over masked regions and loses non-identity-related features, our approach combines segmentation masks with outputs from the dual-path framework that incorporates noise information obtained through inversion (see \cref{sec:method}). Precisely, during the denoising process, the segmentation mask \( M \) is applied by:
\begin{equation}
  \begin{aligned}
    \tilde{\epsilon}_\theta(x_t, c_{id}, \varnothing) & = M \cdot \hat{\epsilon}_\theta(x_t, c_{id}, \varnothing) \\
                                                      & + (1 - M) \cdot \epsilon_\theta(x_t, \varnothing).
  \end{aligned}
\end{equation}
Here, \( M \) is applied to the combination of both paths to ensure anonymization in masked regions while preserving non-identifying features. Simultaneously, the complement of the mask, \( 1 - M \), is applied to the unconditional path to keep details in unmasked areas visible. This integration allows customizable control, ensuring the relevant regions remain clear while anonymizing other parts of the face.


\section{EXPERIMENTS}

We analyze the impact of key components in our approach---$T_{\text{skip}}$, the anonymization parameter ($\lambda_{id}$), the guidance scale ($\lambda_{cfg}$), and segmentation masks---on image alignment, identity anonymization, and attribute preservation. Evaluations on two public datasets demonstrate our method's competitive performance against existing baselines. An ablation study further highlights the critical role of diffusion model inversion in achieving these results. Beyond full-face anonymization, we also demonstrate the usefulness of localized facial anonymization in applications like medical case sharing.

\subsection{Controlling Image Alignment}

\begin{figure}[h]
  \footnotesize
    \begin{tabular}{*{5}{>{\centering\arraybackslash}m{\dimexpr.2\linewidth-2\tabcolsep}}}
      \hline
      \(T_{\text{skip}}=0\) & \(T_{\text{skip}}=29\) & \(T_{\text{skip}}=59\) & \(T_{\text{skip}}=89\) & Original \\ [\defaultaddspace]
      \multicolumn{5}{@{}c@{}}{\includegraphics[width=\linewidth]{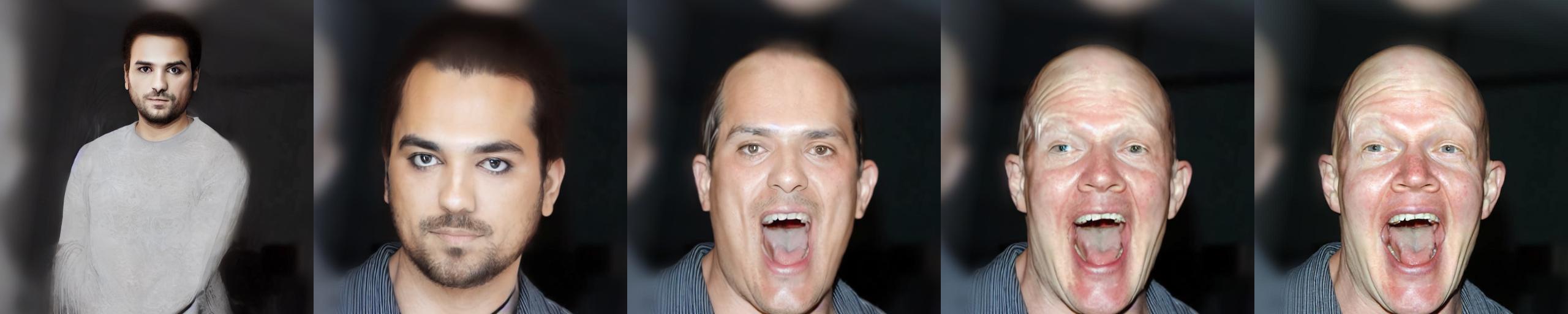}} \\
      \multicolumn{5}{@{}c@{}}{\includegraphics[width=\linewidth]{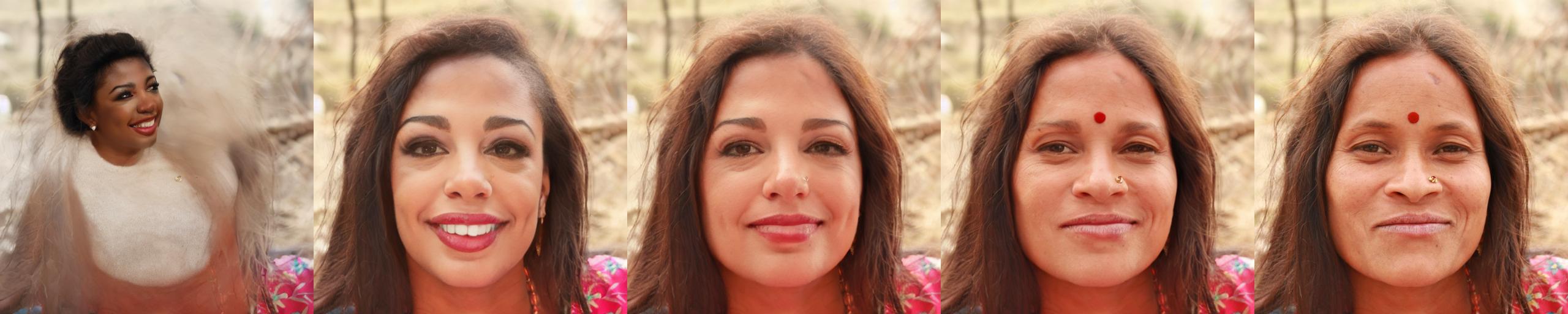}} \\
    \end{tabular}
    \caption{As $T_{\text{skip}}$ increases from 0 to higher values, the generated image progressively aligns more closely with the input, ultimately achieving near-perfect reconstruction.}
  \label{fig:skip}
\end{figure}

Our face anonymization method introduces a parameter, $T_{\text{skip}}$, to regulate the degree of alignment between the generated image and the input. $T_{\text{skip}}$ specifies the point in the denoising process where modified face embeddings are first injected. In a process with $T$ total steps, embeddings are introduced starting at step \( T - T_{\text{skip}} \).

Increasing $T_{\text{skip}}$ enhances alignment with the input image. Prior research~\cite{huberman2024edit} suggests this adjustment refines semantic features while preserving the overall structure. \Cref{fig:skip} illustrates this effect using a denoising process with \( T = 100 \) steps. At \( T_{\text{skip}} = 0 \), the generated image deviates significantly from the input, with notable differences in body posture. As $T_{\text{skip}}$ increases, the generated image progressively aligns with the input's expression and structure, achieving near-perfect reconstruction at higher values.

\subsection{Identity Change with Anonymization Parameter}

Our method anonymizes faces by scaling the face embeddings by a factor of $-\lambda_{id}$, where $\lambda_{id}$ is a positive value and serves as the anonymization parameter. Increasing $\lambda_{id}$ generates faces that are progressively less similar to the original identity, allowing controlled adjustments in anonymity.

\begin{figure}[h]
  \footnotesize
    \begin{tabular}{*{5}{>{\centering\arraybackslash}m{\dimexpr.2\linewidth-2\tabcolsep}}}
      \hline
      Original & \( \lambda_{id}=0.0 \) & \( \lambda_{id}=.33 \) & \( \lambda_{id}=.67 \) & \( \lambda_{id}=1.0 \) \\ [\defaultaddspace]
      \multicolumn{5}{@{}c@{}}{\includegraphics[width=\linewidth]{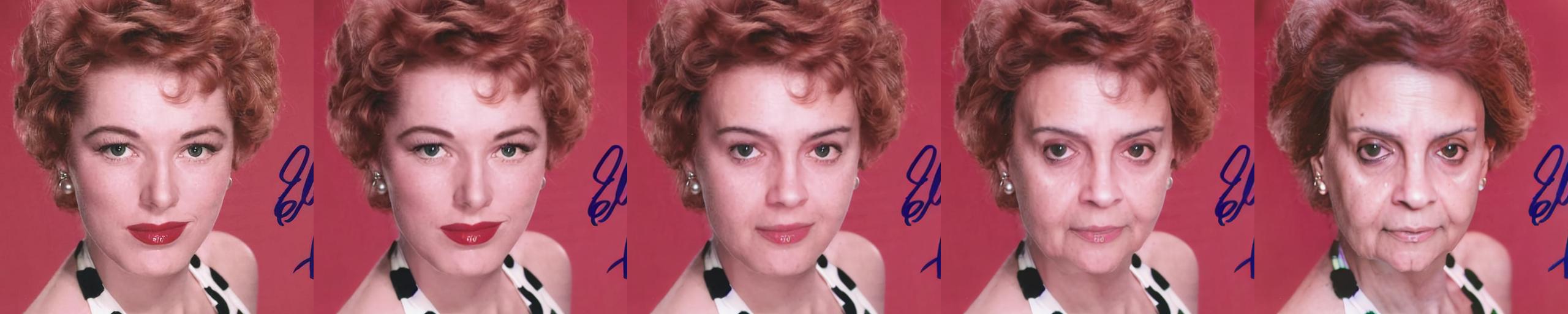}} \\
      \multicolumn{1}{@{}c@{}}{ID Dist.} & 0.012 & 0.654 & 0.961 & 1.147 \\ [\defaultaddspace]
      \multicolumn{5}{@{}c@{}}{\includegraphics[width=\linewidth]{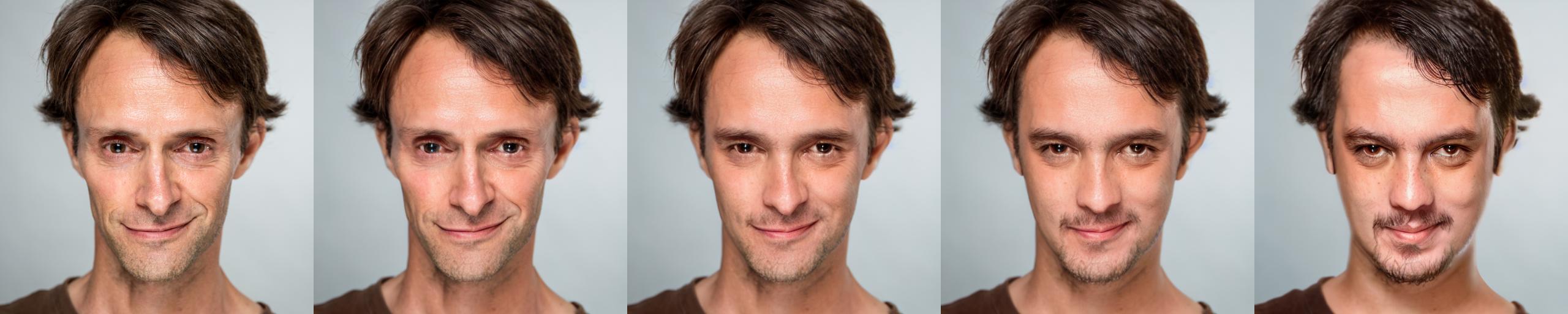}} \\
      \multicolumn{1}{@{}c@{}}{ID Dist.} & 0.013 & 0.488 & 0.905 & 1.339 \\ [\defaultaddspace]
      \multicolumn{1}{@{}c@{}}{} & \multicolumn{4}{@{}c@{}}{\includegraphics[width=.8\linewidth]{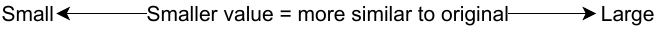}} \\
    \end{tabular}
    \caption{Increasing $\lambda_{id}$ generates faces that are less similar to the original, with identity distances shown for each example.}
  \label{fig:embd}
\end{figure}

\Cref{fig:embd} visualizes this process, displaying the original face alongside four anonymized versions generated with \( \lambda_{id} = 0 \), $0.33$, $0.67$, and $1.0$. We quantified the identity change by calculating the identity distance from the original face using the FaceNet~\cite{schroff2015facenet} model. As $\lambda_{id}$ increases, both visual inspection and identity distance confirm greater divergence from the original identity.

\subsection{Effect of Guidance Scale on Anonymization}

\begin{figure}[h]
  \footnotesize
  \begin{tabular}{*{5}{>{\centering\arraybackslash}m{\dimexpr.2\linewidth-2\tabcolsep}}}
    \hline
    Original & \( \lambda_{cfg}=5 \) & \( \lambda_{cfg}=10 \) & \( \lambda_{cfg}=15 \) & \( \lambda_{cfg}=20 \) \\ [\defaultaddspace]
    \multicolumn{5}{@{}c@{}}{\includegraphics[width=.96\linewidth]{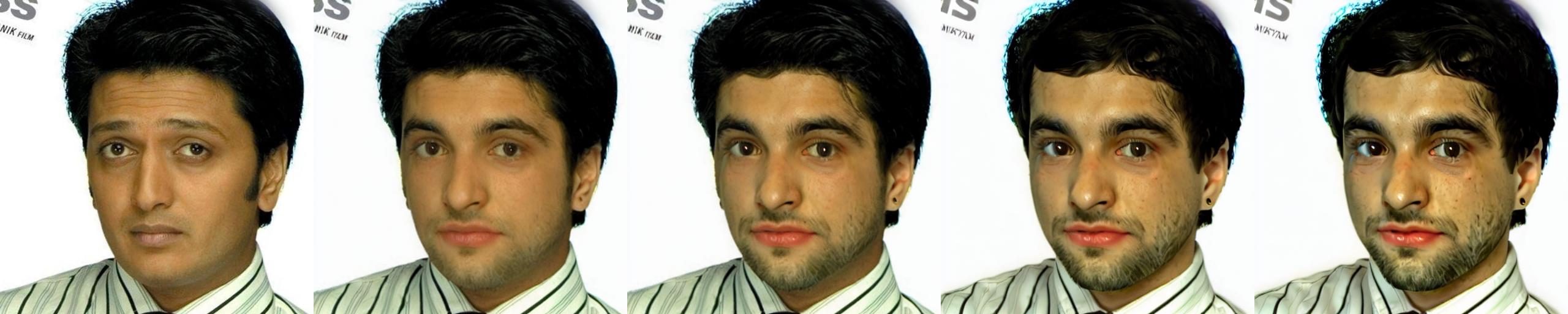}} \\
    \multicolumn{1}{@{}c@{}}{ID Dist.} & 0.743 & 0.936 & 0.952 & 0.975 \\ [\defaultaddspace]
    \multicolumn{5}{@{}c@{}}{\includegraphics[width=.96\linewidth]{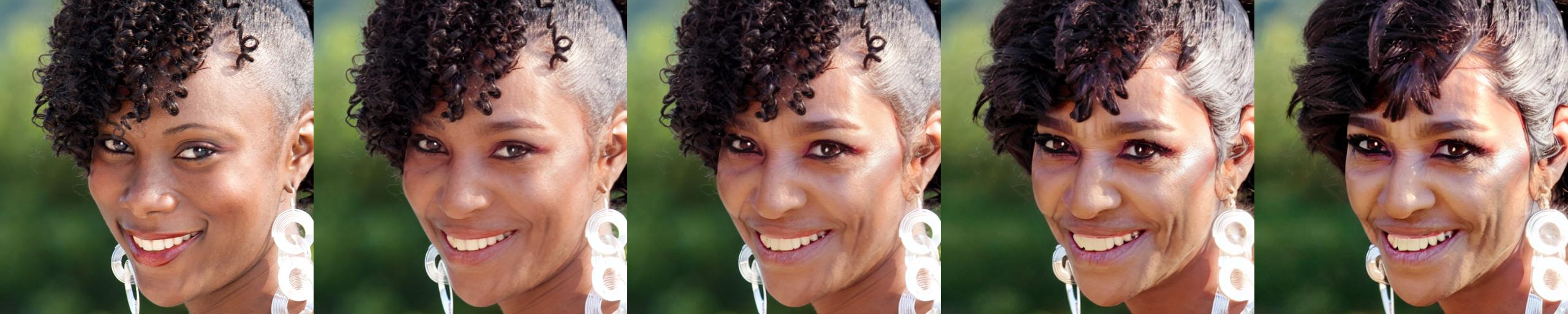}} \\
    \multicolumn{1}{@{}c@{}}{ID Dist.} & 0.457 & 0.667 & 0.671 & 0.725 \\ [\defaultaddspace]
    \multicolumn{1}{@{}c@{}}{} & \multicolumn{4}{@{}c@{}}{\includegraphics[width=.8\linewidth]{./images/embd/arrow.pdf}} \\
  \end{tabular}
  \caption{As the guidance scale increases, the anonymized identities become increasingly distinct from the originals, as confirmed by identity distance measurements using FaceNet~\cite{schroff2015facenet}. However, the version generated with a guidance scale of 20 reveals that excessively high guidance scales, while widening identity distance, compromise the photorealism of the resulting images. }
  \label{fig:cfg}
\end{figure}

The guidance scale $\lambda_{cfg}$ influences the degree of identity change during image generation. \Cref{fig:cfg} demonstrates this effect with two original images and four anonymized versions generated using \( \lambda_{cfg} = 5 \), $10$, $15$, and $20$. Higher guidance scale values yield anonymized identities that are more distinct from the originals. However, excessive guidance (e.g., \( \lambda_{cfg} = 20 \)) reduces photorealism.

\subsection{Parameter Impact on Anonymization}

\begin{figure}[h]
  \centering
  \begin{subfigure}{0.49\linewidth}
    \centering
    \includegraphics[width=1.0\linewidth]{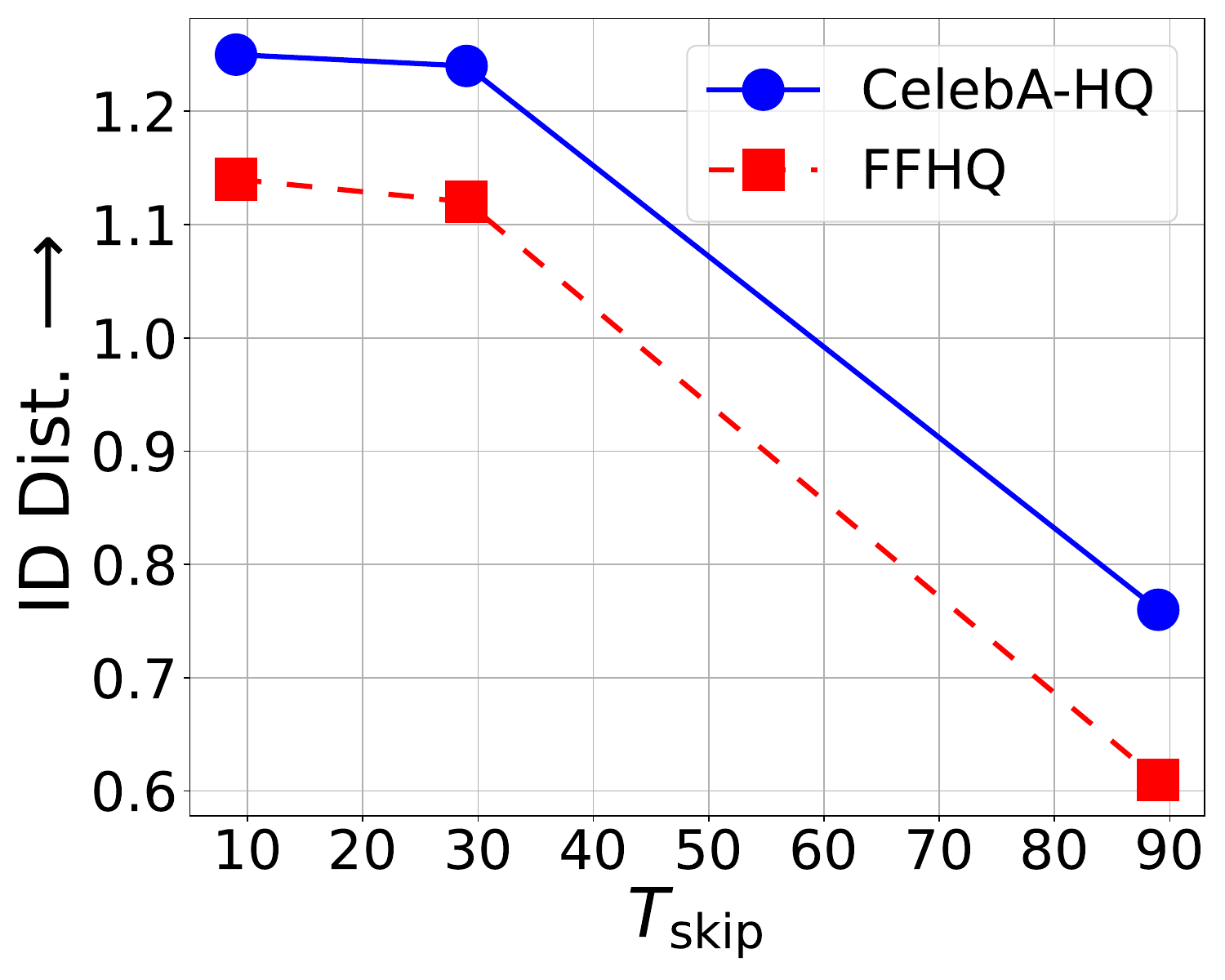}
    \caption{}
    \label{fig:identity_distance_vs_Tskip}
  \end{subfigure}
  \hfill
  \begin{subfigure}{0.49\linewidth}
    \centering
    \includegraphics[width=1.0\linewidth]{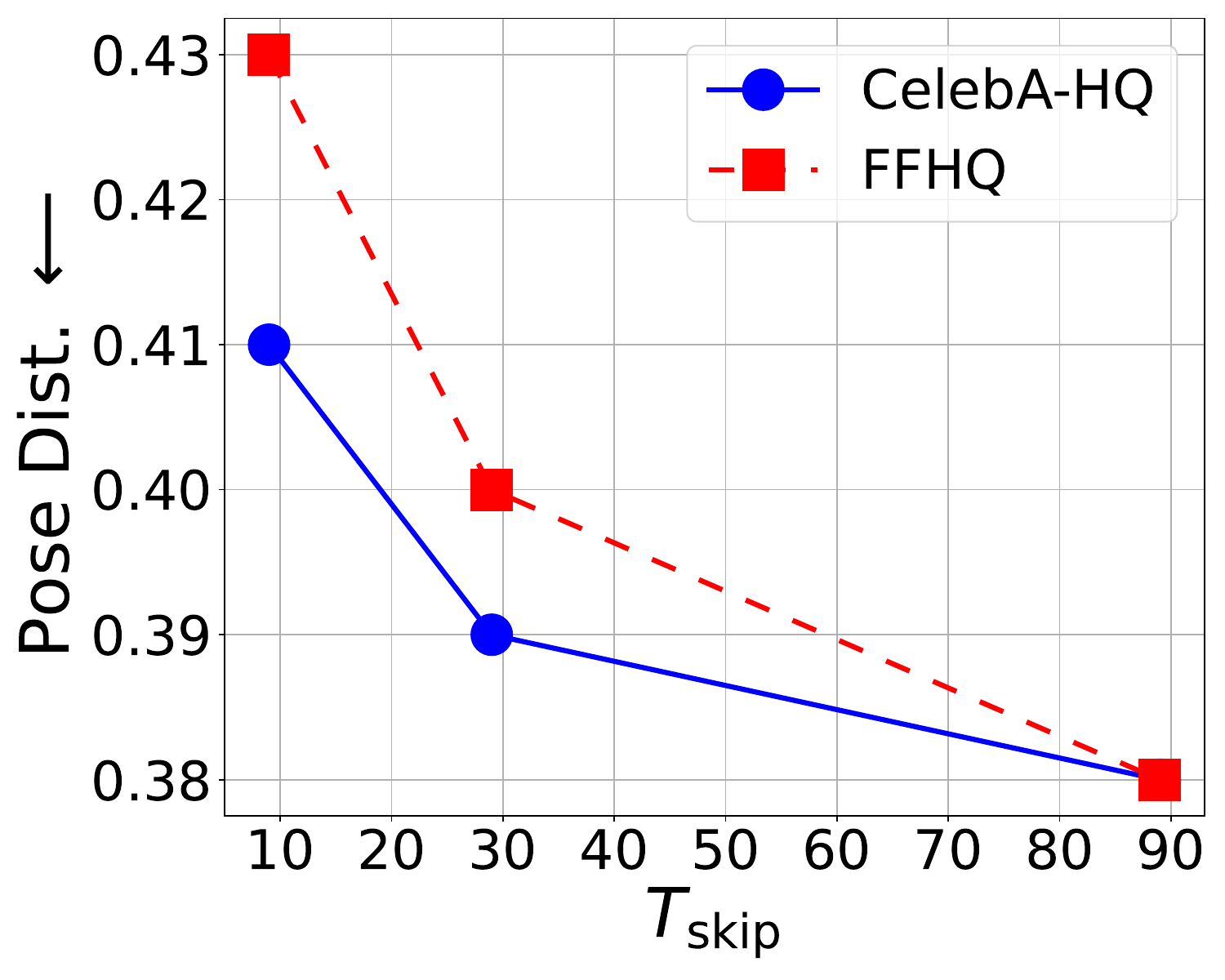}
    \caption{}
    \label{fig:pose_distance_vs_Tskip}
  \end{subfigure}
  \\
  \begin{subfigure}{0.49\linewidth}
    \centering
    \includegraphics[width=1.0\linewidth]{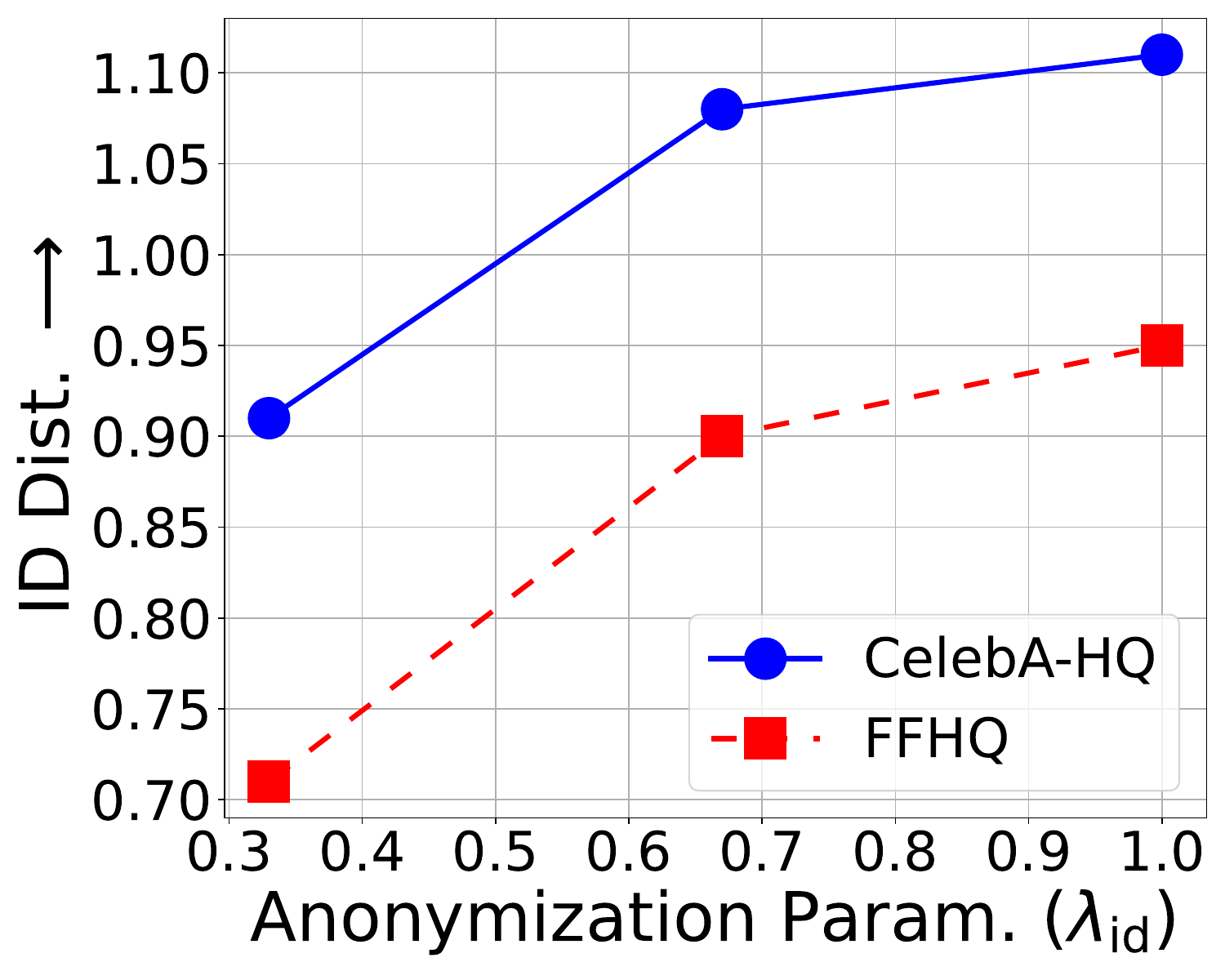}
    \caption{}
    \label{fig:identity_distance_vs_lambda_id}
  \end{subfigure}
  \hfill
  \begin{subfigure}{0.49\linewidth}
    \centering
    \includegraphics[width=1.0\linewidth]{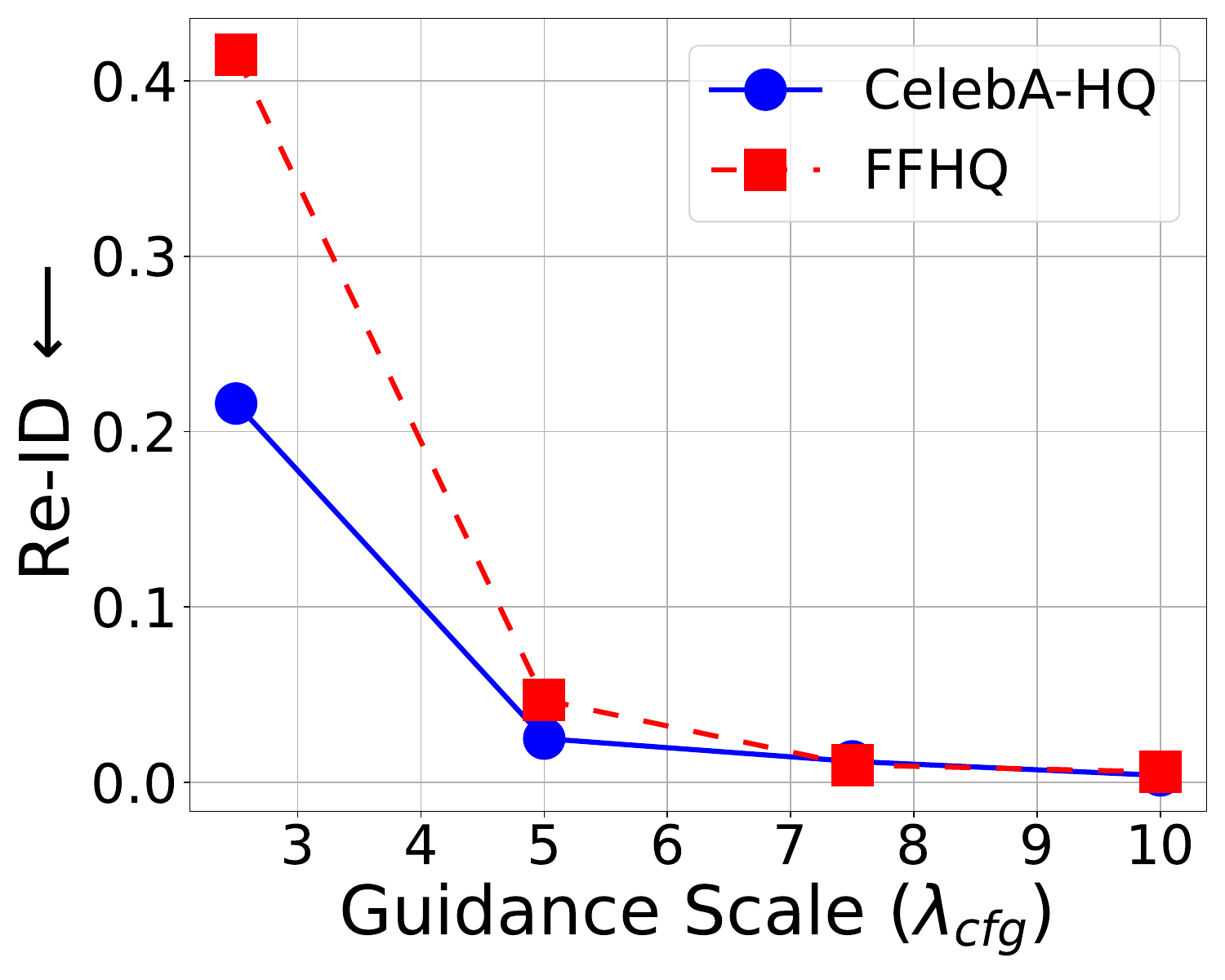}
    \caption{}
    \label{fig:cfg_impact_reid}
  \end{subfigure}
  \\
  \caption{Effects of key parameters on anonymization quality.}
\end{figure}

We quantitatively evaluate anonymization on 1,000 identities from CelebA-HQ~\cite{karras2017progressive} and FFHQ~\cite{karras2019style}, focusing on how the parameters \( T_{\text{skip}} \), \( \lambda_{id} \), and \( \lambda_{cfg} \) influence anonymization quality. Lowering \( T_{\text{skip}} \) increases identity distance (see \cref{fig:identity_distance_vs_Tskip}), but this comes at the cost of weaker alignment with the input image. For example, pose preservation degrades at smaller \( T_{\text{skip}} \) values, as reflected by higher pose distances in \cref{fig:pose_distance_vs_Tskip}. Similarly, increasing \( \lambda_{id} \) also enlarges identity distance (\cref{fig:identity_distance_vs_lambda_id}). For \( \lambda_{cfg} \), we measure re-identification rates---the proportion of anonymized faces correctly matched to their originals by recognition models. \Cref{fig:cfg_impact_reid} demonstrates that higher \( \lambda_{cfg} \) values decrease re-identification rates for both datasets.

\subsection{Segmentation Masks on Anonymization}

\begin{figure}[h]
  \scriptsize
  \centering
  \begin{tabular}{*{5}{>{\centering\arraybackslash}m{\dimexpr.2\linewidth-2\tabcolsep}}}
    & Change whole face & Keep eyes & Keep nose & Keep mouth \\
    \multicolumn{5}{@{}c@{}}{\includegraphics[width=.95\linewidth]{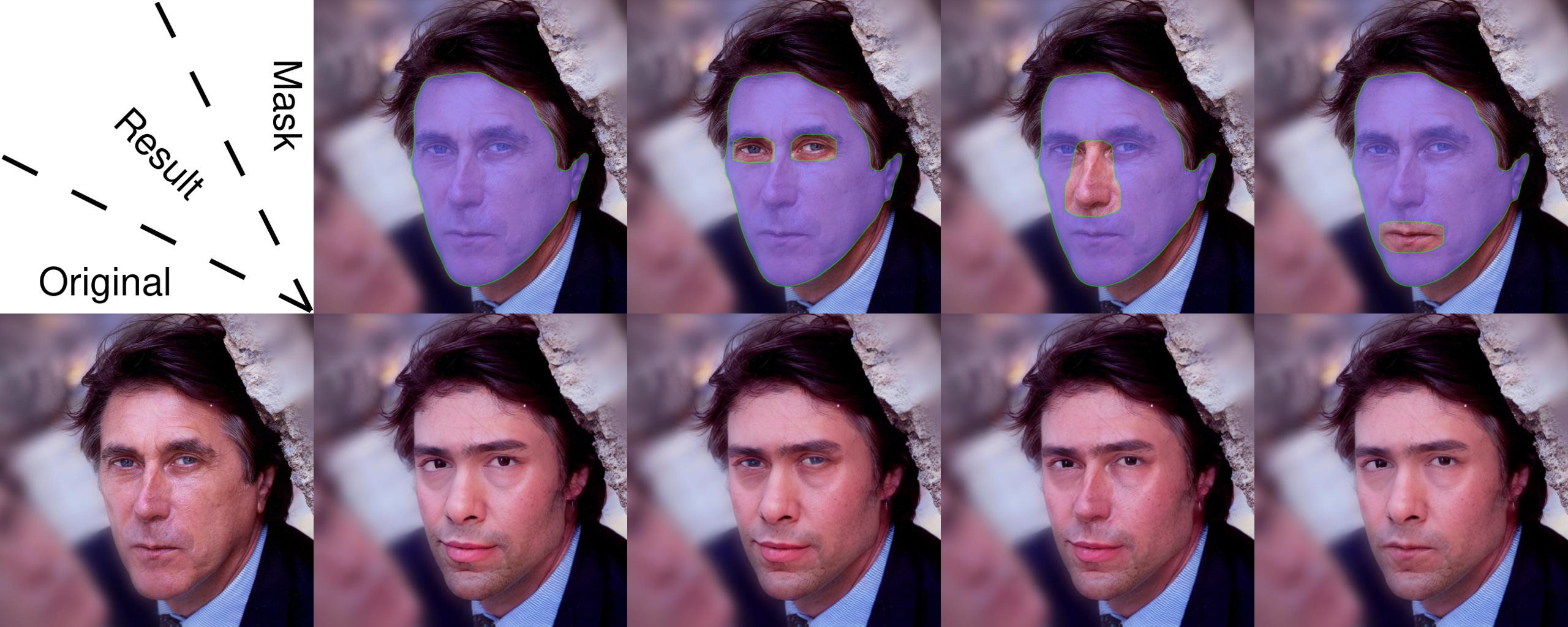}} \\
    \multicolumn{5}{@{}c@{}}{\includegraphics[width=.95\linewidth]{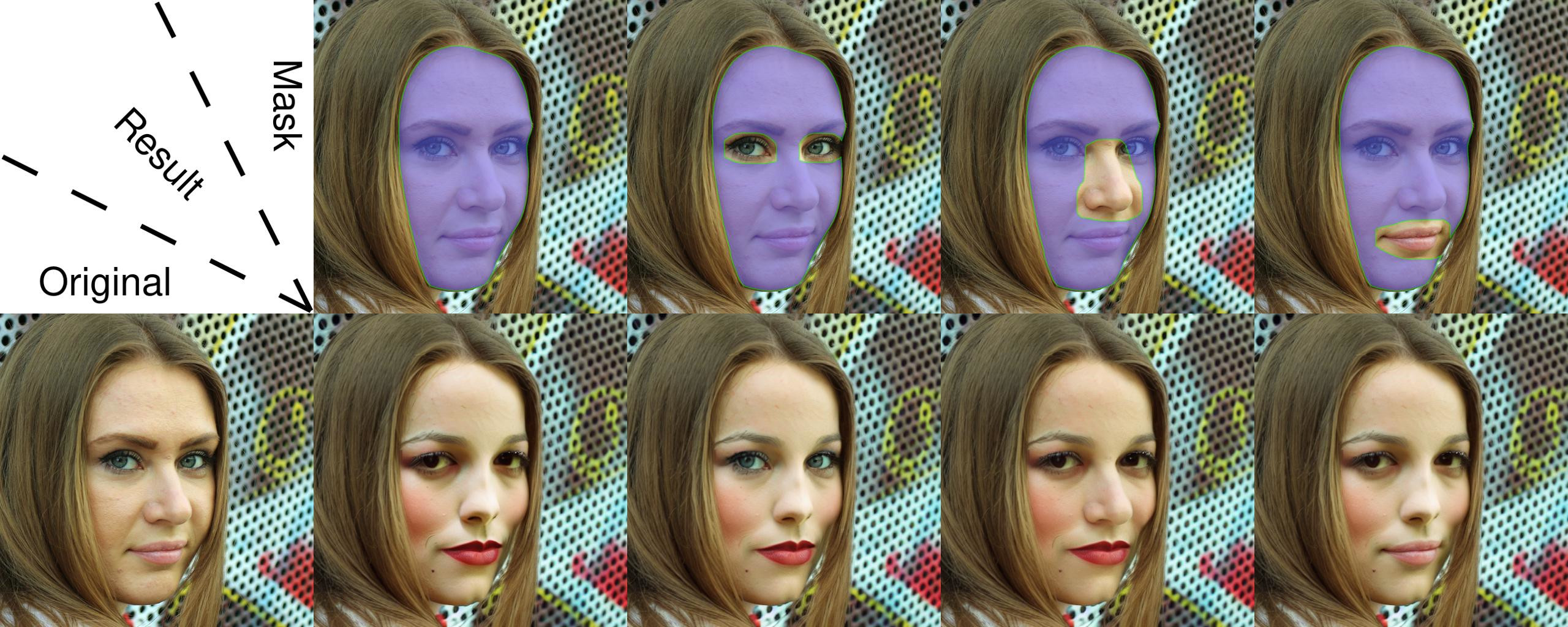}} \\
  \end{tabular}
  \caption{Localized anonymization using segmentation masks. The full-face mask fully anonymizes the face, while individual masks reveal specific facial regions (eyes, nose, mouth).}
  \label{fig:mask}
\end{figure}

While our method does not rely on segmentation masks, they offer additional control by specifying which facial regions are revealed or concealed. \Cref{fig:mask} illustrates this flexibility with various masking configurations. A full-face mask ensures complete anonymization, while masks selectively revealing only the eyes, nose, or mouth preserve those regions and anonymize the rest.

We also investigated the impact of mask application timing during the denoising process and found a tradeoff between attribute retention and anonymity. Detailed results, including re-identification rates, attribute preservation at different timesteps, and the effects of keeping various facial regions visible on anonymization performance, are provided in the supplementary materials.

\begin{table*}[t]
  \caption{Quantitative results on face anonymization methods for CelebA-HQ~\cite{karras2017progressive} (CHQ) and FFHQ~\cite{karras2019style} (FHQ) datasets. Best results are in bold, and second-best are underlined.}
  \label{tab:comp_quan_4k}
  \centering
  \resizebox{\textwidth}{!}{
    \begin{tabular}{l cc cc cc cc cc cc}
      \toprule
      & \multicolumn{4}{c}{Re-ID (\%) $\downarrow$} & \multicolumn{6}{c}{Attribute Distance $\downarrow$} & \multicolumn{2}{c}{Image Quality $\downarrow$} \\
      \cmidrule(lr){2-5} \cmidrule(lr){6-11} \cmidrule(lr){12-13}
      & \multicolumn{2}{c}{AdaFace} & \multicolumn{2}{c}{FaceNet} & \multicolumn{2}{c}{Expression} & \multicolumn{2}{c}{Gaze} & \multicolumn{2}{c}{Pose} & \multicolumn{2}{c}{FID} \\
      \cmidrule(lr){2-3} \cmidrule(lr){4-5} \cmidrule(lr){6-7} \cmidrule(lr){8-9} \cmidrule(lr){10-11} \cmidrule(lr){12-13}
      & CHQ & FHQ & CHQ & FHQ & CHQ & FHQ & CHQ & FHQ & CHQ & FHQ & CHQ & FHQ \\
      \midrule
      Ours & \underline{0.208} & \textbf{0.343} & \underline{0.289} & \underline{0.383} & 10.034 & \underline{9.852} & \underline{0.165} & \underline{0.186} & \textbf{0.053} & \underline{0.056} & \underline{8.223} & \textbf{8.779} \\
      FAMS~\cite{kung2025face} & 3.131 & 14.152 & 0.866 & 5.570 & 10.001 & \textbf{8.823} & \textbf{0.164} & \textbf{0.176} & \textbf{0.053} & \textbf{0.047} & 17.128 & 11.215 \\
      FALCO~\cite{barattin2023attribute} & \textbf{0.104} & - & \textbf{0.100} & - & 10.208 & - & 0.277 & - & 0.088 & - & 39.168 & - \\
      RiDDLE~\cite{li2023riddle} & - & \underline{0.510} & - & \textbf{0.042} & - & 10.038 & - & 0.215 & - & 0.081 & - & 69.259 \\
      LDFA~\cite{klemp2023ldfa} & 10.284 & 11.152 & 4.275 & 4.925 & \textbf{8.647} & 10.387 & 0.260 & 0.353 & 0.092 & 0.113 & \textbf{8.058} & \underline{9.946} \\
      DP2~\cite{hukkelaas2023deepprivacy2} & 0.835 & 1.881 & 0.722 & 1.186 & \underline{9.912} & 10.183 & 0.262 & 0.299 & 0.161 & 0.163 & 16.935 & 18.632 \\
      \bottomrule
    \end{tabular}
  }
\end{table*}

\subsection{Comparisons with Baselines}
\label{sec:comp_base}

We evaluated our method on the CelebA-HQ~\cite{karras2017progressive} and FFHQ~\cite{karras2019style} datasets, comparing it against five baselines (FAMS~\cite{kung2025face}, FALCO~\cite{barattin2023attribute}, RiDDLE~\cite{li2023riddle}, LDFA~\cite{klemp2023ldfa}, and DP2~\cite{hukkelaas2023deepprivacy2}) using quantitative and qualitative metrics. Our method used \( T_{\text{skip}} = 70 \), \( \lambda_{id} = 1.0 \), \( \lambda_{cfg} = 10 \), and an eye-and-mouth-revealing mask applied at timestep 80.

\paragraph{Quantitative analysis.}

To evaluate our model's performance, we consider three aspects: identity anonymization, preservation of non-identity-related attributes, and image quality. Identity anonymization is quantified using the re-identification rate---the proportion of anonymized faces correctly matched to their originals by recognition models. For fairness, we employ two models different from the one used during anonymization: AdaFace~\cite{kim2022adaface} and FaceNet~\cite{schroff2015facenet}. Non-identity-related attributes include head pose, eye gaze, and facial expression. Head pose and gaze distances are computed from angular differences in predicted orientations~\cite{ruiz2018fine} and gaze directions~\cite{abdelrahman2023l2cs}, respectively, while expression distance is measured via 3DMM coefficients~\cite{blanz2023morphable}. Finally, image quality is assessed using the Fréchet Inception Distance (FID)~\cite{heusel2017gans}, a metric commonly used in prior studies~\cite{zhai2022a3gan,hukkelaas2023deepprivacy2,dall2022graph,helou2023vera,barattin2023attribute} to quantify the realism of generated images by comparing their statistics with those of real images.

\begin{figure*}[h]
  \centering

  \par\medskip 
  \begin{subfigure}{1.0\linewidth}
    \centering
    \includegraphics[width=0.24\linewidth]{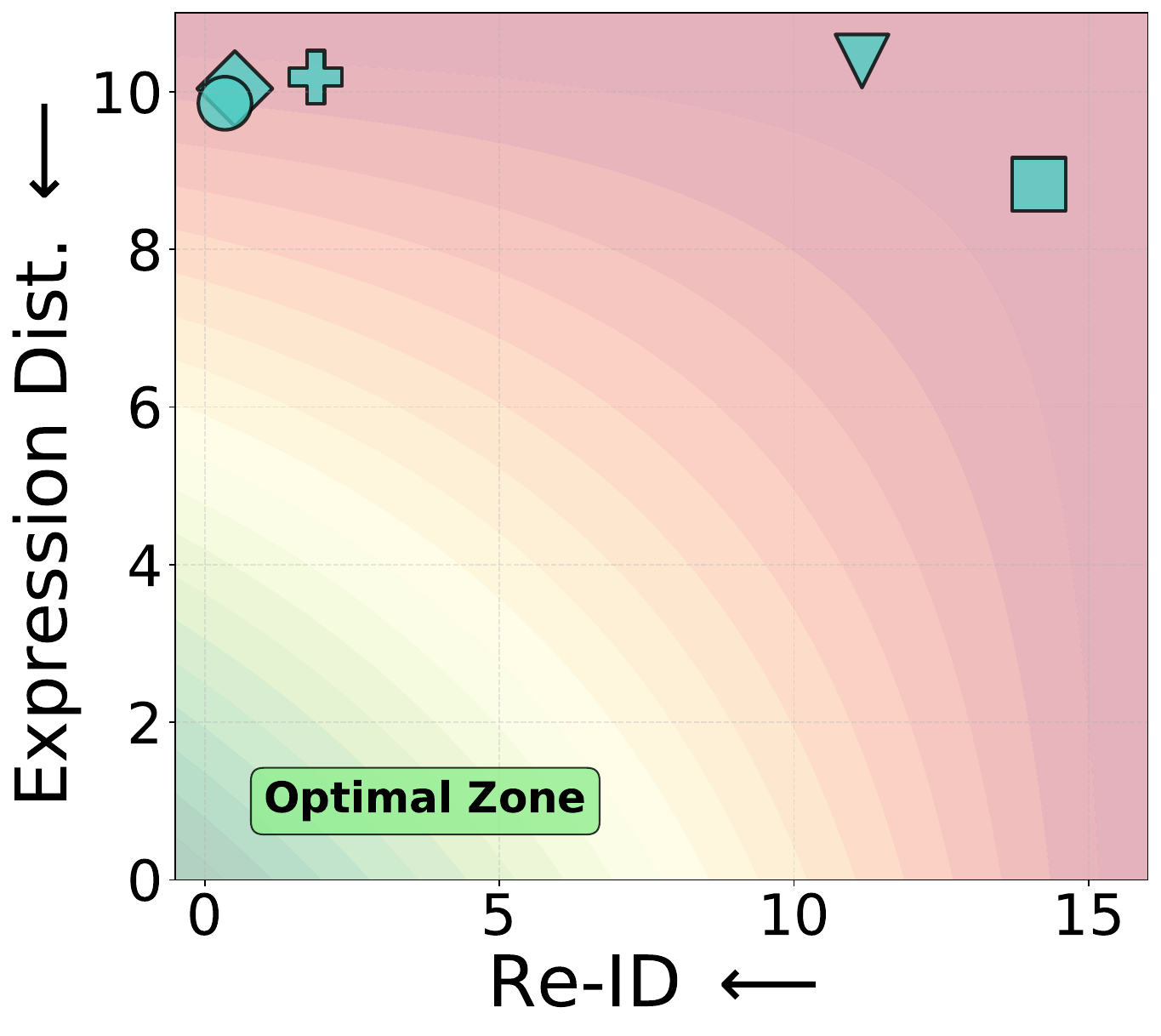}
    \hfill
    \includegraphics[width=0.24\linewidth]{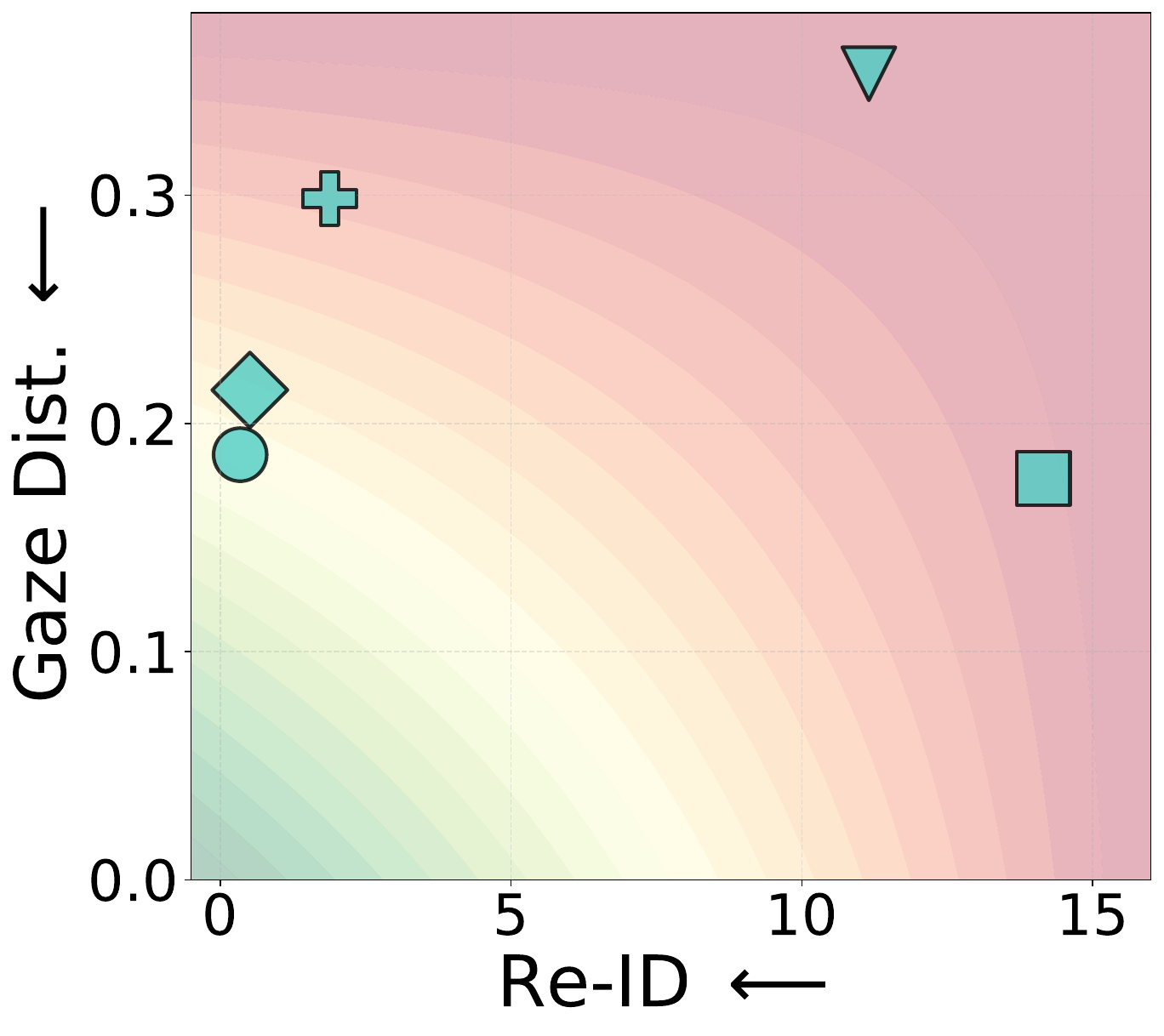}
    \hfill
    \includegraphics[width=0.24\linewidth]{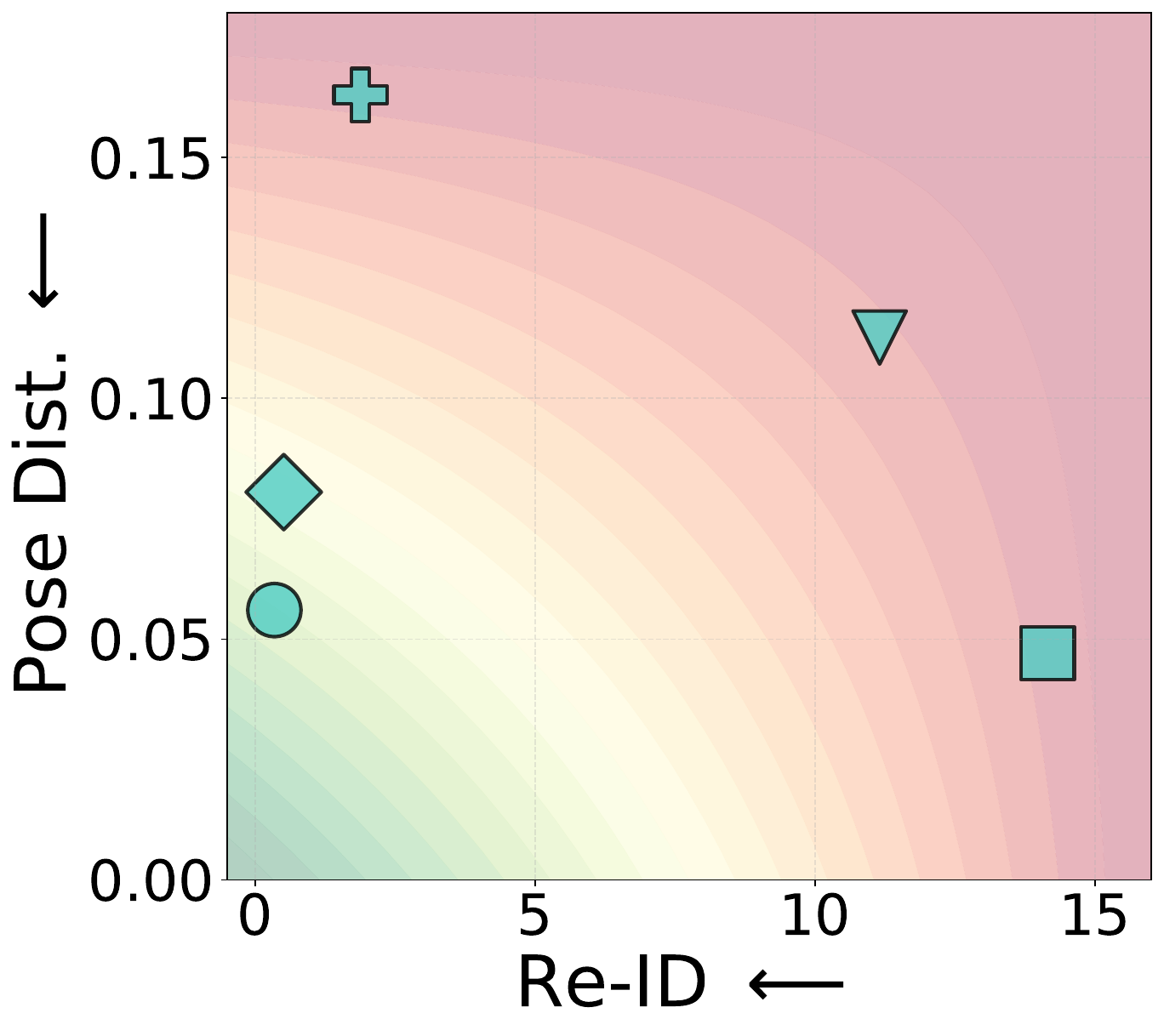}
    \hfill
    \includegraphics[width=0.24\linewidth]{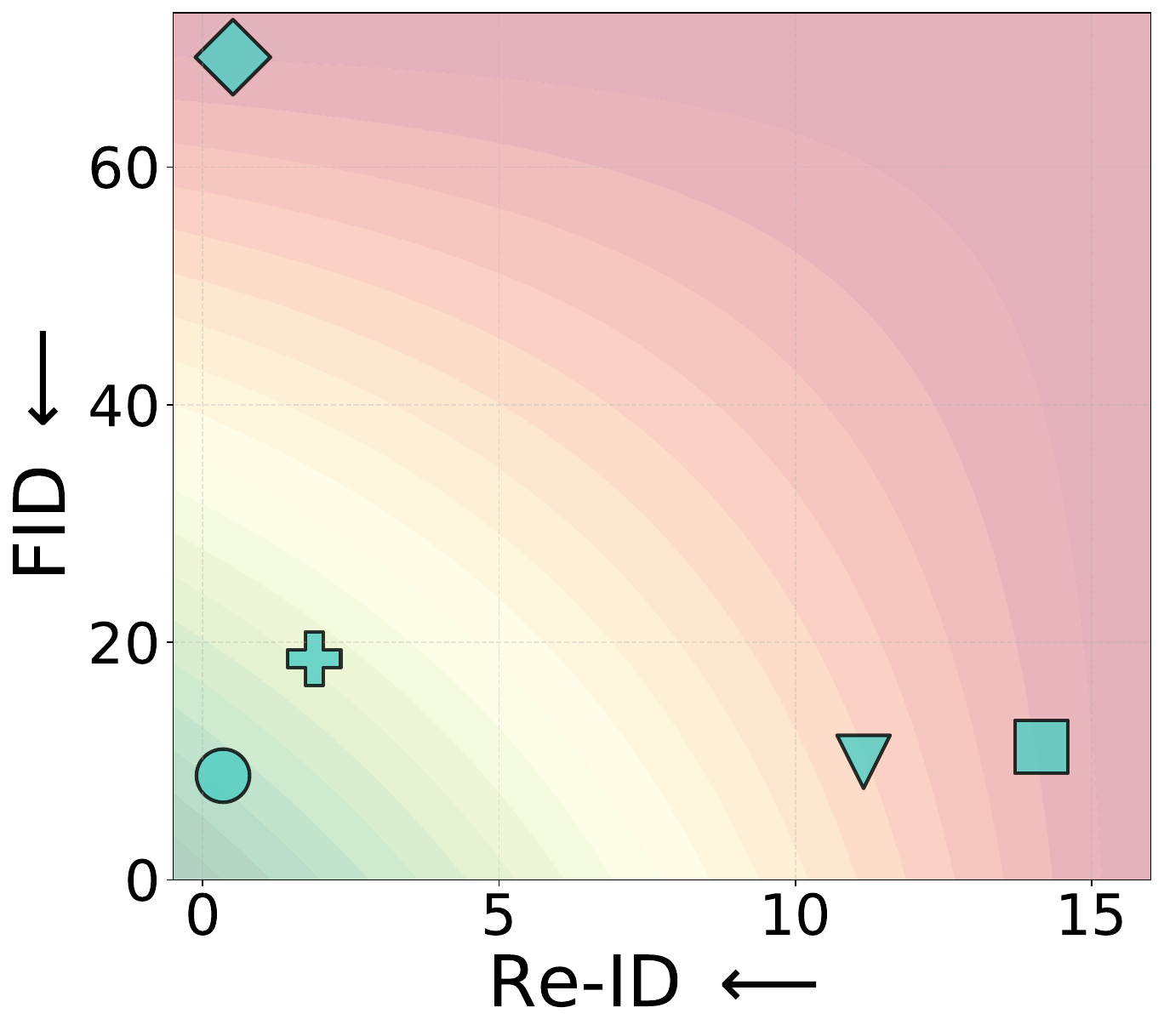}
  \end{subfigure}

  \par\medskip 
  \begin{subfigure}{1.0\linewidth}
    \centering
    \includegraphics[width=0.24\linewidth]{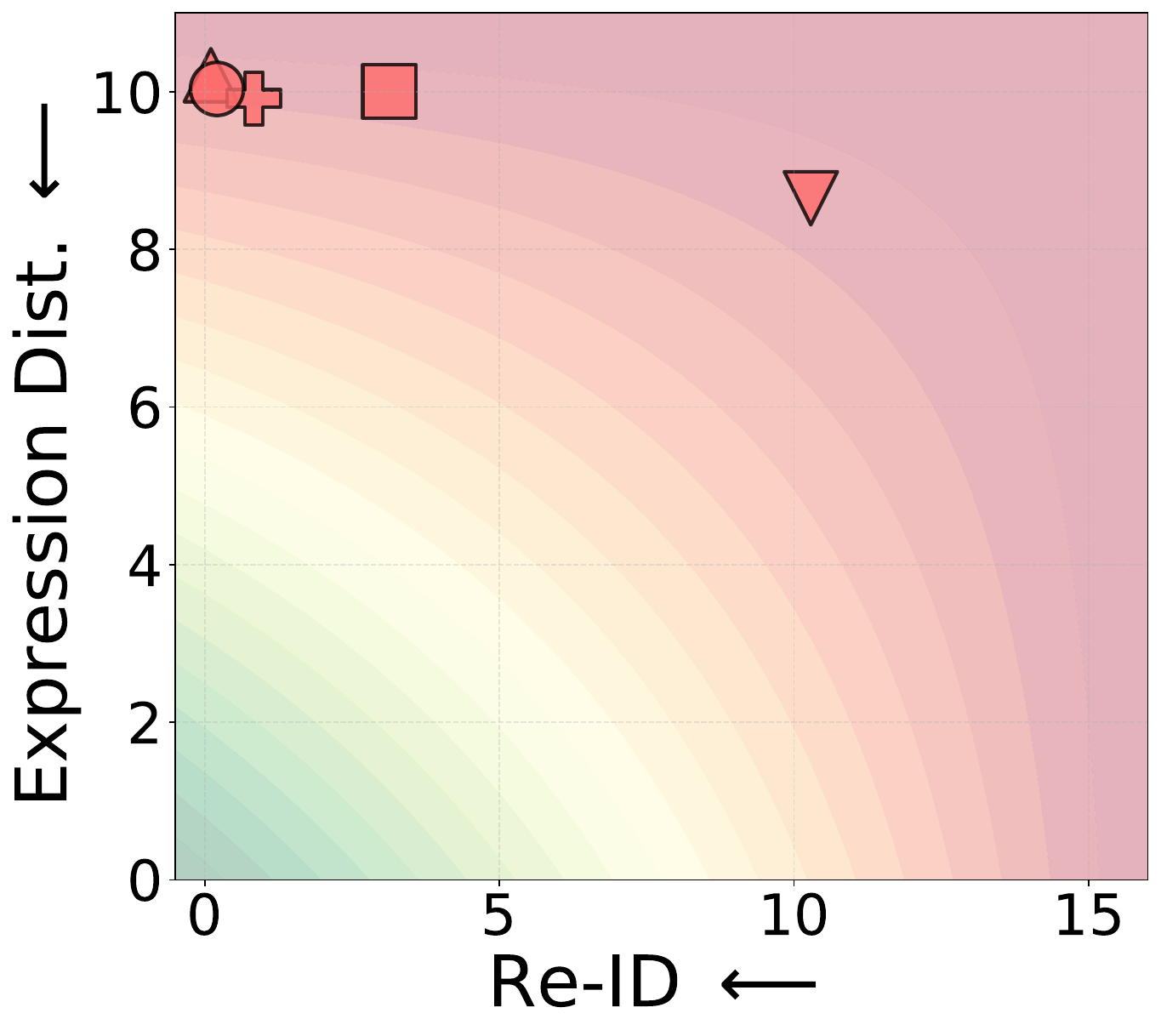}
    \hfill
    \includegraphics[width=0.24\linewidth]{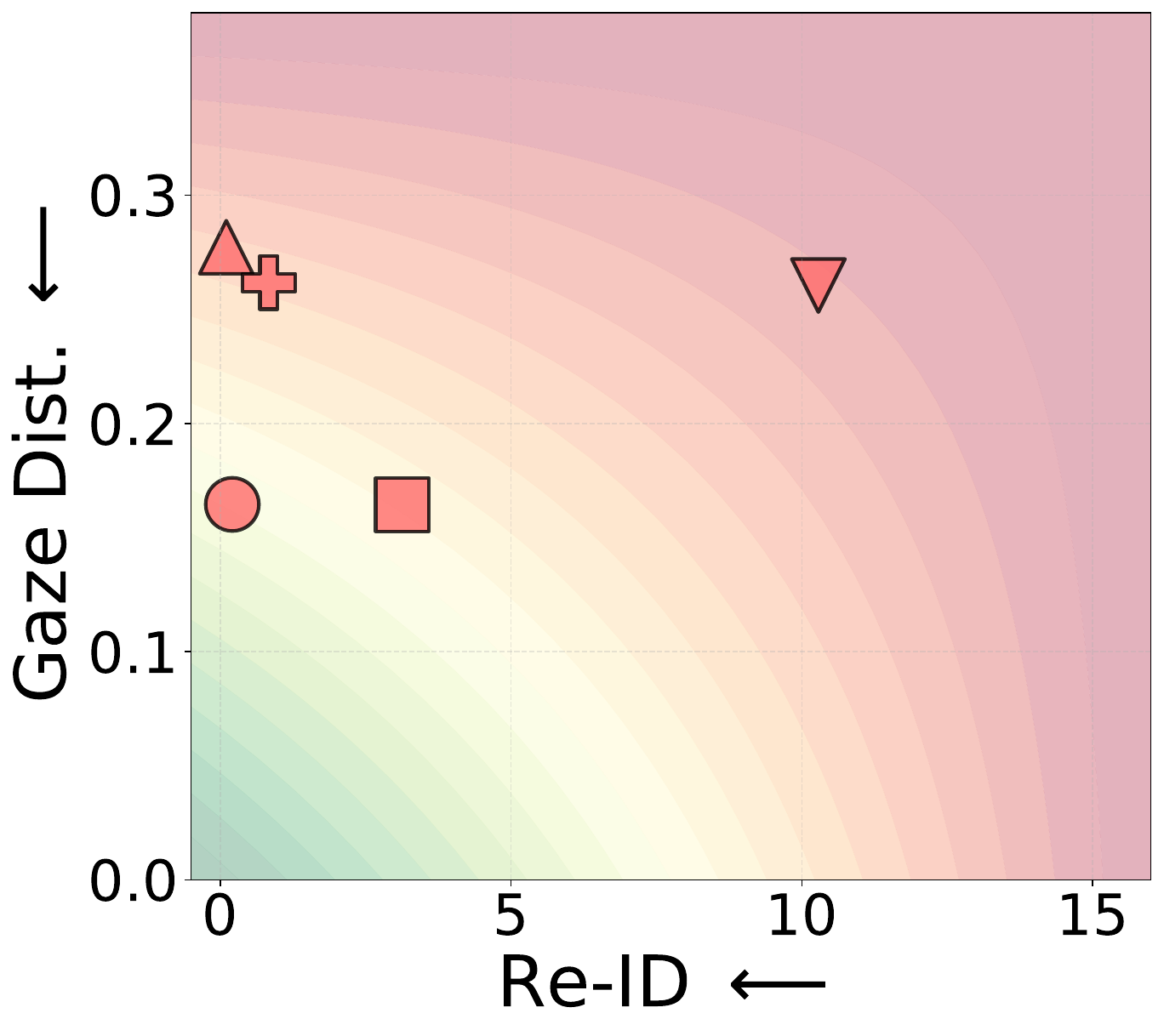}
    \hfill
    \includegraphics[width=0.24\linewidth]{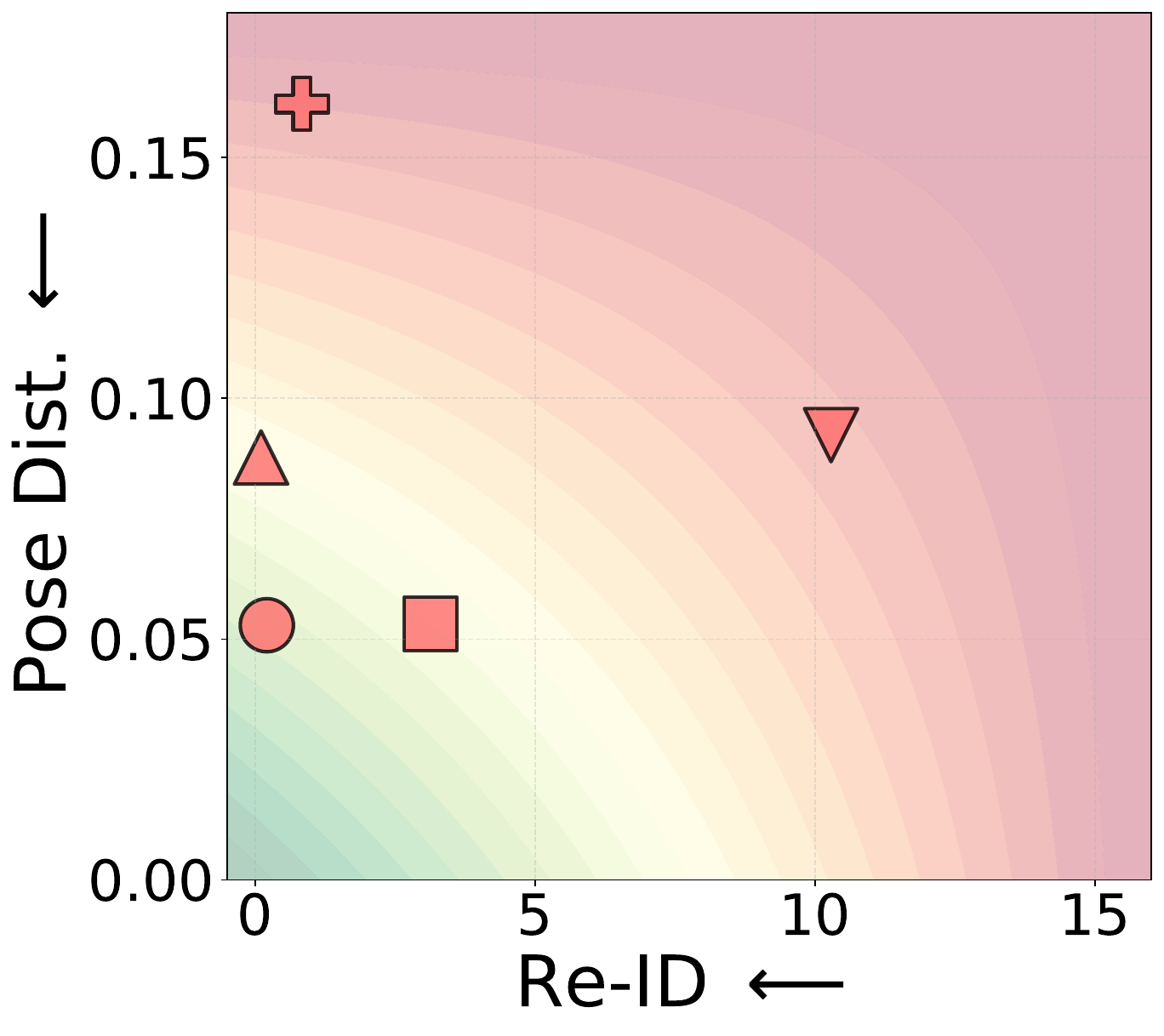}
    \hfill
    \includegraphics[width=0.24\linewidth]{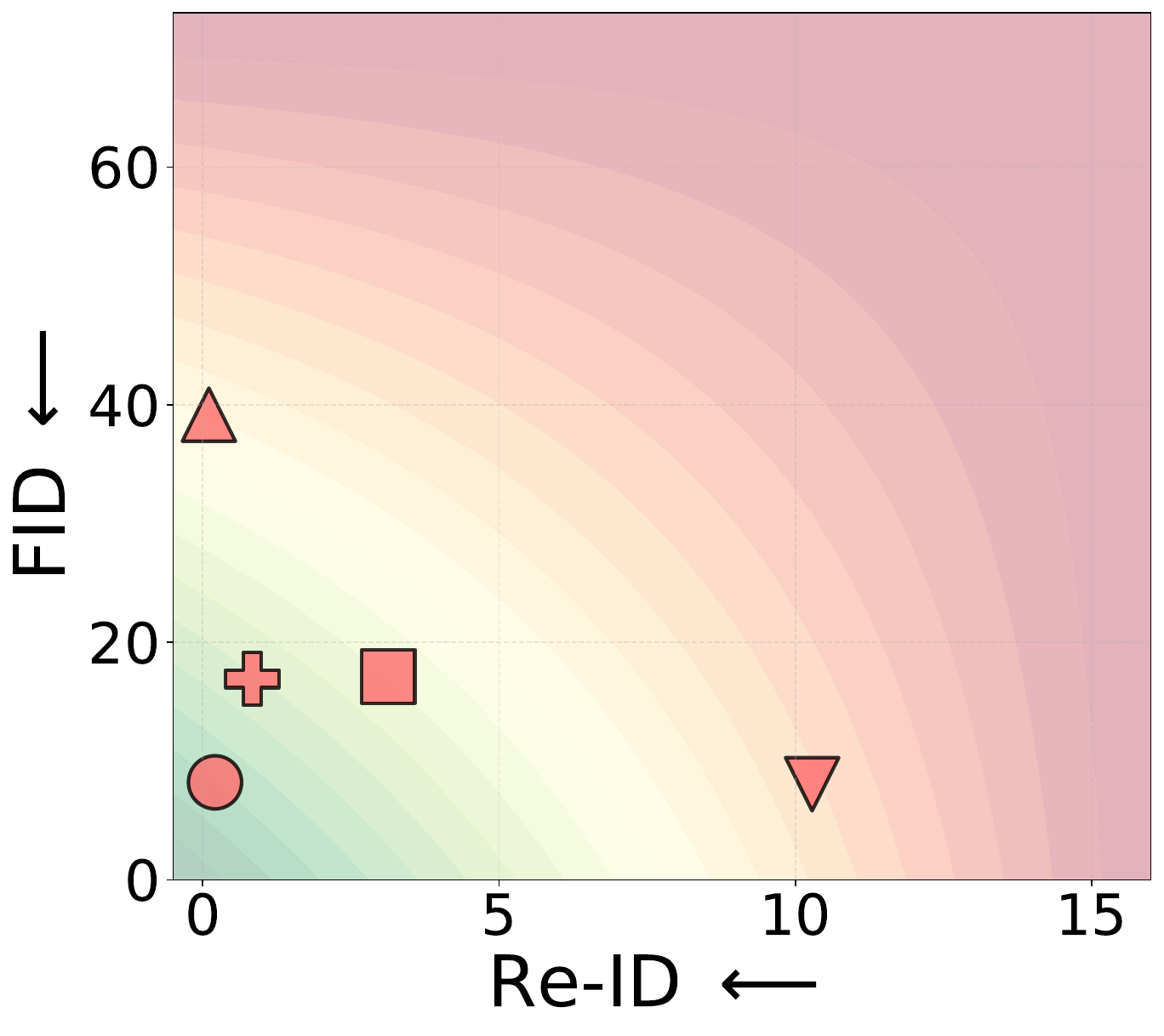}
  \end{subfigure}

  \par\medskip 
  \begin{subfigure}{1.0\linewidth}
    \centering
    \includegraphics[width=.75\linewidth]{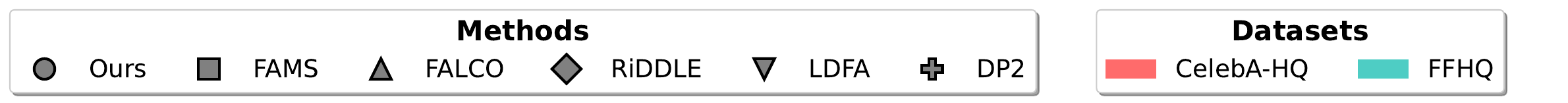}
  \end{subfigure}

  \caption{Privacy-utility trade-off evaluation using the AdaFace~\cite{kim2022adaface} model and four utility metrics (expression, gaze, pose, FID~\cite{heusel2017gans}) on CelebA-HQ~\cite{karras2017progressive} and FFHQ~\cite{karras2019style} datasets. The green gradient highlights the optimal zone (lower-left) where both privacy protection (low Re-ID) and utility preservation (low metric values) are achieved.}
  \label{fig:trade-offs-adaface}
\end{figure*}

\Cref{tab:comp_quan_4k} presents a quantitative comparison across 4,852 CelebA-HQ~\cite{karras2017progressive} and 4,722 FFHQ~\cite{karras2019style} test subjects. \Cref{fig:trade-offs-adaface,fig:trade-offs-facenet} illustrate the corresponding privacy-utility trade-offs using AdaFace~\cite{kim2022adaface} and FaceNet~\cite{schroff2015facenet} for re-identification rate calculations, respectively (\cref{fig:trade-offs-facenet} is provided in the supplementary material), with optimal zones (low re-identification rates, low utility metric values) highlighted in green. Both figures show that our method offers a better balance across all evaluated metrics---expression, gaze, pose, and image quality---compared to baseline methods, which have trade-offs. Specifically, FALCO~\cite{barattin2023attribute} and RiDDLE~\cite{li2023riddle} attain low re-identification rates but at the cost of utility: FALCO~\cite{barattin2023attribute} exhibits larger expression and gaze distances, while RiDDLE~\cite{li2023riddle} suffers from degraded image quality. In contrast, FAMS~\cite{kung2025face} preserves utility well but compromises privacy, and LDFA~\cite{klemp2023ldfa} yields higher re-identification rates despite comparable FID~\cite{heusel2017gans}. Finally, DP2~\cite{hukkelaas2023deepprivacy2} shows poor pose preservation across both datasets.

\begin{table*}[h]
  \caption{Ablation study of face anonymization implementations for CelebA-HQ~\cite{karras2017progressive} and FFHQ~\cite{karras2019style} datasets.}
  \label{tab:abla_adaface}
  \centering
  \resizebox{\textwidth}{!}{
    \begin{tabular}{l cc cc cc cc cc cc}
      \toprule
      & \multicolumn{4}{c}{Re-ID (\%) $\downarrow$} & \multicolumn{6}{c}{Attribute Distance $\downarrow$} & \multicolumn{2}{c}{Image Quality $\downarrow$} \\
      \cmidrule(lr){2-5} \cmidrule(lr){6-11} \cmidrule(lr){12-13}
      & \multicolumn{2}{c}{AdaFace} & \multicolumn{2}{c}{FaceNet} & \multicolumn{2}{c}{Expression} & \multicolumn{2}{c}{Gaze} & \multicolumn{2}{c}{Pose} & \multicolumn{2}{c}{FID} \\
      \cmidrule(lr){2-3} \cmidrule(lr){4-5} \cmidrule(lr){6-7} \cmidrule(lr){8-9} \cmidrule(lr){10-11} \cmidrule(lr){12-13}
      & CHQ & FHQ & CHQ & FHQ & CHQ & FHQ & CHQ & FHQ & CHQ & FHQ & CHQ & FHQ \\
      \midrule
      Ours & 0.10 & 0.50 & 0.40 & 0.60 & \textbf{11.621} & \textbf{10.941} & \textbf{0.214} & \textbf{0.237} & \textbf{0.058} & \textbf{0.060} & \textbf{18.838} & \textbf{21.121} \\
      Inpainting~\cite{rombach2022high} & \textbf{0.00} & \textbf{0.10} & \textbf{0.10} & \textbf{0.10} & 18.143 & 17.922 & 0.293 & 0.339 & 0.127 & 0.142 & 35.802 & 40.213 \\
      \bottomrule
    \end{tabular}
  }
\end{table*}

\paragraph{Qualitative analysis.}

\begin{figure}[h]
  \scriptsize
  \centering
  \begin{tabularx}{\linewidth}{@{}X@{}X@{}X@{}X@{}X@{}X@{}}
    \multicolumn{6}{@{}c@{}}{\includegraphics[width=\linewidth]{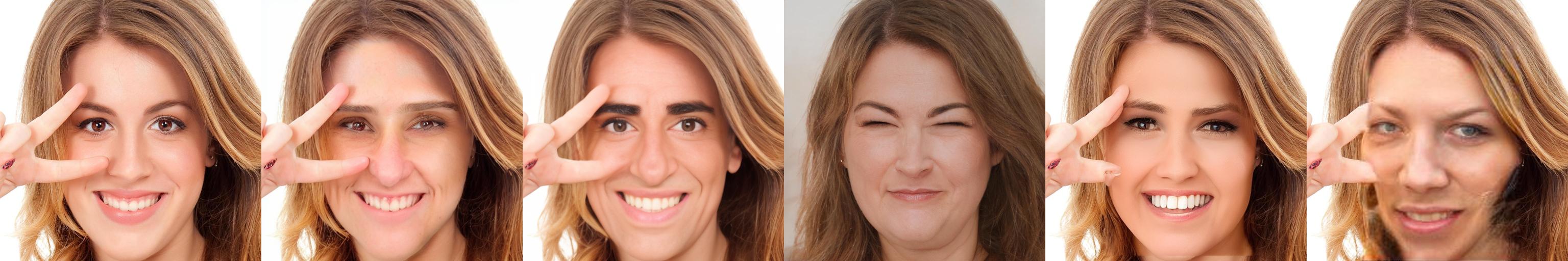}} \\
    \multicolumn{6}{@{}c@{}}{\includegraphics[width=\linewidth]{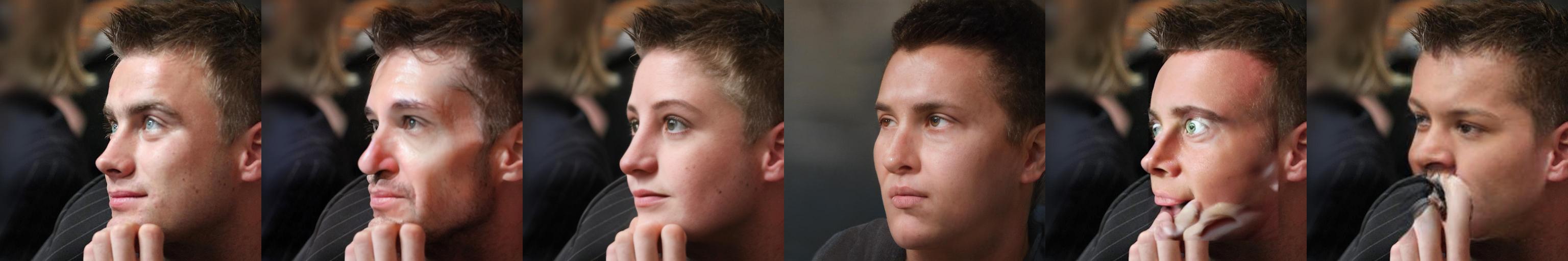}} \\
    \multicolumn{6}{@{}c@{}}{\includegraphics[width=\linewidth]{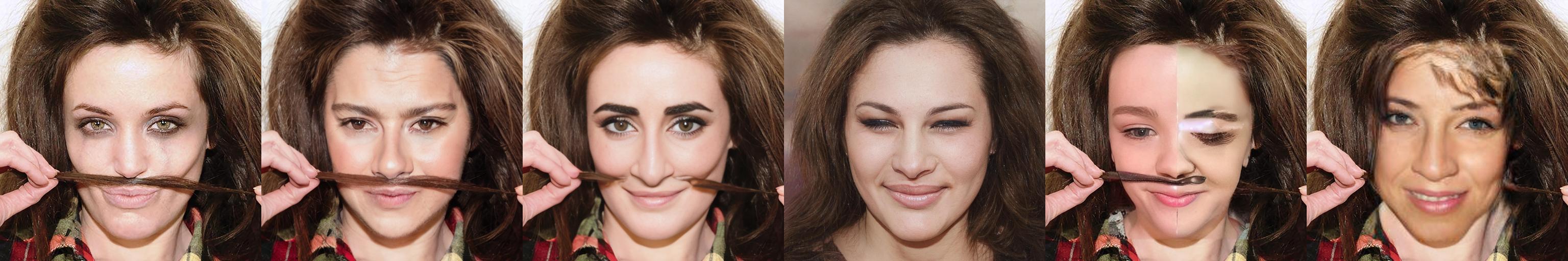}} \\
    \multicolumn{6}{@{}c@{}}{\includegraphics[width=\linewidth]{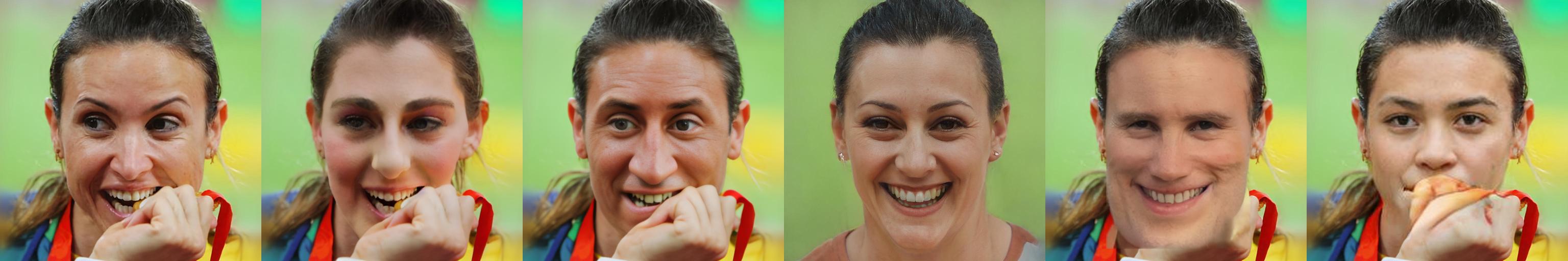}} \\
    \centering Original & \centering Ours & \centering FAMS~\cite{kung2025face} & \centering FALCO~\cite{barattin2023attribute} & \centering LDFA~\cite{klemp2023ldfa} & \centering DP2~\cite{hukkelaas2023deepprivacy2} \\
  \end{tabularx}
  \caption{Face anonymization results on CelebA-HQ~\cite{karras2017progressive} set.}
  \label{fig:comp_cele}
\end{figure}

\begin{figure}[h]
  \scriptsize
  \centering
  \begin{tabularx}{\columnwidth}{@{}X@{}X@{}X@{}X@{}X@{}X@{}}
    \multicolumn{6}{@{}c@{}}{\includegraphics[width=\linewidth]{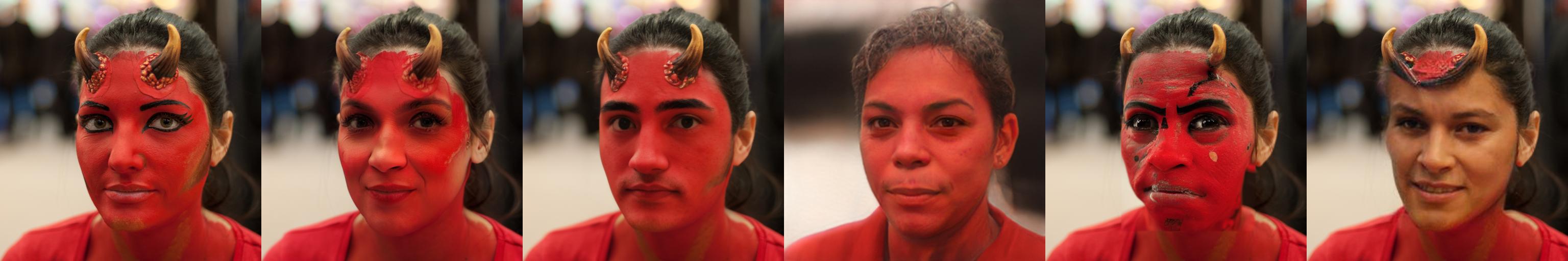}} \\
    \multicolumn{6}{@{}c@{}}{\includegraphics[width=\linewidth]{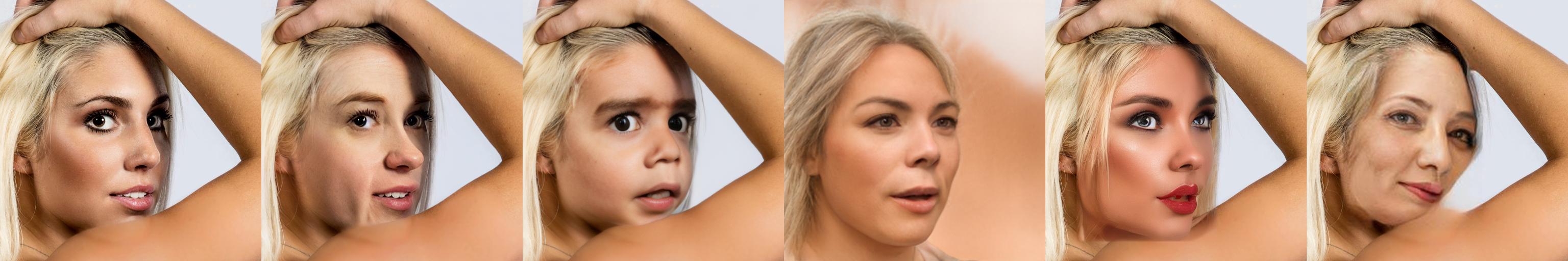}} \\
    \multicolumn{6}{@{}c@{}}{\includegraphics[width=\linewidth]{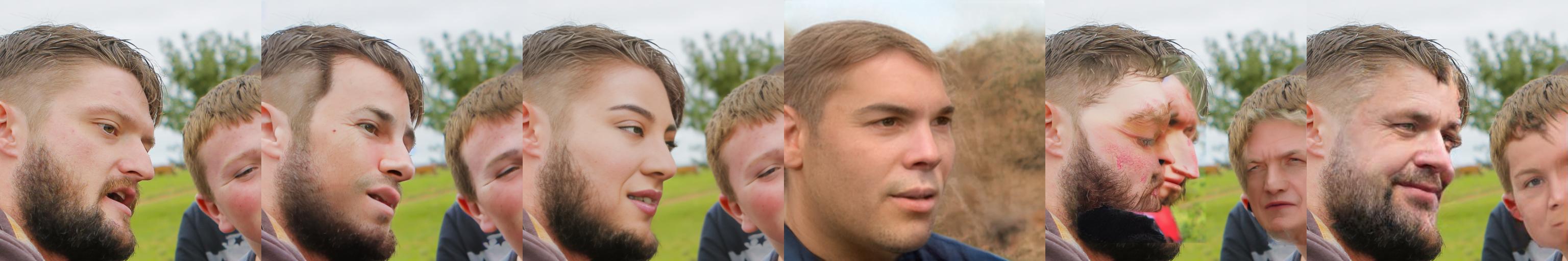}} \\
    \multicolumn{6}{@{}c@{}}{\includegraphics[width=\linewidth]{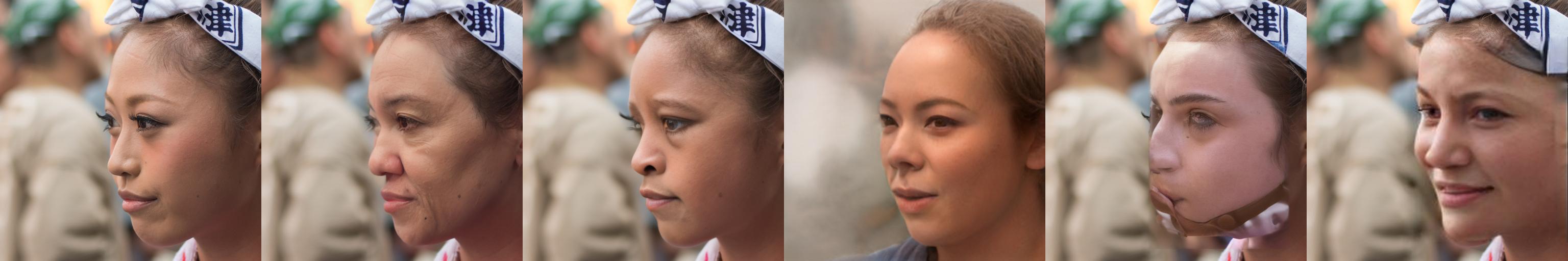}} \\
    \centering Original & \centering Ours & \centering FAMS~\cite{kung2025face} & \centering RiDDLE~\cite{li2023riddle} & \centering LDFA~\cite{klemp2023ldfa} & \centering DP2~\cite{hukkelaas2023deepprivacy2} \\
  \end{tabularx}
  \caption{Face anonymization results on FFHQ~\cite{karras2019style} set.}
  \label{fig:comp_ffhq}
\end{figure}

\cref{fig:comp_cele,fig:comp_ffhq} present qualitative results comparing our method to baseline approaches on challenging cases from CelebA-HQ~\cite{karras2017progressive} and FFHQ~\cite{karras2019style}. These include faces with occlusions or extreme angles. Our method effectively anonymized identities while preserving non-identity-related details, such as pose, expression, gaze, and background information, and produced photorealistic outputs. In contrast, FALCO~\cite{barattin2023attribute} and RiDDLE~\cite{li2023riddle} failed to preserve background details, DP2~\cite{hukkelaas2023deepprivacy2} struggled with pose consistency, and LDFA~\cite{klemp2023ldfa} generated less realistic outputs. Additional qualitative examples are provided in the supplementary material.


\subsection{Ablation Studies}

We assess the impact of the inversion component in our face anonymization framework by comparing it to the Stable Diffusion Inpainting baseline~\cite{rombach2022high}. We anonymized 1,000 test subjects each from the CelebA-HQ~\cite{karras2017progressive} and FFHQ~\cite{karras2019style} datasets and conducted a quantitative evaluation based on the same metrics outlined in \cref{sec:comp_base}, focusing on identity anonymization, attribute retention, and image quality. For the baseline, we removed the inversion step from our framework and replaced the standard Stable Diffusion model with the Stable Diffusion Inpainting model. To ensure fairness, the inpainting baseline used the same identity conditioning as our method, and both methods applied identical masks covering the entire face and used the same guidance scale value of 10. For our method, we maintained a \( T_{\text{skip}} \) of 70 and an anonymization parameter \( \lambda_{id} \) of 1.0.

\Cref{tab:abla_adaface} shows that our method outperforms the inpainting baseline in attribute preservation across all metrics and produces images that more closely match the original quality. While the inpainting baseline achieves slightly lower re-identification rates, it falls significantly short in maintaining attribute fidelity. This limitation could make the inpainting approach unsuitable for applications requiring retention of non-identity-related attributes.

\subsection{Localized Anonymization Applications}

\begin{figure}[h]
  \centering
  \resizebox{0.8\columnwidth}{!}{
    \begin{tabular}{
        *{3}{>{\centering\arraybackslash}m{\dimexpr.33\linewidth-2\tabcolsep-.33\arrayrulewidth}}
      }
      \hline
      \multicolumn{1}{c}{Original} & \multicolumn{1}{c}{Mask (keep cheek)} & \multicolumn{1}{c}{Anonymized} \\ [\defaultaddspace]
      \multicolumn{1}{c}{\includegraphics[width={\dimexpr.33\linewidth}]{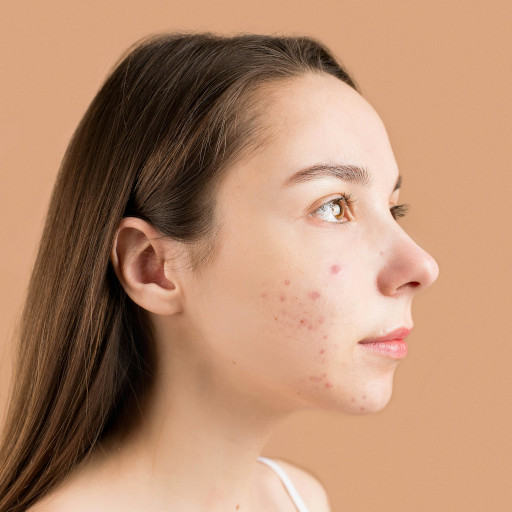}} & \multicolumn{1}{c}{\includegraphics[width={\dimexpr.33\linewidth}]{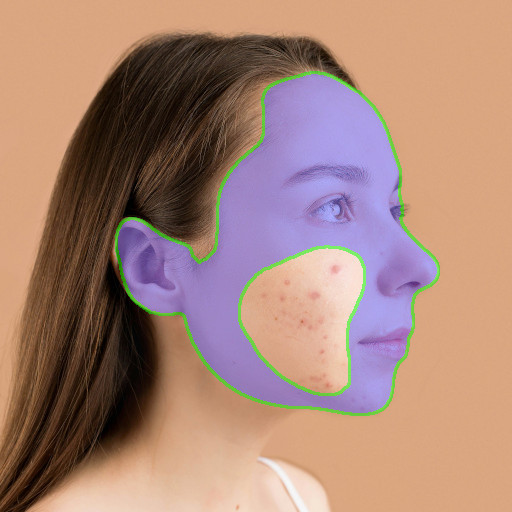}} & \multicolumn{1}{c}{\includegraphics[width={\dimexpr.33\linewidth}]{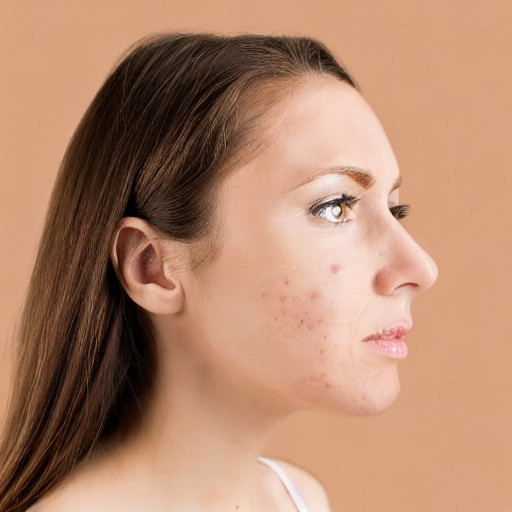}} \\
    \end{tabular}
  }
  \caption{Demonstration of localized facial anonymization preserving dermatological symptoms (acne) while anonymizing identity for medical image sharing.}
  \label{fig:app}
\end{figure}

The motivation for selectively anonymizing facial regions stems from real-world scenarios where complete face obfuscation would eliminate diagnostic or analytical information. In medical contexts, practitioners often need to document and share facial symptoms (such as dermatological conditions~\cite{rudd2022facial,nathanson2006medical,mayanja2002chronic}, asymmetries~\cite{lantieri2016face}, or expressions~\cite{zan2024autologous}) while maintaining patient confidentiality. Our method enables healthcare professionals to share case studies with colleagues for consultation or educational purposes without compromising patient privacy, as demonstrated in \cref{fig:app} where facial identity is anonymized while preserving pimples on the cheeks. Similarly, in human-computer interaction and behavioral research, preserving specific regions like the eyes while anonymizing other facial features allows for privacy-preserved gaze analysis~\cite{du2024privategaze}, facilitating studies on attention patterns~\cite{shagass1976eye}, emotional responses~\cite{armstrong2012eye,noyes2023eye}, and user experience~\cite{bergstrom2014eye} without unnecessary exposure of subjects' identities. These applications highlight that localized anonymization is not merely a face-editing technique but a privacy-enhancing method with clear practical benefits in sensitive domains.

\section{CONCLUSIONS}

We developed a face anonymization method that uses diffusion model inversion without requiring additional training. It allows precise control over which facial features remain visible. This approach established new standards for privacy-protected facial data use. While our current diffusion-based method has longer generation times compared to traditional GAN-based approaches, emerging advances in Flow~\cite{esser2024scaling} and SDXL-Turbo~\cite{sauer2024adversarial} models promise significant speed improvements. These faster diffusion models will likely accelerate research adoption in this field.


\section*{ACKNOWLEDGEMENTS}

This work was supported by the EU Horizon project AI4Trust (No. 101070190) and by the the FIS project GUIDANCE (No. FIS2023-03251). We acknowledge ISCRA for awarding this project access to the LEONARDO supercomputer, owned by the EuroHPC Joint Undertaking, hosted by CINECA (Italy), and thank the Finnish Foundation for Technology Promotion.


{
  \small
  \bibliographystyle{ieee}
  \bibliography{main}
}

\input{suppl}

\end{document}

%% file: suppl.tex
\clearpage
\setcounter{page}{1}
\maketitlesupplementary

\section{IMPACT OF MASK APPLICATION TIMING ON ANONYMIZATION AND ATTRIBUTE RETENTION}

\begin{figure}[h]
  \centering
  \begin{subfigure}{0.49\linewidth}
    \centering
    \includegraphics[width=1.0\linewidth]{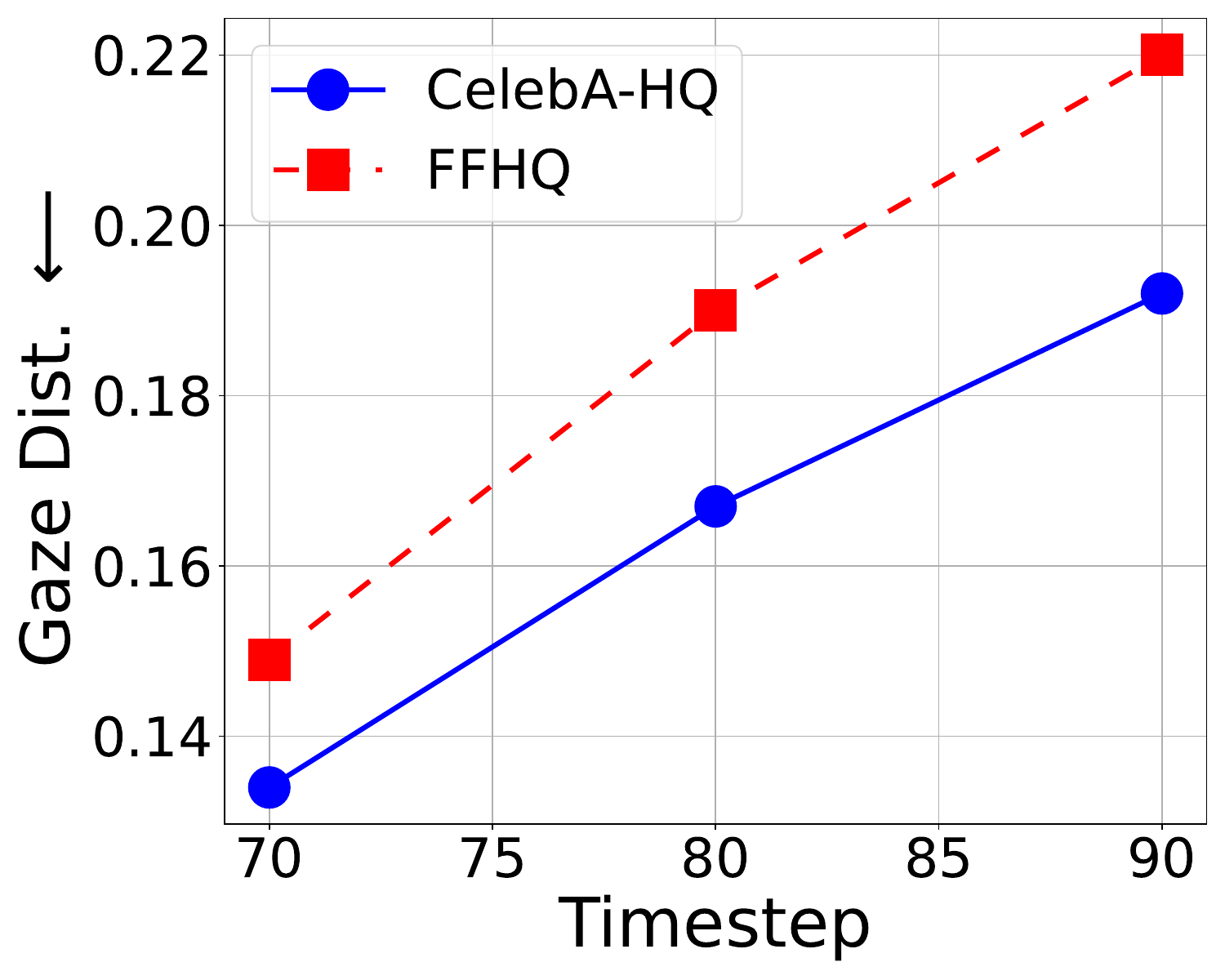}
    \caption{}
    \label{fig:features-a}
  \end{subfigure}
  \hfill
  \begin{subfigure}{0.49\linewidth}
    \centering
    \includegraphics[width=1.0\linewidth]{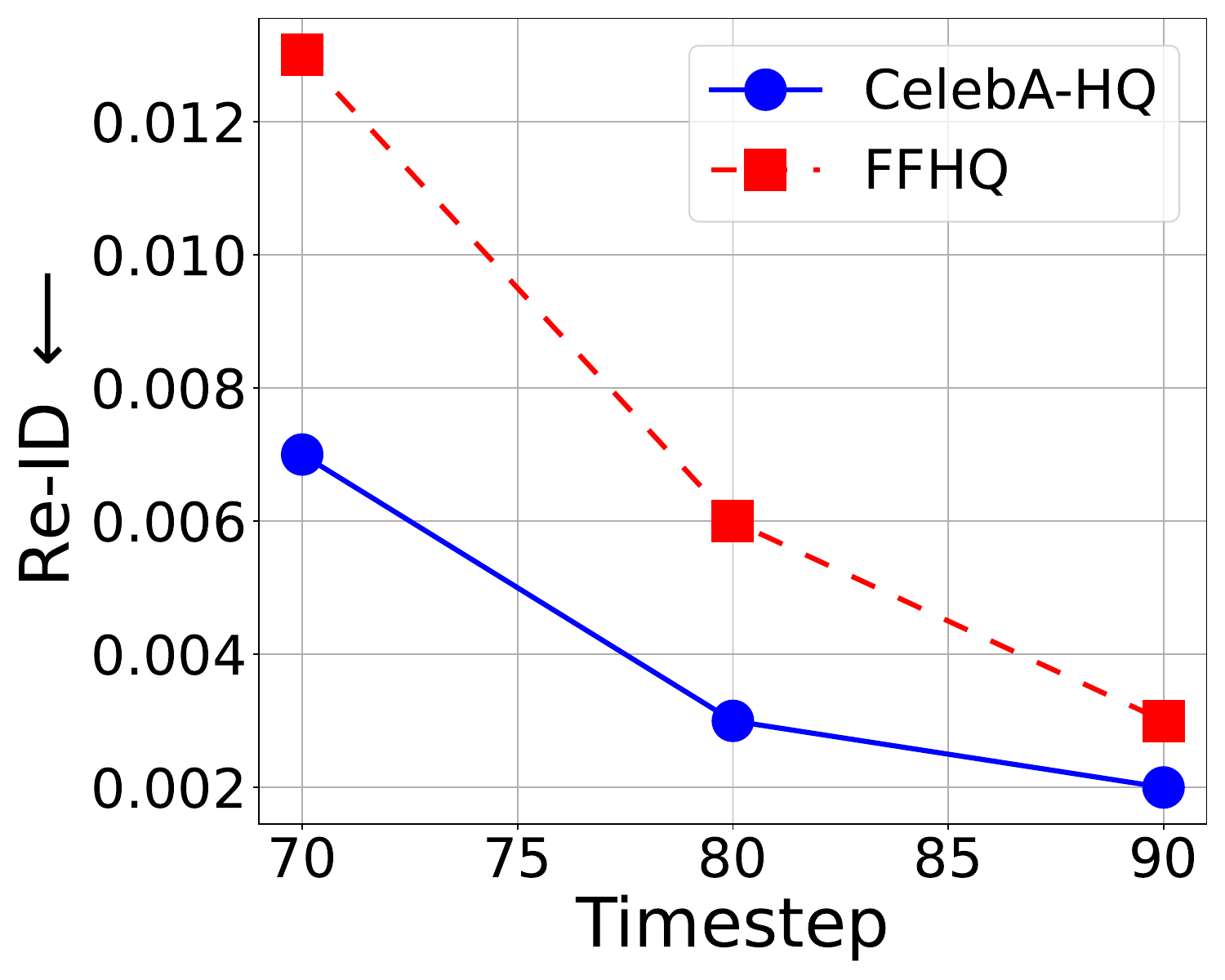}
    \caption{}
    \label{fig:features-b}
  \end{subfigure}
  \\
  \begin{subfigure}{0.49\linewidth}
    \centering
    \includegraphics[width=1.0\linewidth]{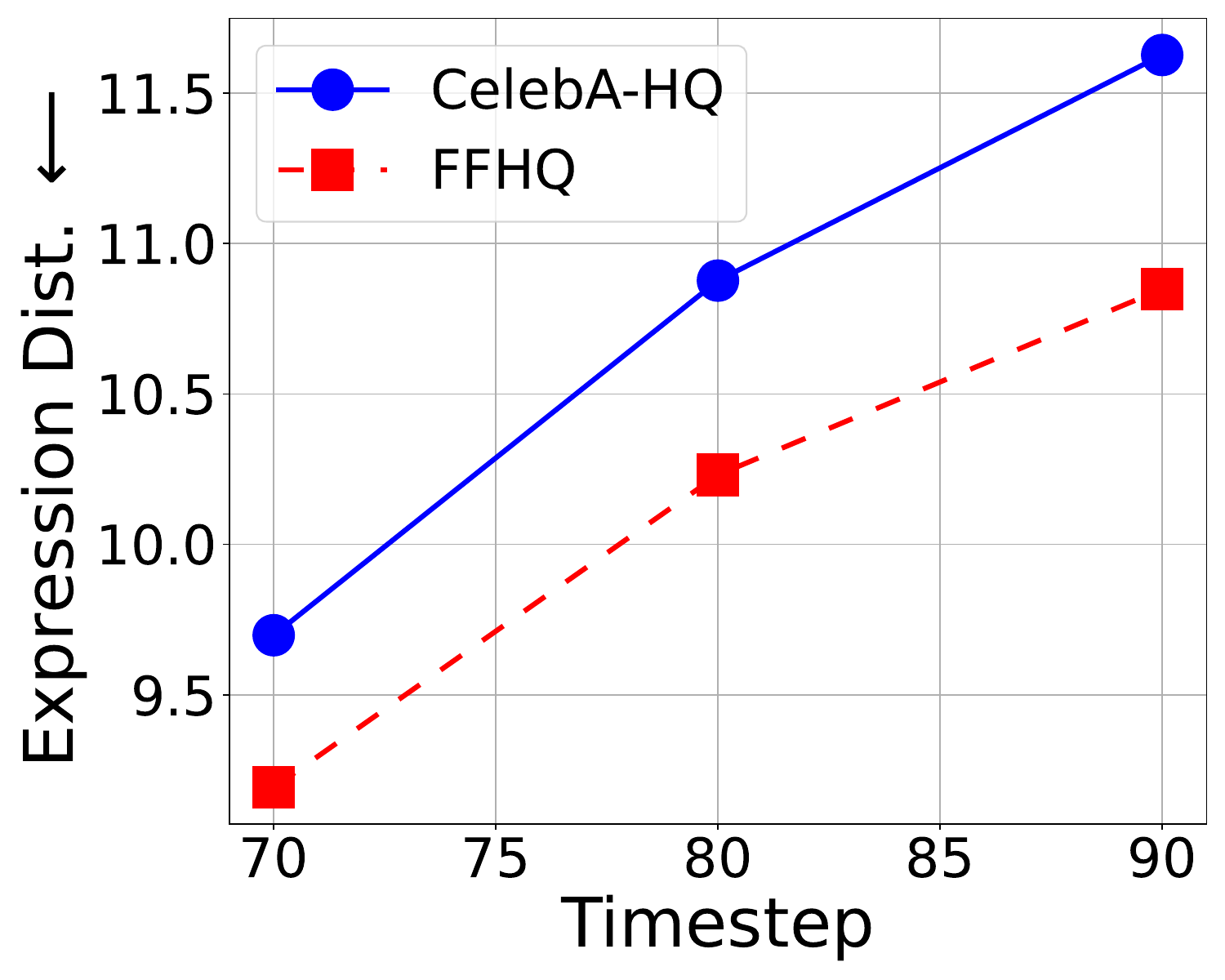}
    \caption{}
    \label{fig:features-c}
  \end{subfigure}
  \hfill
  \begin{subfigure}{0.49\linewidth}
    \centering
    \includegraphics[width=1.0\linewidth]{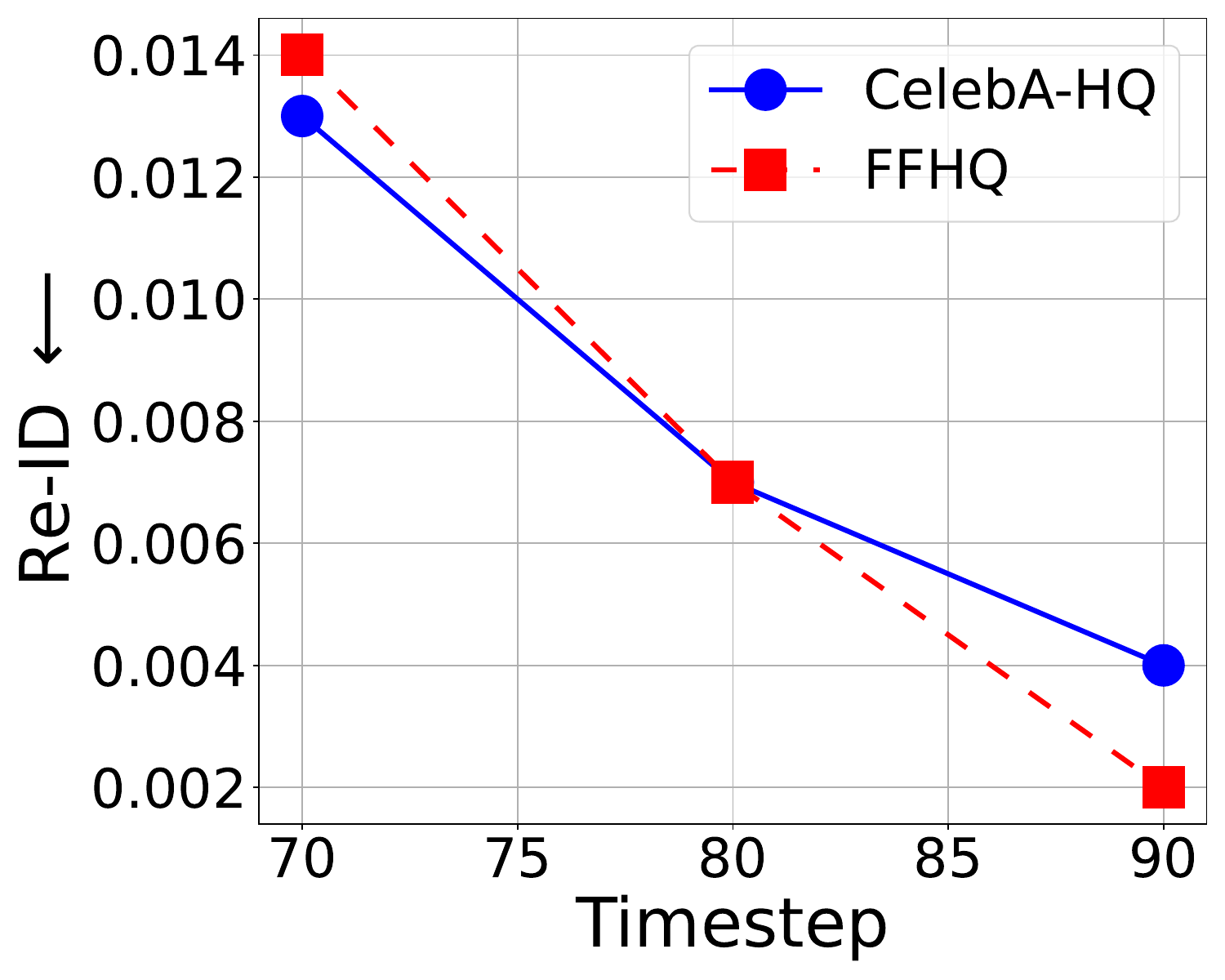}
    \caption{}
    \label{fig:features-d}
  \end{subfigure}
  \\
  \caption{Revealing eyes or mouth earlier in denoising improves gaze and expression retention (\cref{fig:features-a,fig:features-c}) but increases identity recognition risk (\cref{fig:features-b,fig:features-d}, respectively).}
  \label{fig:mask_impact_reid}
\end{figure}

We analyzed the effect of mask application timing during the denoising process. Using 1,000 subjects from CelebA-HQ~\cite{karras2017progressive} and FFHQ~\cite{karras2019style}, we applied eye- or mouth-revealing masks at timesteps 70, 80, and 90. Re-identification rates and gaze distances (for eye masks) or expression distances (for mouth masks) were measured. \Cref{fig:mask_impact_reid} show that earlier mask application (e.g., timestep 70) better preserved original features like gaze or expression but increased re-identification rates. This tradeoff illustrates how mask timing can balance attribute retention with anonymity.

\section{ADDITIONAL PRIVACY-UTILITY TRADE-OFF ANALYSIS}

\begin{figure*}[h]
  \centering

  \begin{subfigure}{1.0\linewidth}
    \centering
    \includegraphics[width=0.24\linewidth]{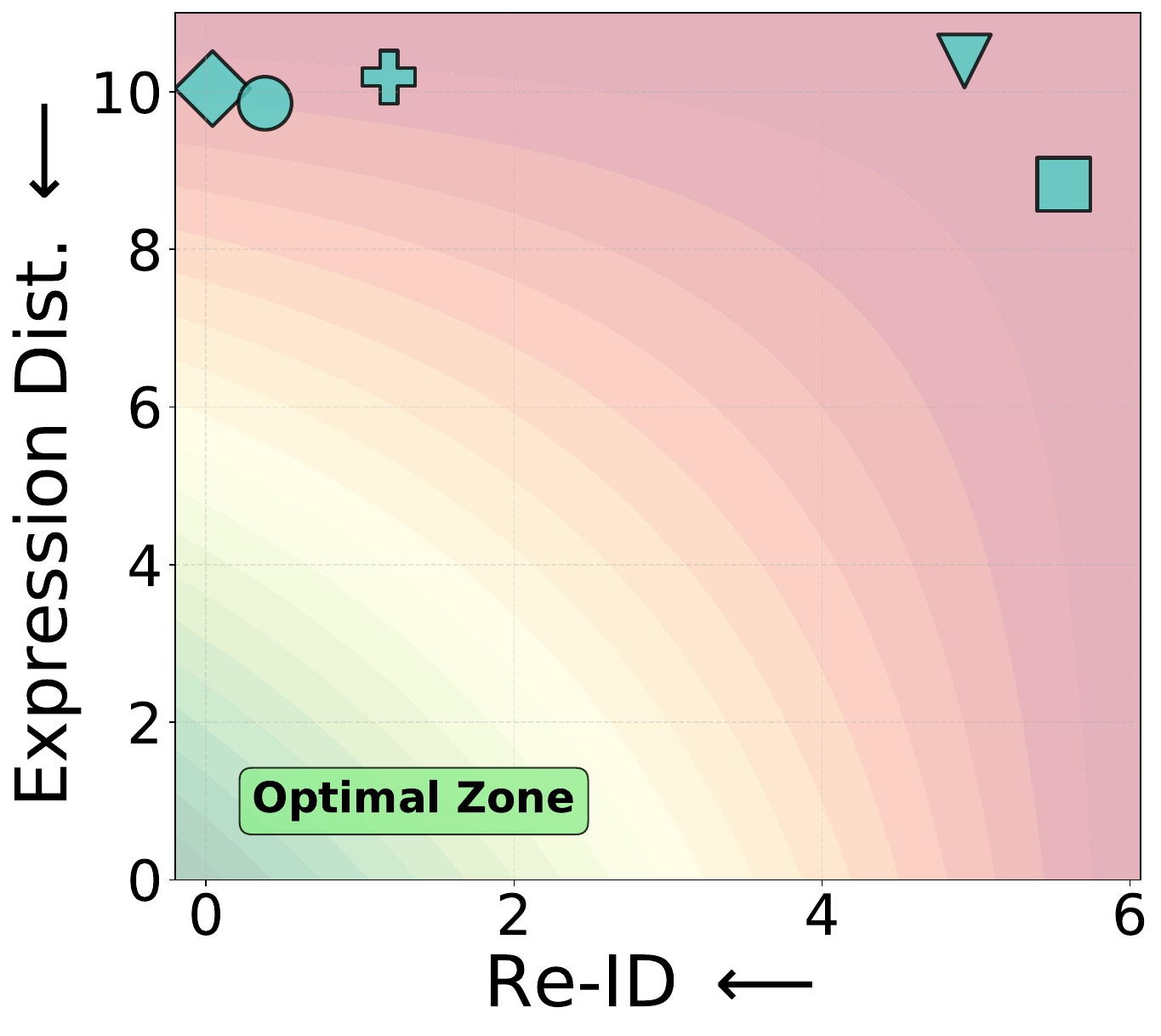}
    \hfill
    \includegraphics[width=0.24\linewidth]{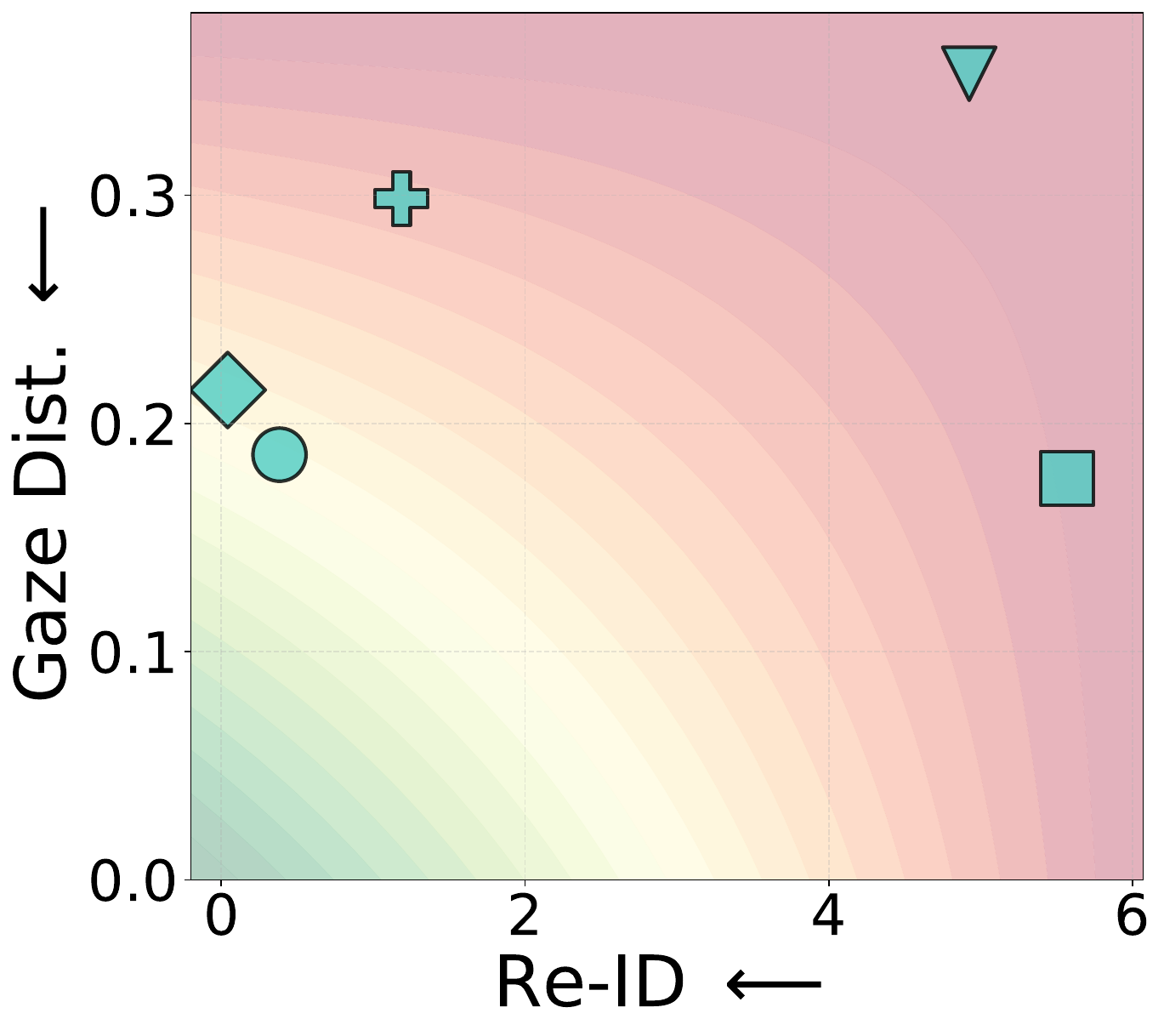}
    \hfill
    \includegraphics[width=0.24\linewidth]{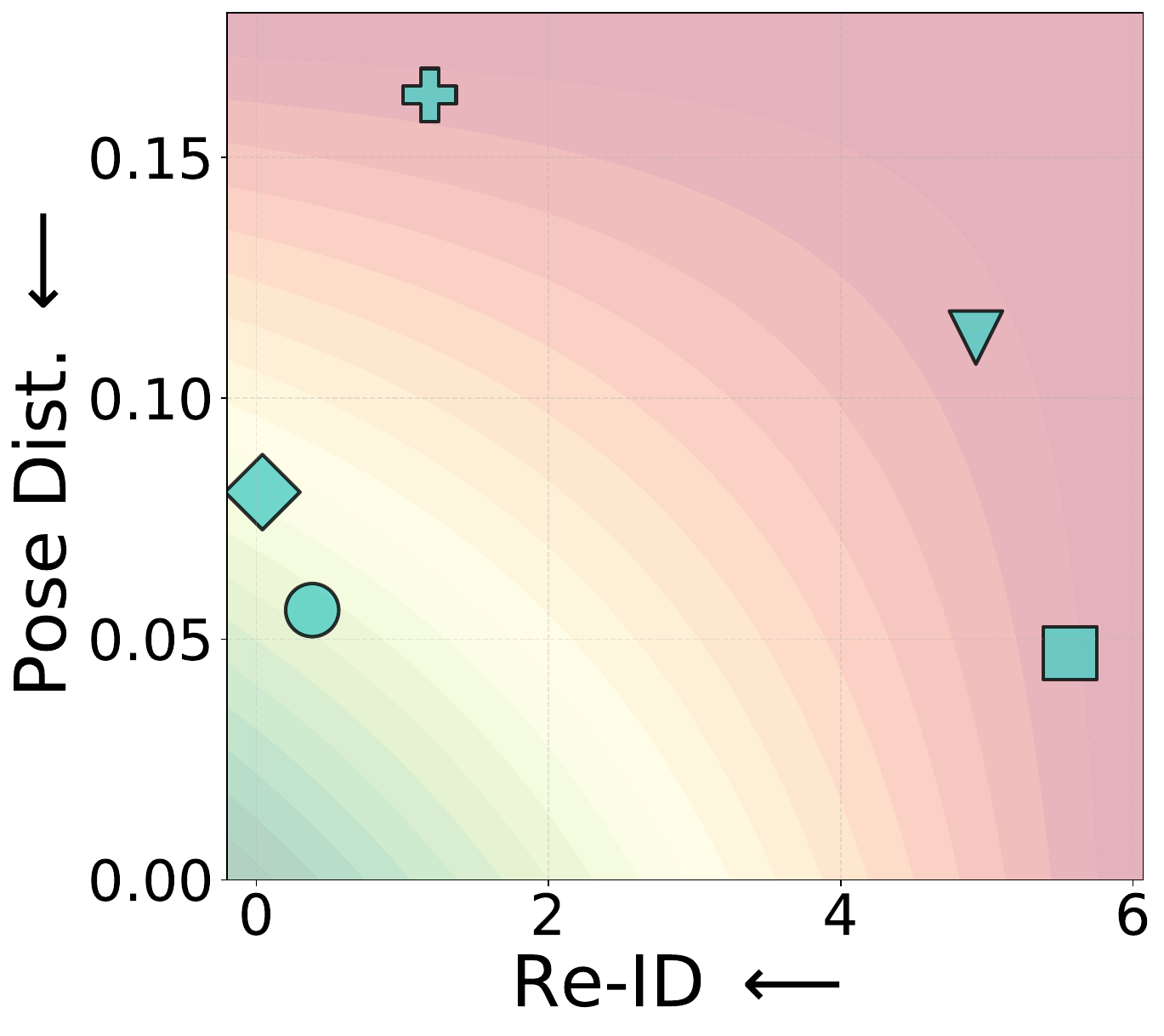}
    \hfill
    \includegraphics[width=0.24\linewidth]{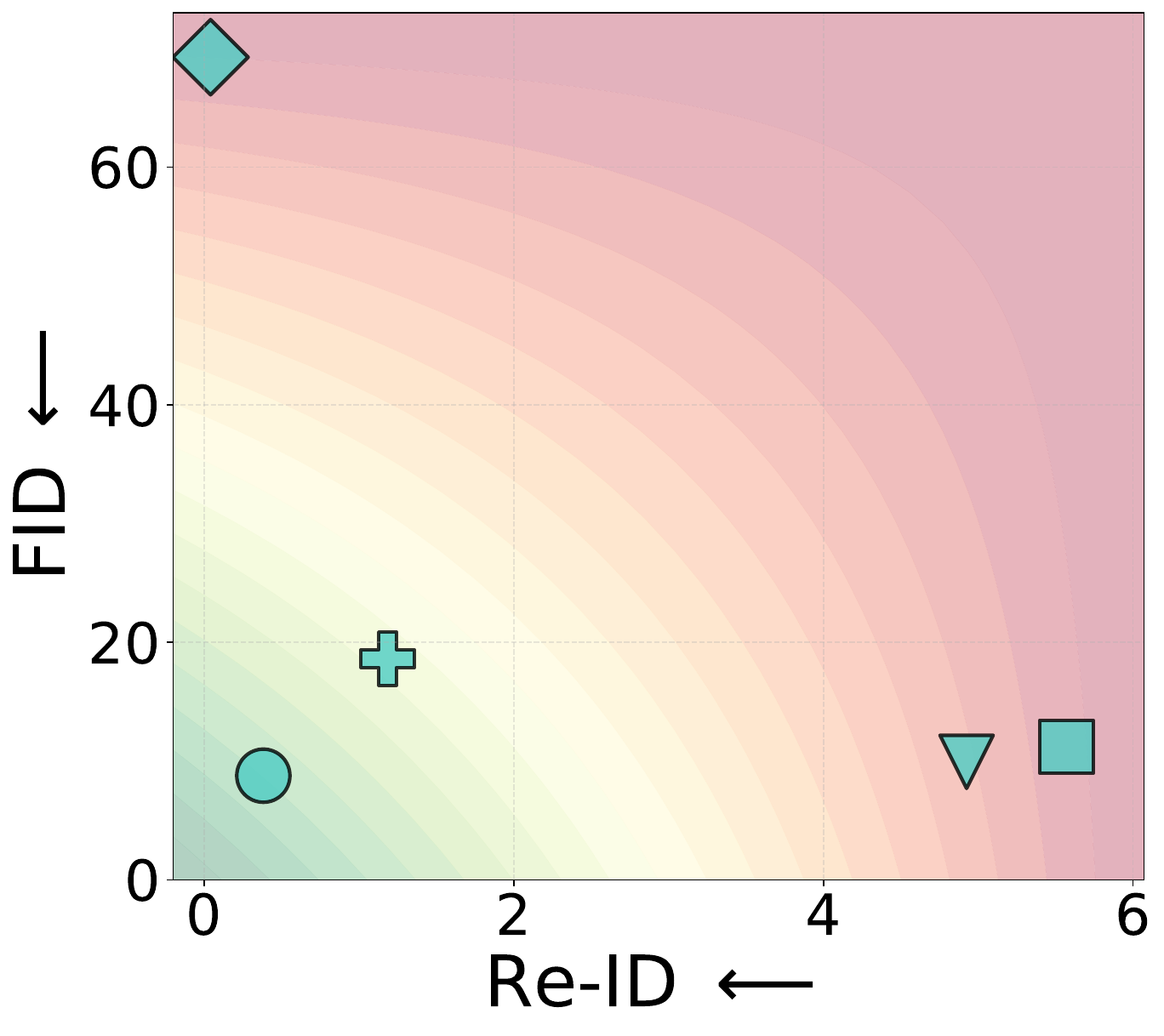}
  \end{subfigure}

  \par\medskip 
  \begin{subfigure}{1.0\linewidth}
    \centering
    \includegraphics[width=0.24\linewidth]{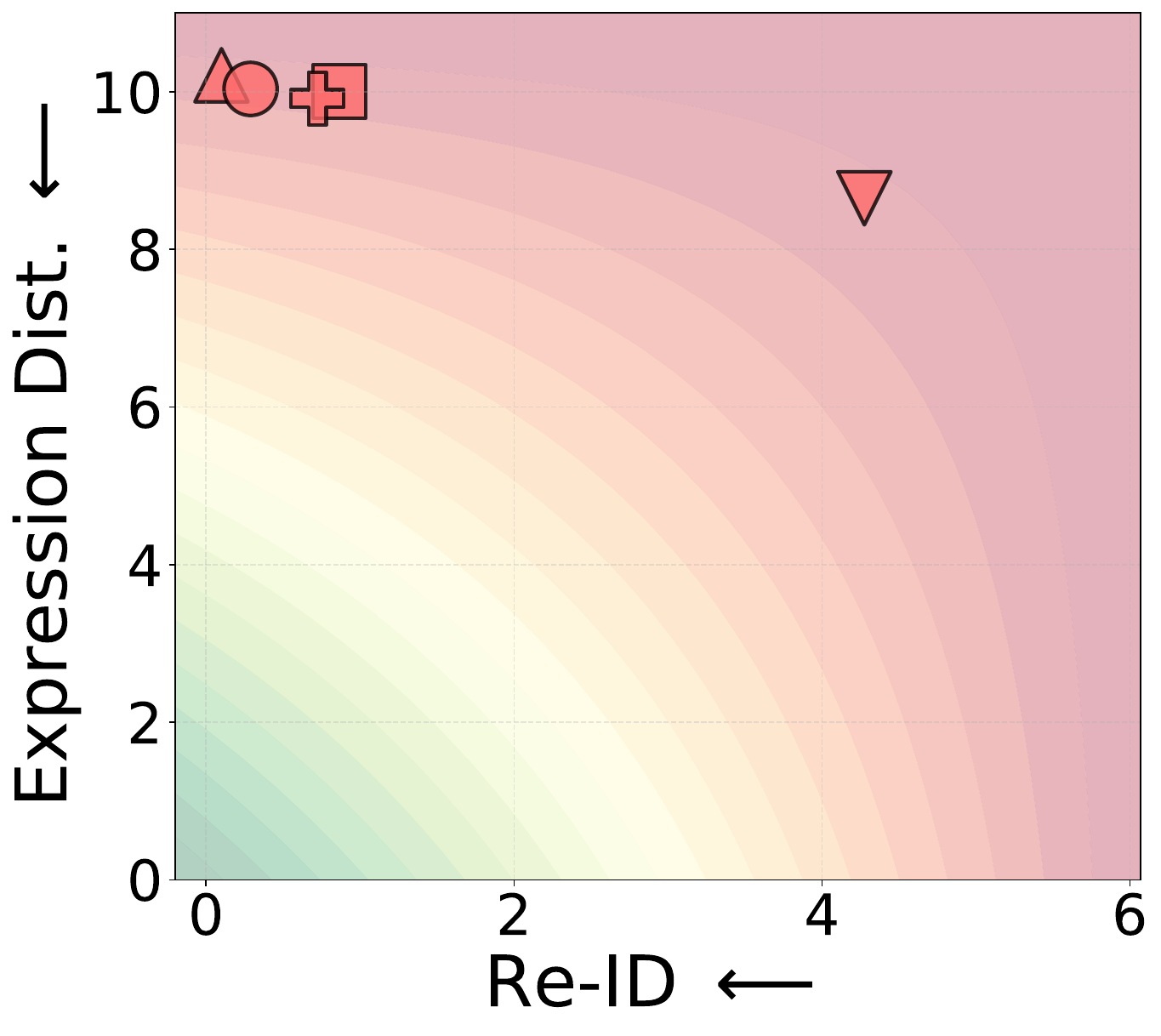}
    \hfill
    \includegraphics[width=0.24\linewidth]{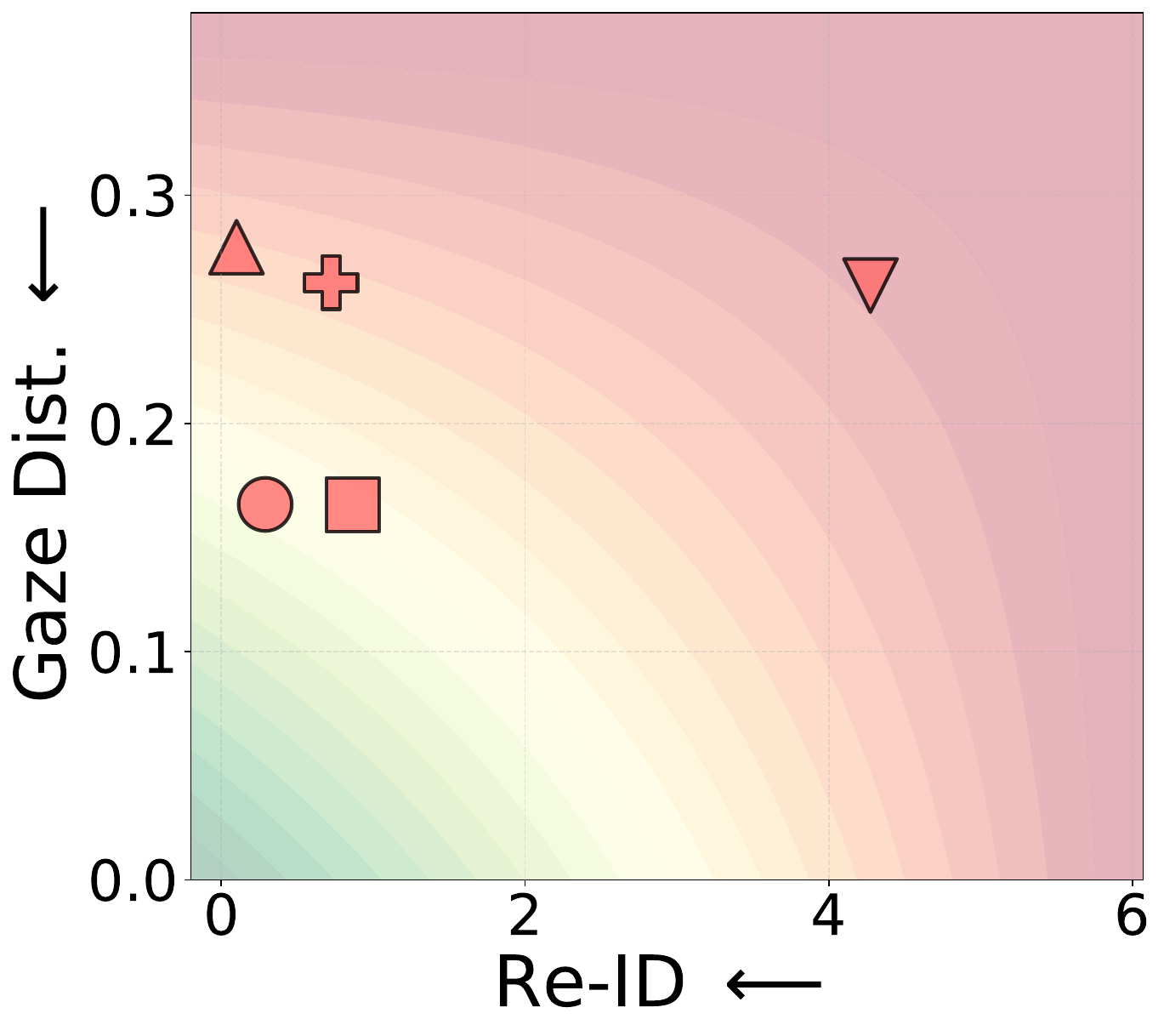}
    \hfill
    \includegraphics[width=0.24\linewidth]{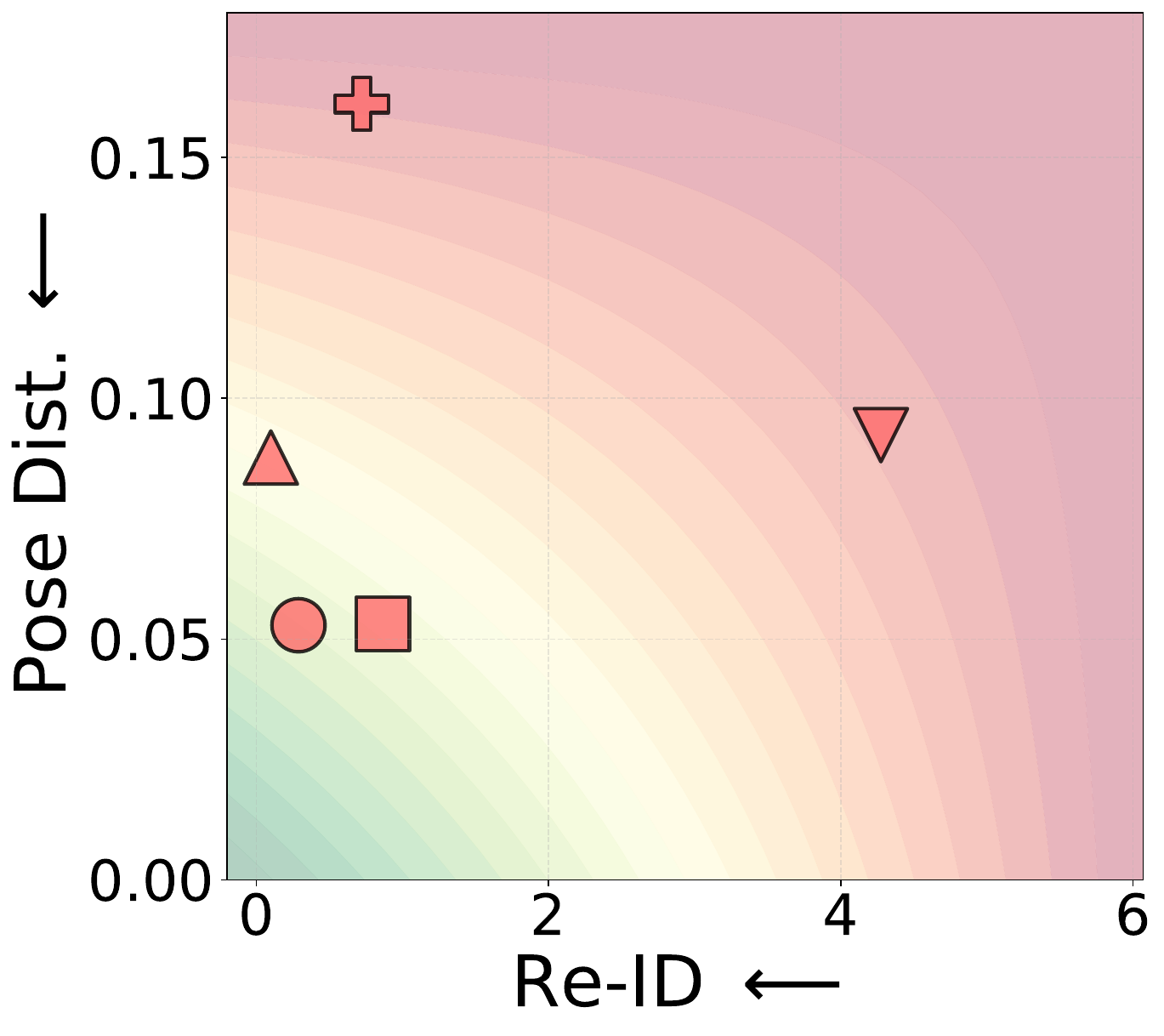}
    \hfill
    \includegraphics[width=0.24\linewidth]{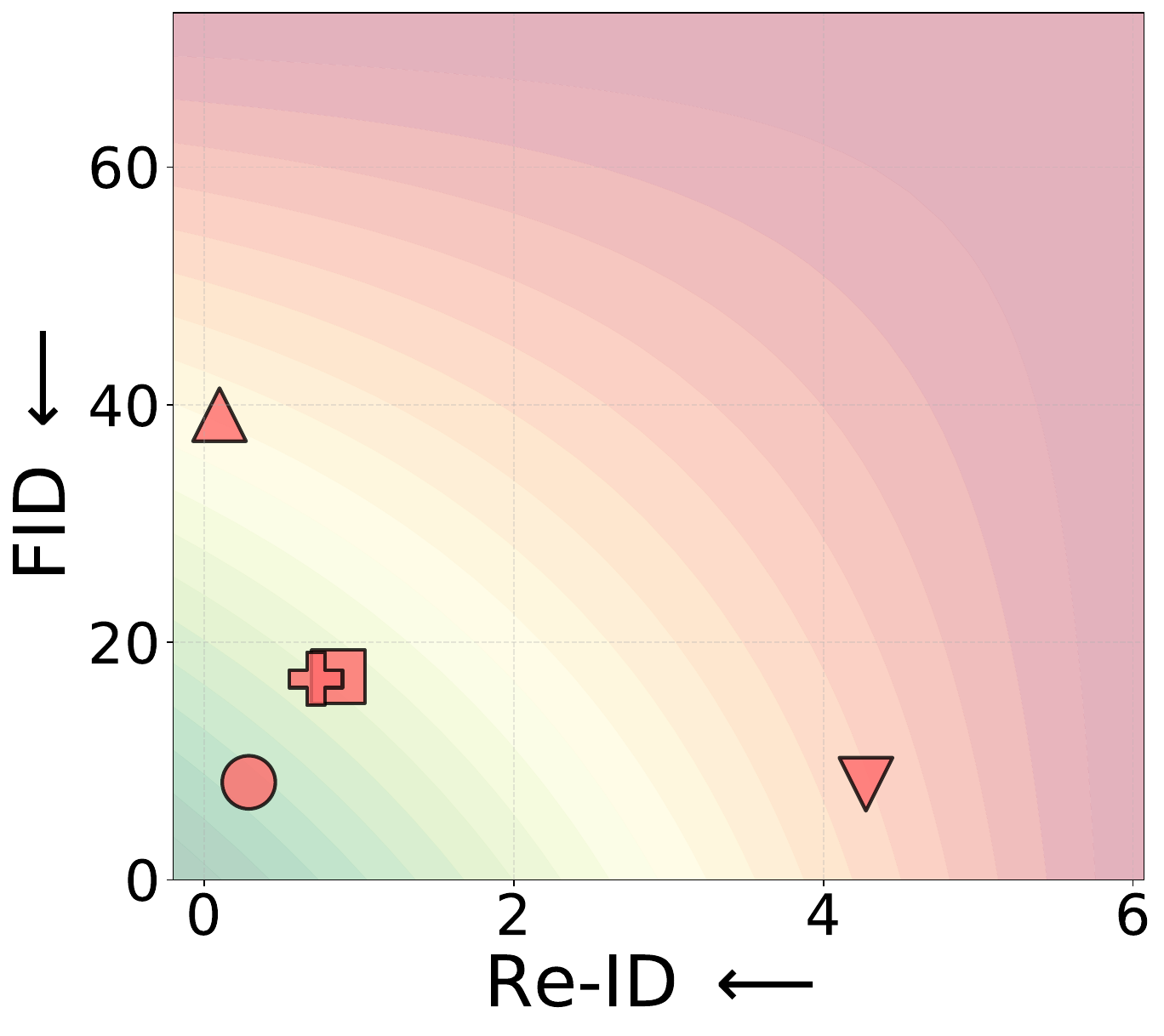}
  \end{subfigure}

  \par\medskip 
  \begin{subfigure}{1.0\linewidth}
    \centering
    \includegraphics[width=.75\linewidth]{./images/plot/trade_offs/legends.pdf}
  \end{subfigure}

  \caption{Privacy-utility trade-off evaluation using the FaceNet~\cite{schroff2015facenet} model and four utility metrics (expression, gaze, pose, FID~\cite{heusel2017gans}) on CelebA-HQ~\cite{karras2017progressive} and FFHQ~\cite{karras2019style} datasets. The green gradient highlights the optimal zone (lower-left) where both privacy protection (low Re-ID) and utility preservation (low metric values) are achieved.}
  \label{fig:trade-offs-facenet}
\end{figure*}

\Cref{fig:trade-offs-facenet} presents privacy-utility trade-offs with re-identification rates calculated using FaceNet~\cite{schroff2015facenet}, complementing \cref{fig:trade-offs-adaface} in the main paper where AdaFace~\cite{kim2022adaface} is used. The figure demonstrates consistent results: our method maintains better balance across all metrics (expression, gaze, pose, and FID~\cite{heusel2017gans}), while baseline methods exhibit trade-offs described in the main text (see \cref{sec:comp_base}).

\section{IMPACT OF FACIAL REGION VISIBILITY ON ANONYMIZATION}

\Cref{tab:visibility_on_anon} provides a quantitative analysis of how keeping different facial regions visible affects anonymization. We evaluated 1,000 subjects from CelebA-HQ~\cite{karras2017progressive} and FFHQ~\cite{karras2019style} using $\lambda_{cfg}=7.0$ and measured re-identification rates with AdaFace~\cite{kim2022adaface} across different masks. The results show that exposing both eyes and nose most significantly compromises anonymity, while keeping only the eyes visible has minimal impact. While this generally confirms that larger visible areas increase recognition, the nose alone has a greater impact than revealing the eyes and mouth combined. Investigating the underlying reasons for this phenomenon could be an interesting direction for future research.

\begin{table}
  \centering
  \resizebox{0.65\columnwidth}{!}{
    \begin{tabular}{@{}lcc@{}}
      \toprule
      & \multicolumn{2}{c}{Re-ID (\%) $\downarrow$} \\
      & CelebA-HQ & FFHQ \\
      \midrule
      Change whole face & 0.50 & 0.51 \\
      Keep eyes & 1.70 & 1.01 \\
      Keep mouth & 2.30 & 1.01 \\
      Keep eyes and mouth & 4.81 & 2.73 \\
      Keep nose & 10.62 & 6.67 \\
      Keep nose and mouth & 29.19 & 22.26 \\
      Keep eyes and nose & 30.56 & 23.68 \\
      \bottomrule
    \end{tabular}
  }
  \caption{Impact of facial region visibility on identity retrieval rates.}
  \label{tab:visibility_on_anon}
\end{table}

\section{IDENTITY REVERSAL ATTACKS}

\begin{figure}[h]
  \centering
  \resizebox{.85\columnwidth}{!}{
    \begin{tabular}{
        *{3}{>{\centering\arraybackslash}m{\dimexpr.33\linewidth-2\tabcolsep-.33\arrayrulewidth}}
      }
      \multicolumn{1}{c}{Original} & \multicolumn{1}{c|}{Anonymized} & \multicolumn{1}{c}{Attacked} \\ 
      \multicolumn{1}{c}{\includegraphics[width={\dimexpr.33\linewidth}]{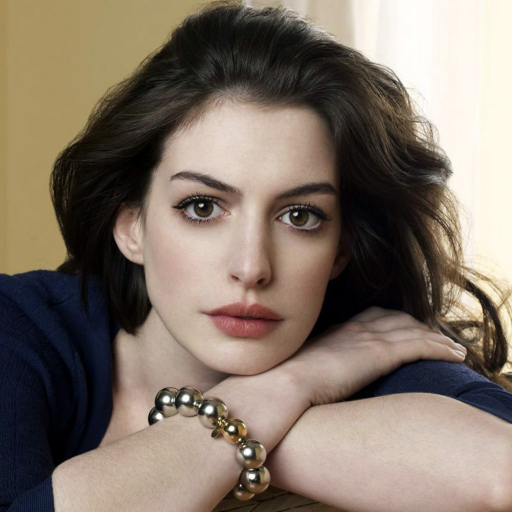}} & \multicolumn{1}{c|}{\includegraphics[width={\dimexpr.33\linewidth}]{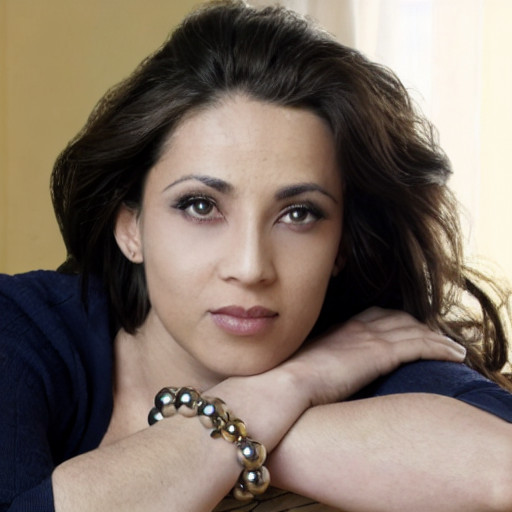}} & \multicolumn{1}{c}{\includegraphics[width={\dimexpr.33\linewidth}]{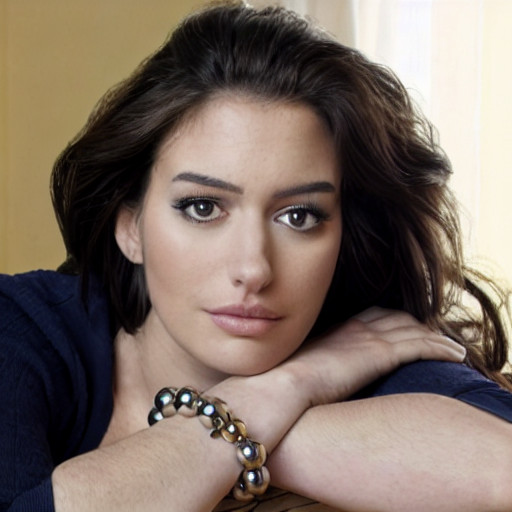}} \\
      \multicolumn{1}{c}{\includegraphics[width={\dimexpr.33\linewidth}]{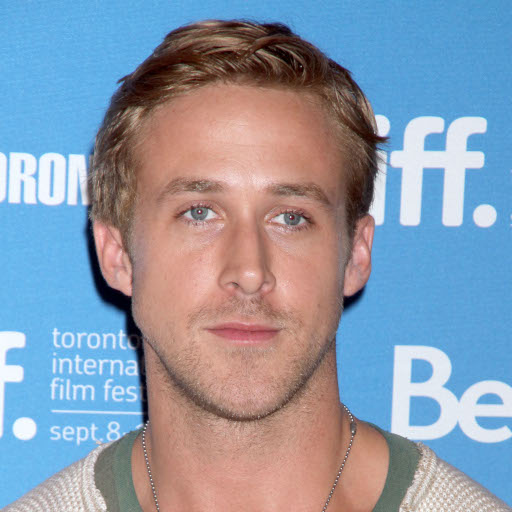}} & \multicolumn{1}{c|}{\includegraphics[width={\dimexpr.33\linewidth}]{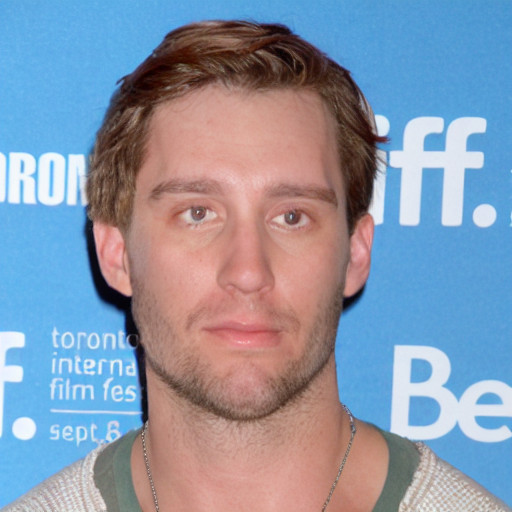}} & \multicolumn{1}{c}{\includegraphics[width={\dimexpr.33\linewidth}]{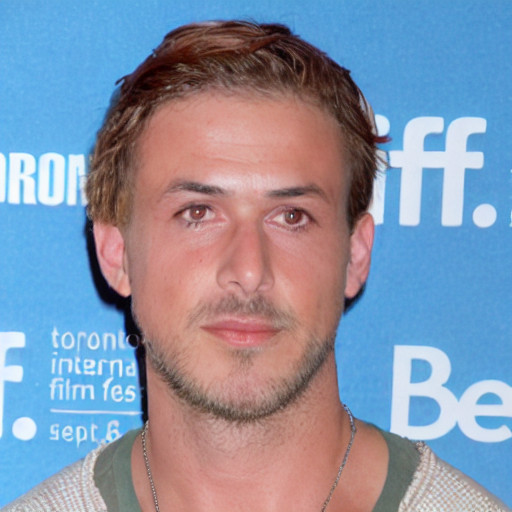 }}  \\
    \end{tabular}
  }
  \caption{Demonstration of our method's reliability against identity recovery attacks.}
  \label{fig:de_anon}
\end{figure}

While one might assume that reapplying a negative embedding to our anonymized images could reverse the anonymization process, our method remains robust due to the influence of multiple factors. As shown in \cref{eq:cfg}, the anonymization outcome depends not only on the embedding but also on the guidance scale and null condition---parameters that are not accessible to potential attackers. \Cref{fig:de_anon} illustrates that simply reapplying a negative embedding fails to restore the source identity. Moreover, our quantitative results, presented in \cref{tab:deanon}, show that re-identification rates remain low after this procedure is applied to anonymized images where entire faces have been altered. These rates are among the lowest compared to other baselines reported in \cref{tab:comp_quan_4k}, indicating that such attacks are ineffective at recovering the original identity.

\begin{table}[h]
  \centering
	\resizebox{.75\linewidth}{!}{
    \begin{tabular}{@{}l cc@{}}
      \toprule
      & \multicolumn{2}{c}{Re-ID (\%) $\downarrow$} \\
      \cmidrule(lr){2-3}
      & CelebA-HQ & FFHQ \\
      \midrule
      Anonymized (change whole face) & 0.08 & 0.11 \\
      Attacked & 0.31 & 0.35 \\
      \bottomrule
    \end{tabular}
  }
  \caption{Re-identification rates after reapplying negative embeddings to anonymized images.}
  \label{tab:deanon}
\end{table}

\section{ADDITIONAL SEGMENTATION MASK RESULTS}

\begin{figure*}
  \footnotesize
  \centering
  \begin{tabular}{*{8}{>{\centering\arraybackslash}m{\dimexpr.125\linewidth-2\tabcolsep}}}
    Original & Change whole face & Keep eyes & Change eyes & Keep nose & Change nose & Keep mouth & Change mouth \\
    \multicolumn{8}{@{}c@{}}{\includegraphics[width=\linewidth]{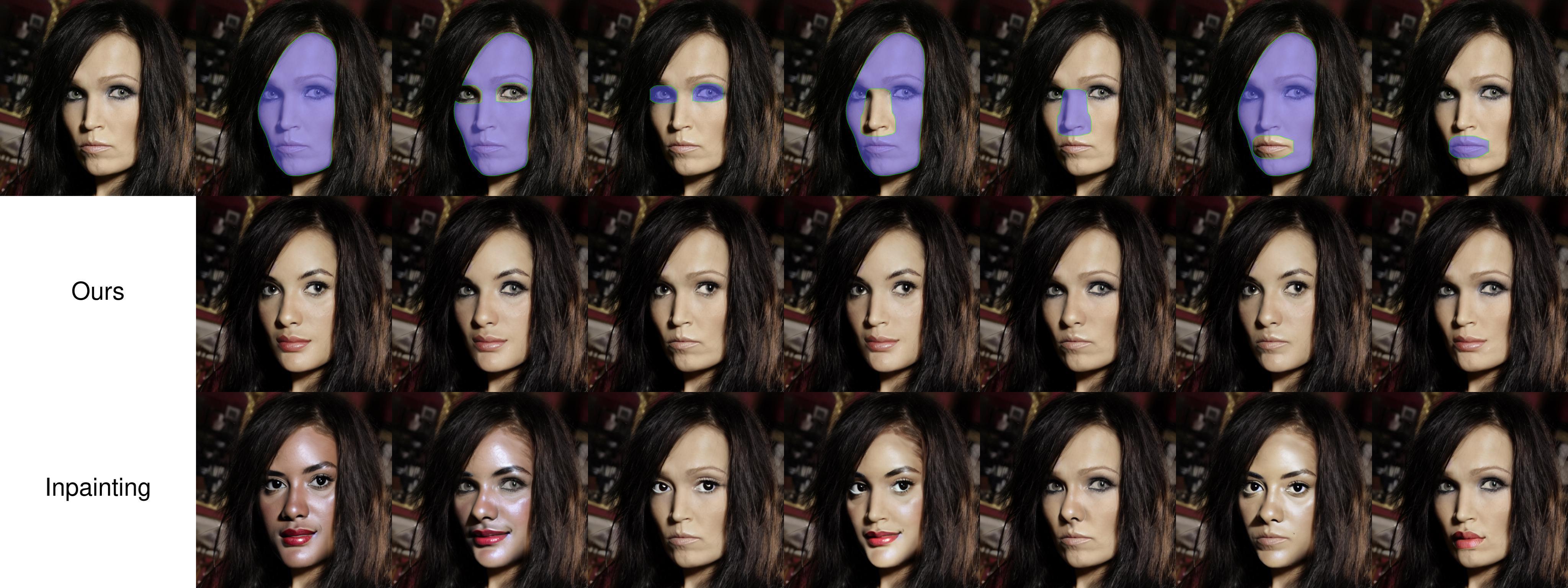}} \\
    \multicolumn{8}{@{}c@{}}{\includegraphics[width=\linewidth]{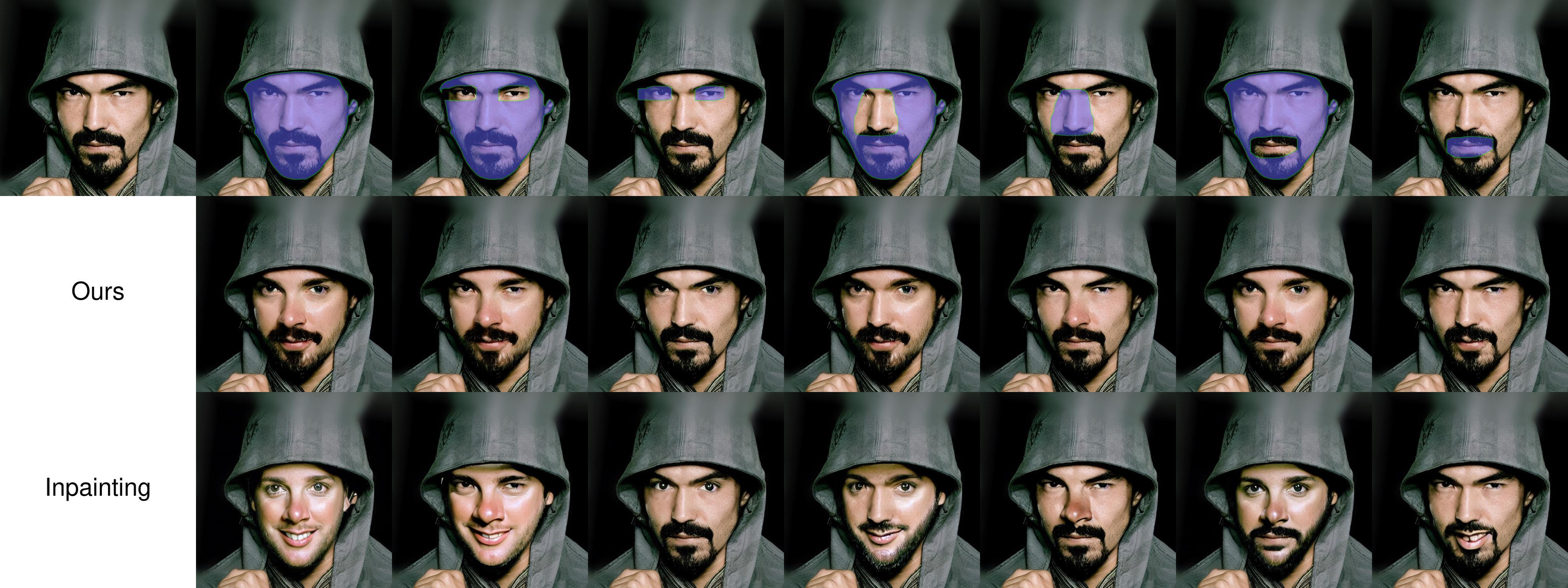}} \\
    \multicolumn{8}{@{}c@{}}{\includegraphics[width=\linewidth]{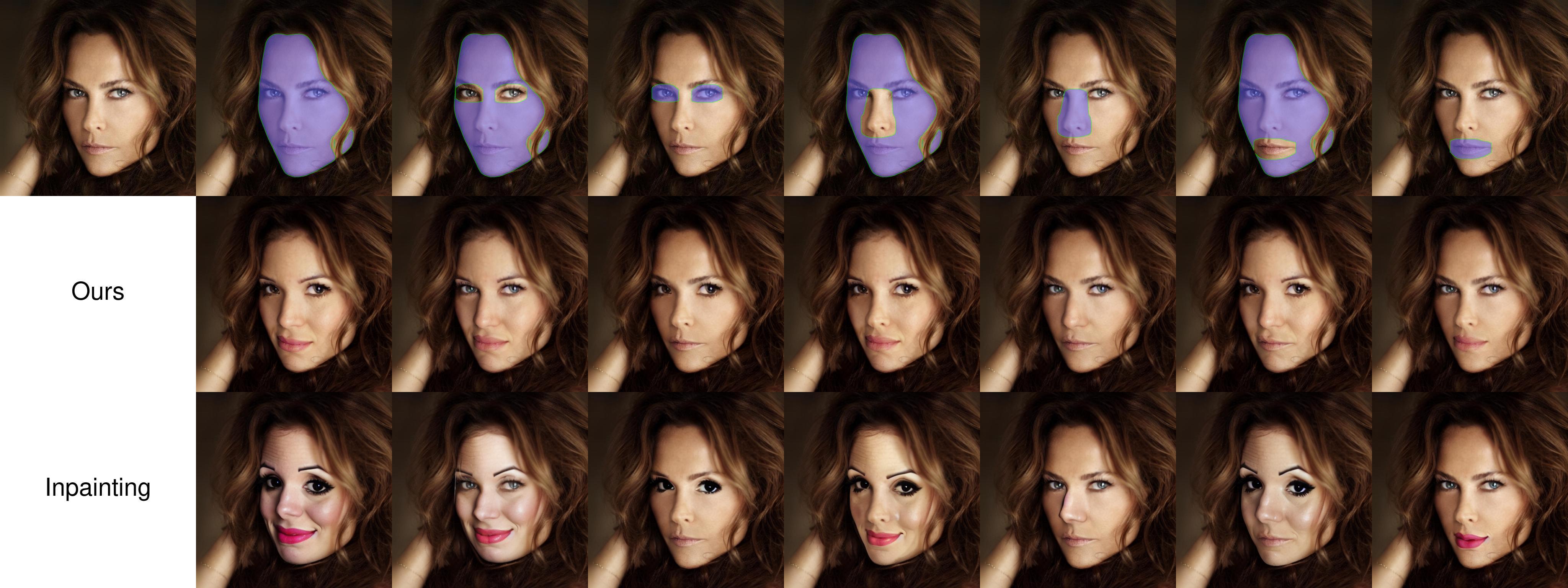}} \\
  \end{tabular}
  \caption{Comparison of facial region anonymization using segmentation masks on CelebA-HQ~\cite{karras2017progressive} images, including a comparison with the Stable Diffusion Inpainting model~\cite{rombach2022high}.}
  \label{fig:mask_cele_plus_0}
\end{figure*}

\begin{figure*}
  \footnotesize
  \centering
  \begin{tabular}{*{8}{>{\centering\arraybackslash}m{\dimexpr.125\linewidth-2\tabcolsep}}}
    Original & Change whole face & Keep eyes & Change eyes & Keep nose & Change nose & Keep mouth & Change mouth \\
    \multicolumn{8}{@{}c@{}}{\includegraphics[width=\linewidth]{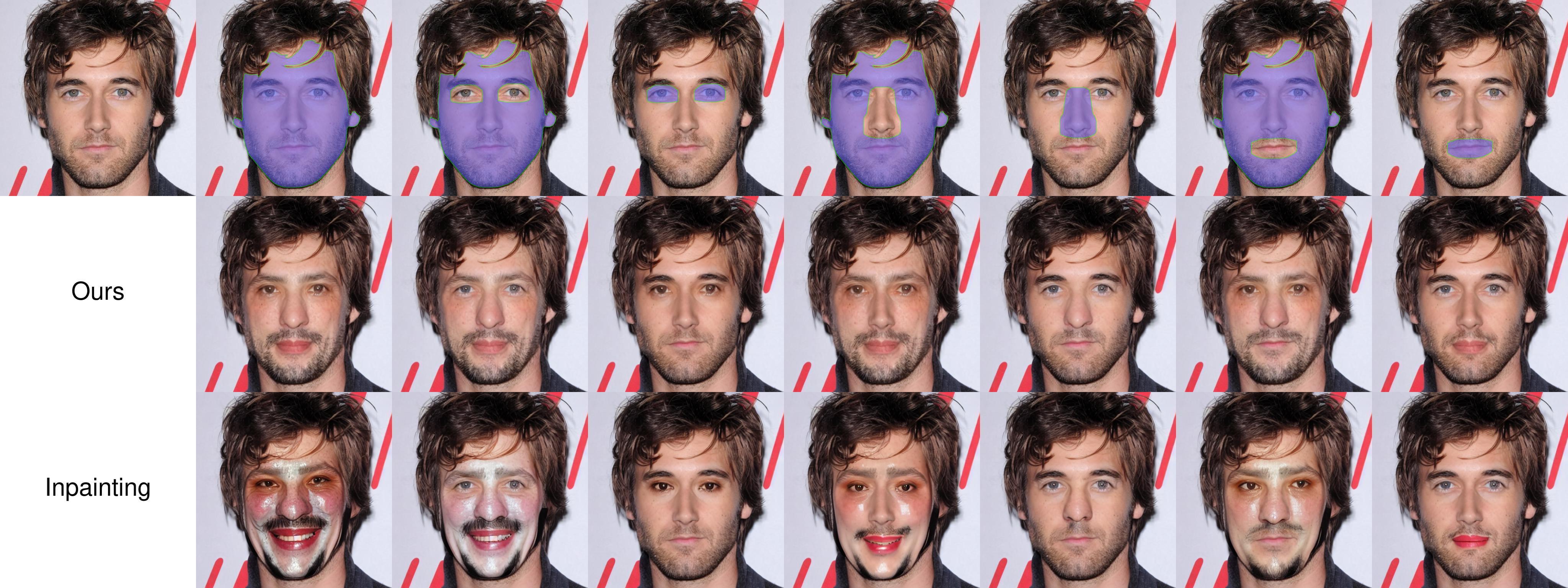}} \\
    \multicolumn{8}{@{}c@{}}{\includegraphics[width=\linewidth]{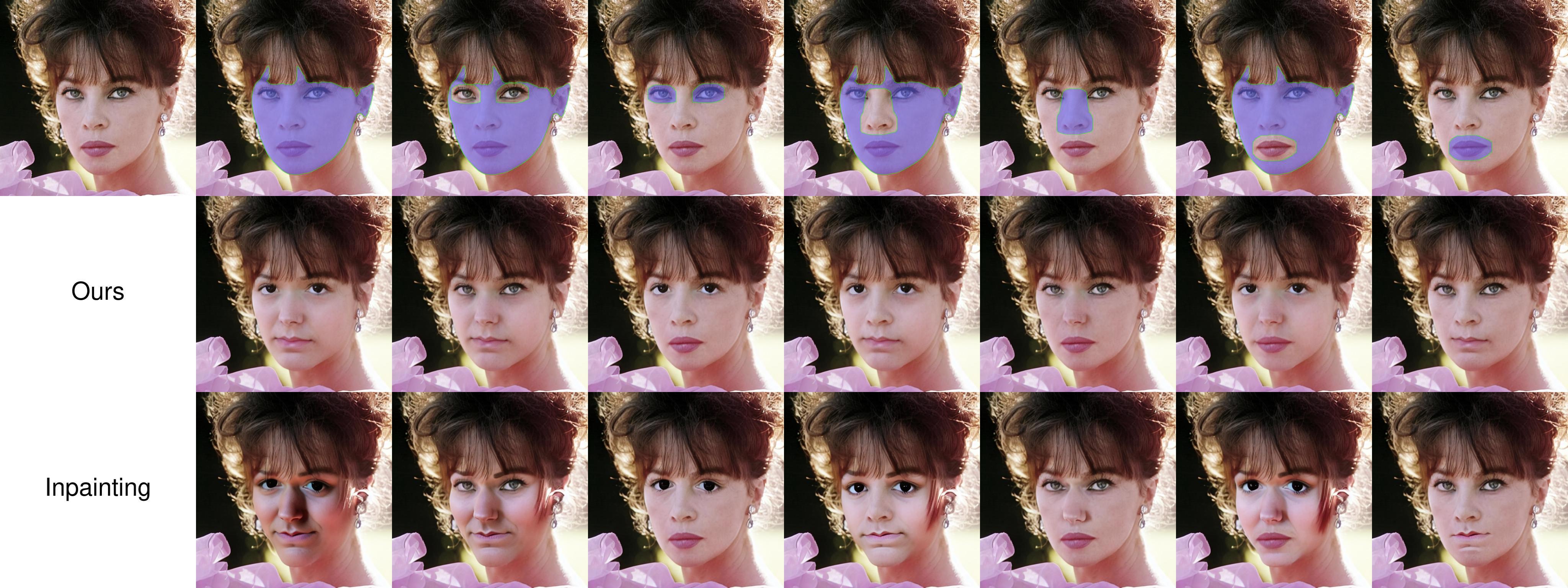}} \\
    \multicolumn{8}{@{}c@{}}{\includegraphics[width=\linewidth]{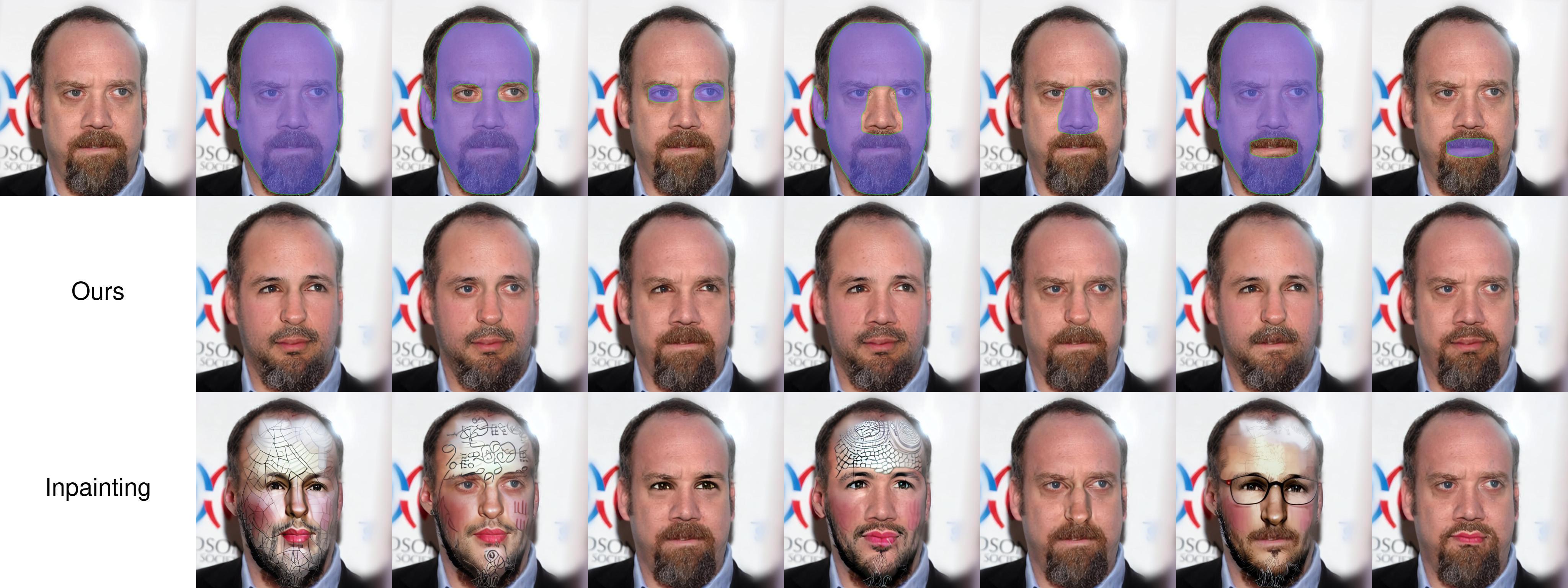}} \\
  \end{tabular}
  \caption{Comparison of facial region anonymization using segmentation masks on CelebA-HQ~\cite{karras2017progressive} images, including a comparison with the Stable Diffusion Inpainting model~\cite{rombach2022high}.}
  \label{fig:mask_cele_plus_1}
\end{figure*}

\begin{figure*}
  \footnotesize
  \centering
  \begin{tabular}{*{8}{>{\centering\arraybackslash}m{\dimexpr.125\linewidth-2\tabcolsep}}}
    Original & Change whole face & Keep eyes & Change eyes & Keep nose & Change nose & Keep mouth & Change mouth \\
    \multicolumn{8}{@{}c@{}}{\includegraphics[width=\linewidth]{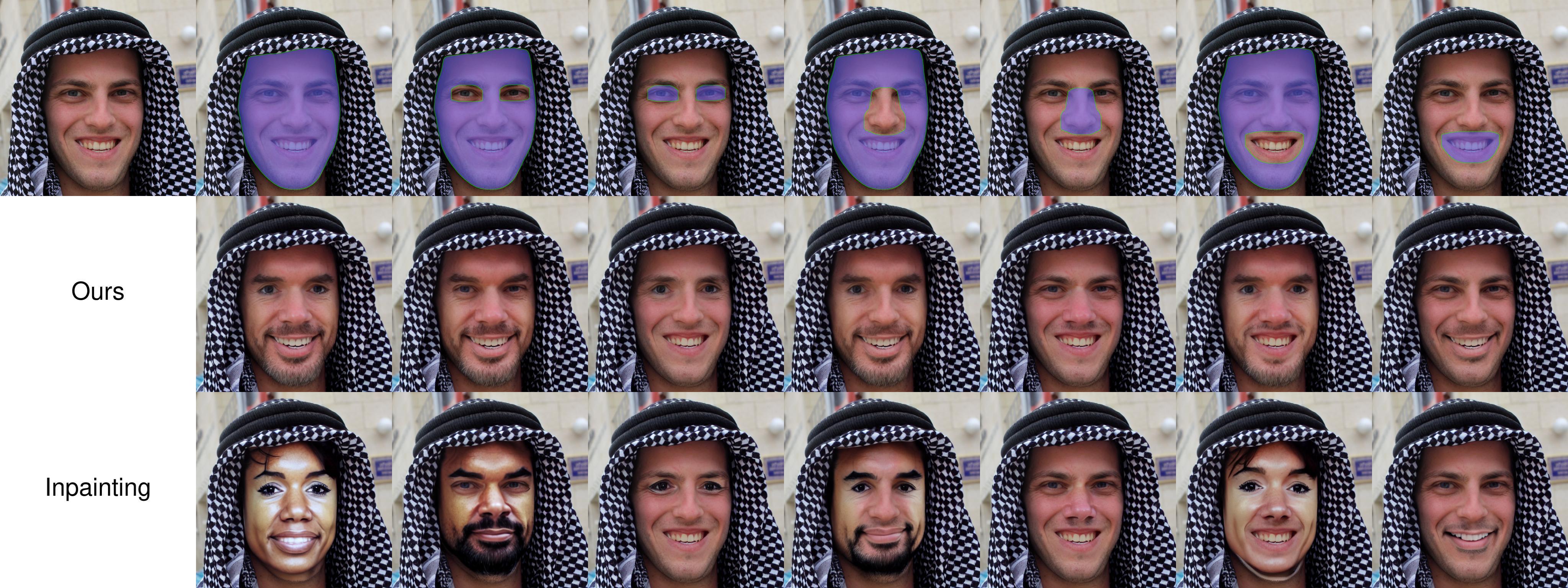}} \\
    \multicolumn{8}{@{}c@{}}{\includegraphics[width=\linewidth]{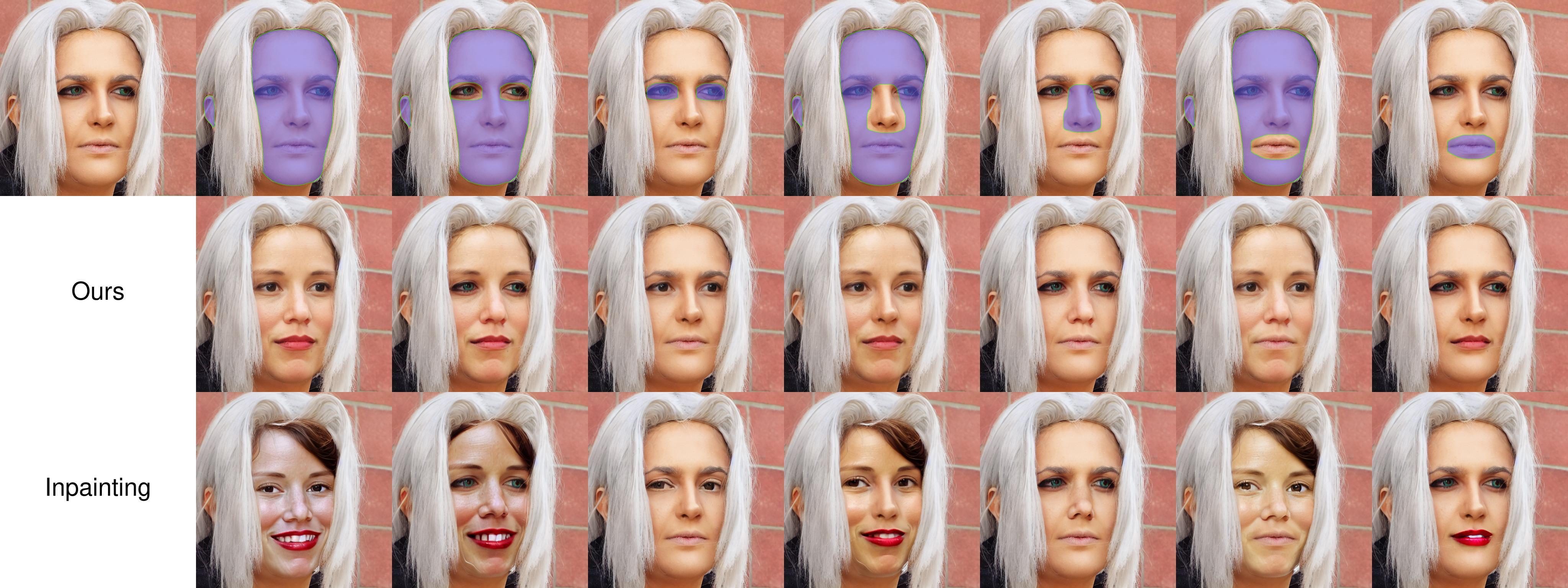}} \\
    \multicolumn{8}{@{}c@{}}{\includegraphics[width=\linewidth]{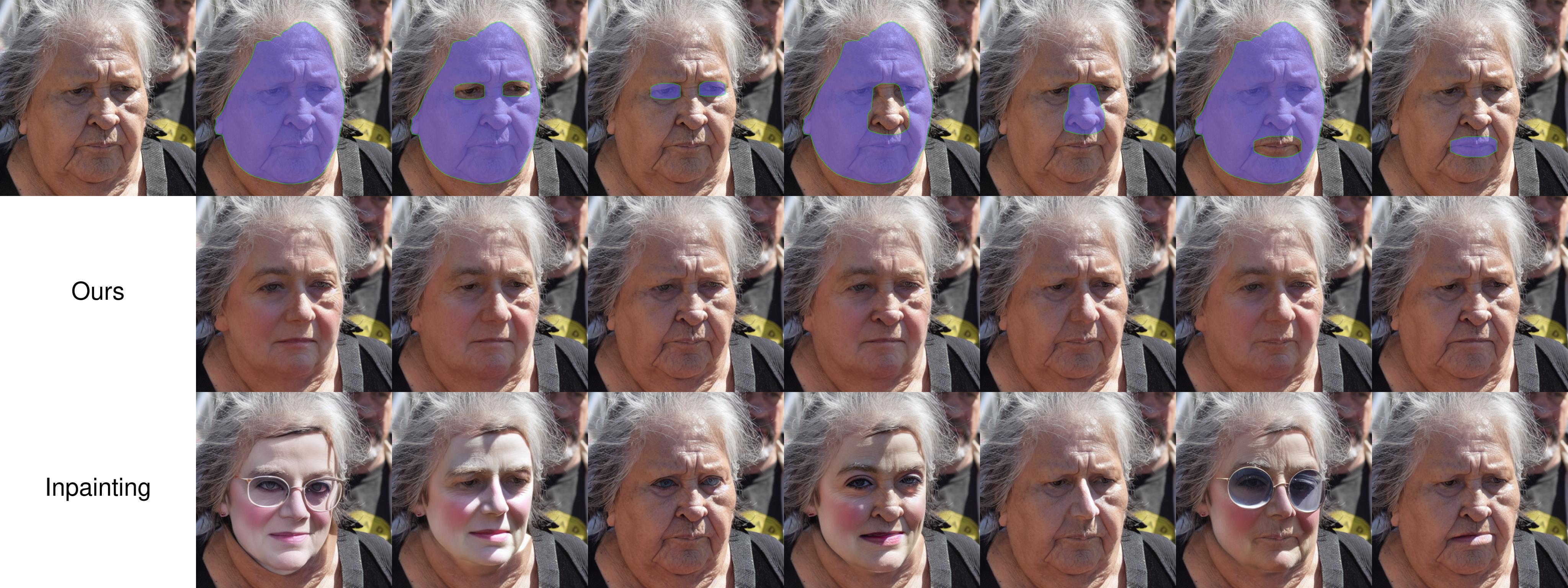}} \\
  \end{tabular}
  \caption{Comparison of facial region anonymization using segmentation masks on FFHQ~\cite{karras2019style} images, including a comparison with the Stable Diffusion Inpainting model~\cite{rombach2022high}.}
  \label{fig:mask_ffhq_plus_0}
\end{figure*}

\begin{figure*}
  \footnotesize
  \centering
  \begin{tabular}{*{8}{>{\centering\arraybackslash}m{\dimexpr.125\linewidth-2\tabcolsep}}}
    Original & Change whole face & Keep eyes & Change eyes & Keep nose & Change nose & Keep mouth & Change mouth \\
    \multicolumn{8}{@{}c@{}}{\includegraphics[width=\linewidth]{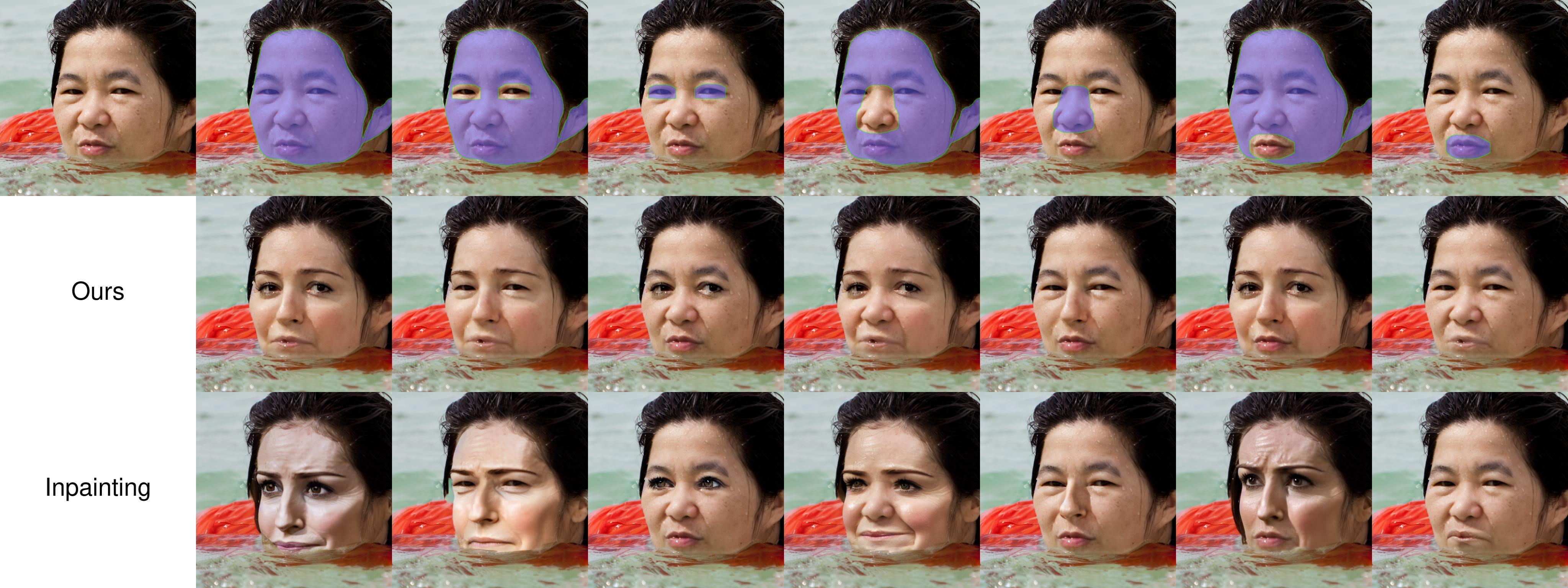}} \\
    \multicolumn{8}{@{}c@{}}{\includegraphics[width=\linewidth]{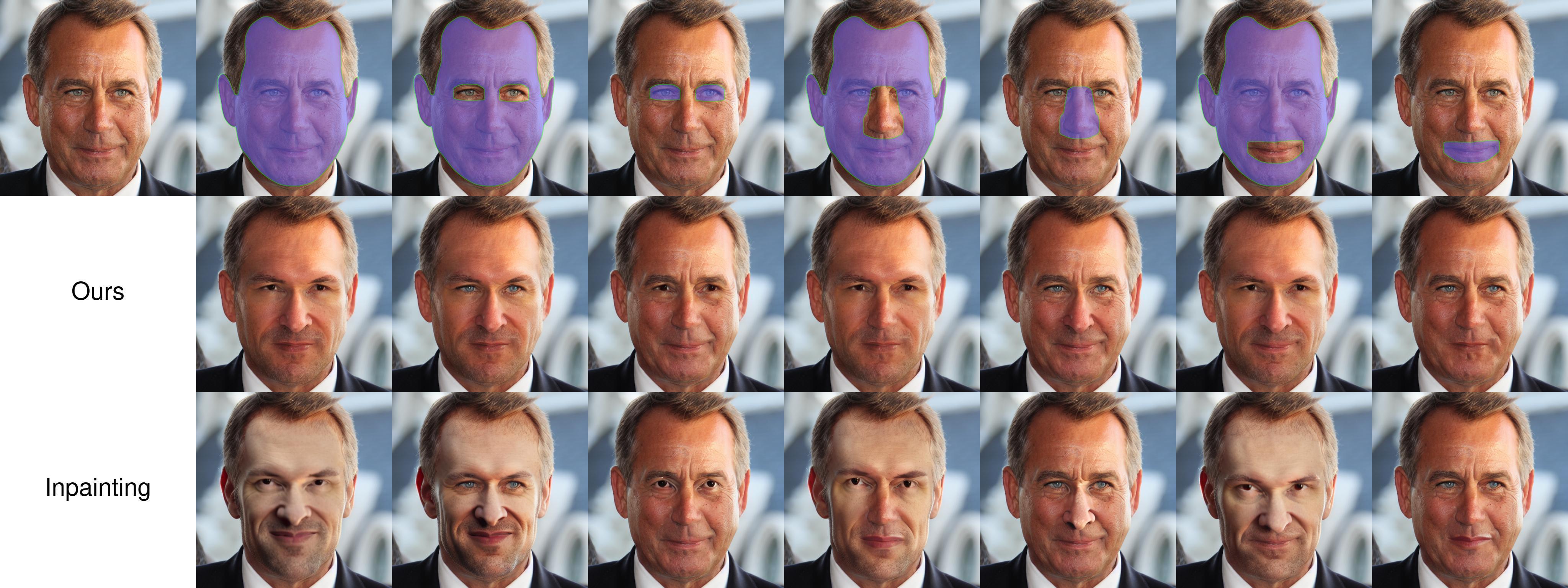}} \\
    \multicolumn{8}{@{}c@{}}{\includegraphics[width=\linewidth]{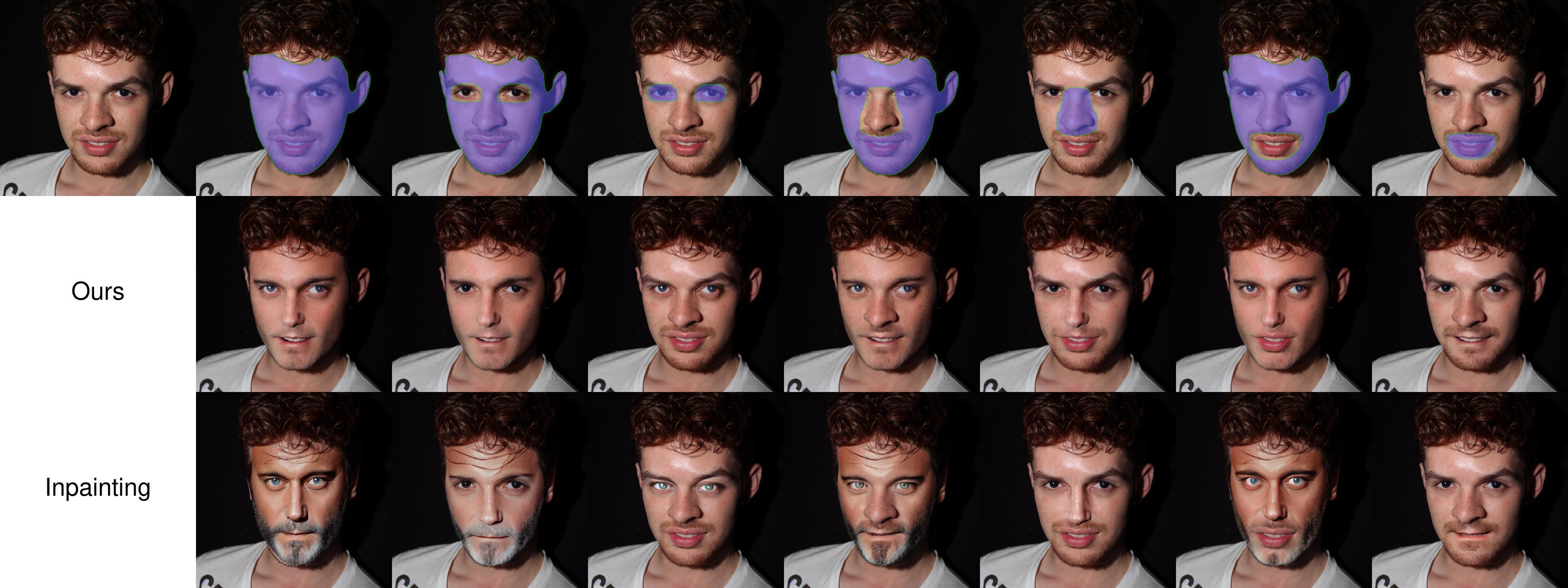}} \\
  \end{tabular}
  \caption{Comparison of facial region anonymization using segmentation masks on FFHQ~\cite{karras2019style} images, including a comparison with the Stable Diffusion Inpainting model~\cite{rombach2022high}.}
  \label{fig:mask_ffhq_plus_1}
\end{figure*}

\Cref{fig:mask_cele_plus_0,fig:mask_cele_plus_1,fig:mask_ffhq_plus_0,fig:mask_ffhq_plus_1} showcase additional results of applying segmentation masks to anonymize specific facial regions using images from the CelebA-HQ~\cite{karras2017progressive} and FFHQ~\cite{karras2019style} datasets. By presenting examples from both datasets, we highlight how segmentation masks enable precise control over which facial attributes, such as the eyes, nose, or mouth, are preserved or hidden.

Additionally, we compare our method with the Stable Diffusion Inpainting model~\cite{rombach2022high} to highlight the strengths of our framework. The results show that our method consistently preserves the same anonymized appearance across different segmentation masks. This ensures that the appearance remains stable when different facial regions are revealed or hidden. In contrast, the inpainting model struggles to maintain a consistent anonymized appearance when segmentation masks change. These qualitative comparisons underscore the robustness of our approach in producing stable and coherent anonymized appearance.

\section{ADDITIONAL QUALITATIVE COMPARISONS}

\begin{figure*}
  \centering
  \begin{tabularx}{\linewidth}{@{}X@{}X@{}X@{}X@{}X@{}X@{}}
    \multicolumn{6}{@{}c@{}}{\includegraphics[width=\linewidth]{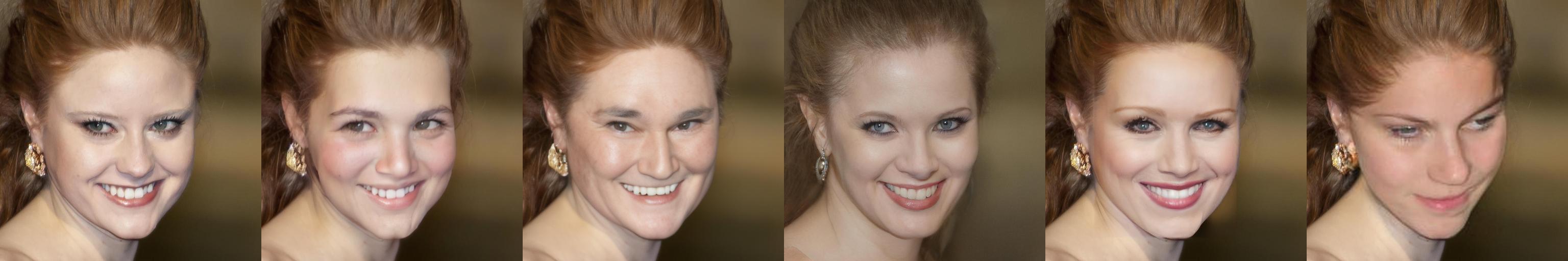}} \\
    \multicolumn{6}{@{}c@{}}{\includegraphics[width=\linewidth]{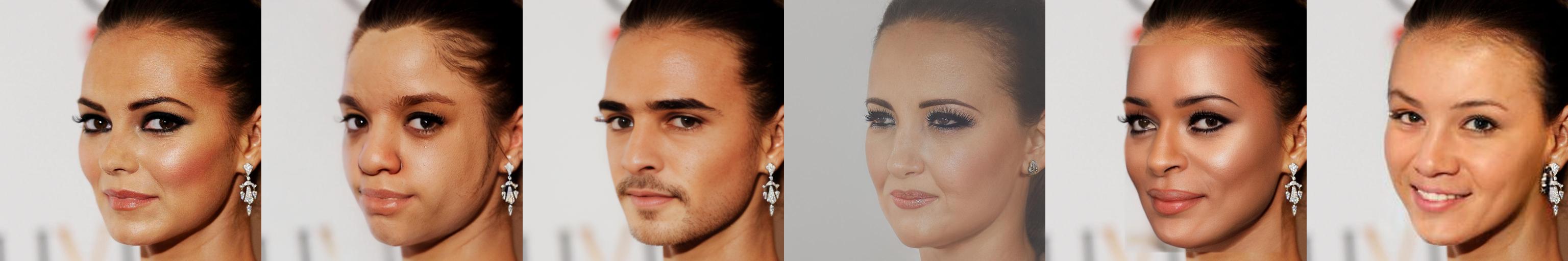}} \\
    \multicolumn{6}{@{}c@{}}{\includegraphics[width=\linewidth]{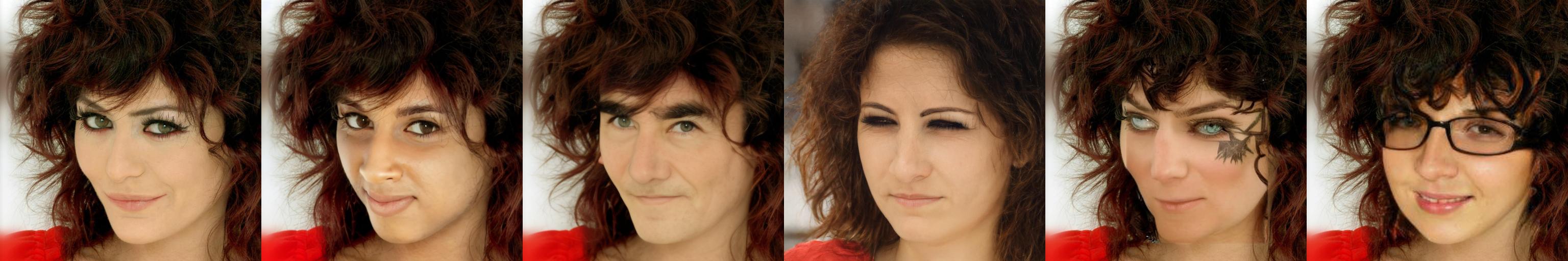}} \\
    \multicolumn{6}{@{}c@{}}{\includegraphics[width=\linewidth]{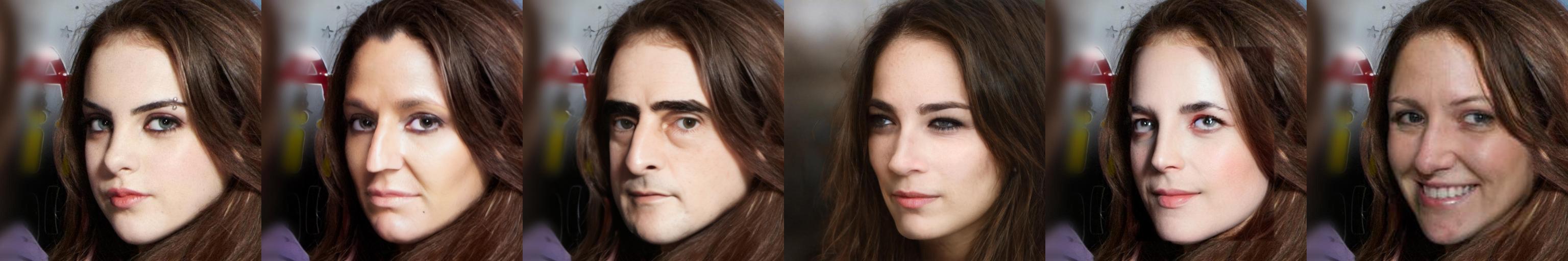}} \\
    \multicolumn{6}{@{}c@{}}{\includegraphics[width=\linewidth]{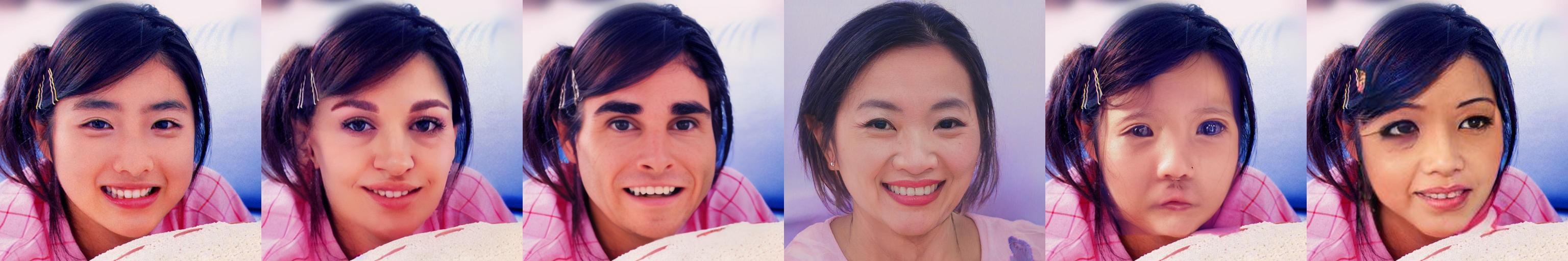}} \\
    \multicolumn{6}{@{}c@{}}{\includegraphics[width=\linewidth]{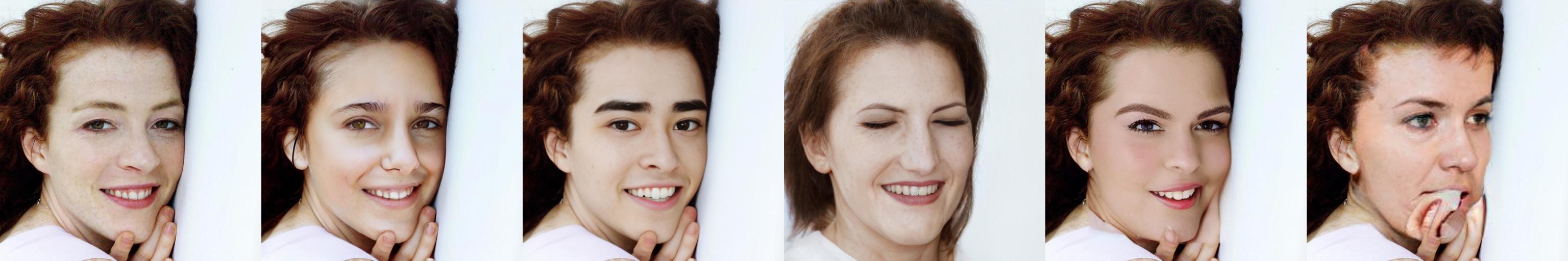}} \\
    \centering Original & \centering Ours & \centering FAMS~\cite{kung2025face} & \centering FALCO~\cite{barattin2023attribute} & \centering LDFA~\cite{klemp2023ldfa} & \centering DP2~\cite{hukkelaas2023deepprivacy2} \\
  \end{tabularx}
  \caption{Qualitative comparison of anonymization results on CelebA-HQ~\cite{karras2017progressive} images.}
  \label{fig:comp_cele_plus_0}
\end{figure*}

\begin{figure*}
  \centering
  \begin{tabularx}{\linewidth}{@{}X@{}X@{}X@{}X@{}X@{}X@{}}
    \multicolumn{6}{@{}c@{}}{\includegraphics[width=\linewidth]{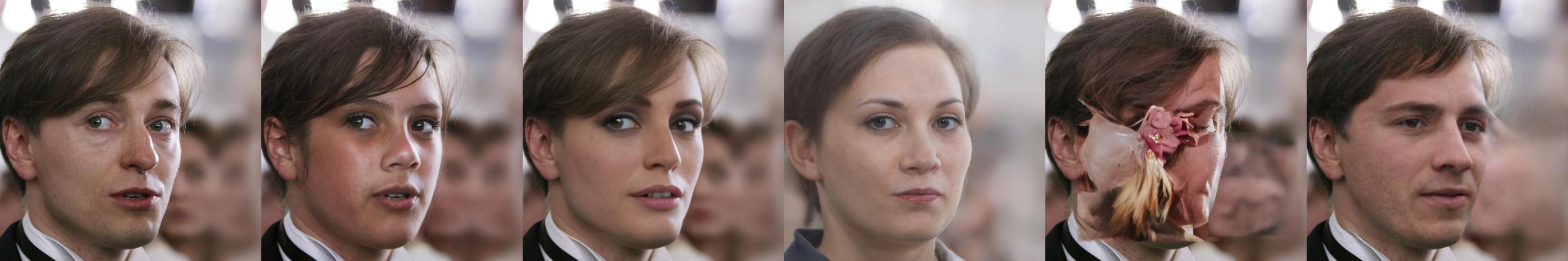}} \\
    \multicolumn{6}{@{}c@{}}{\includegraphics[width=\linewidth]{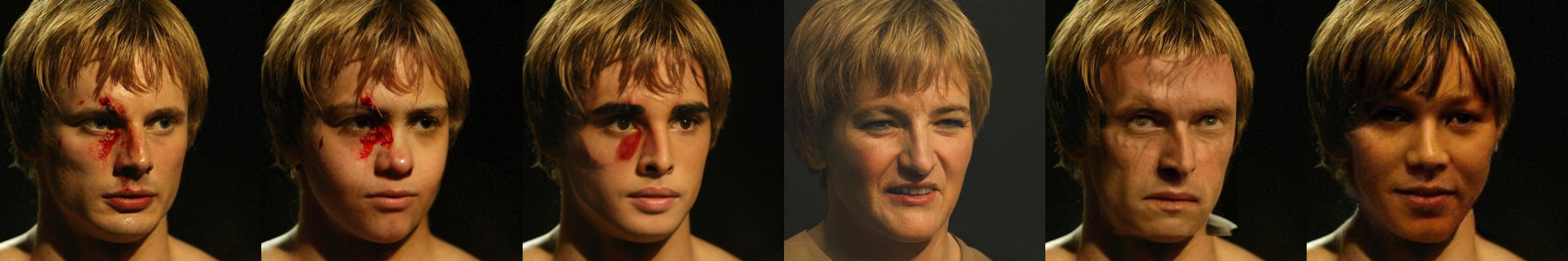}} \\
    \multicolumn{6}{@{}c@{}}{\includegraphics[width=\linewidth]{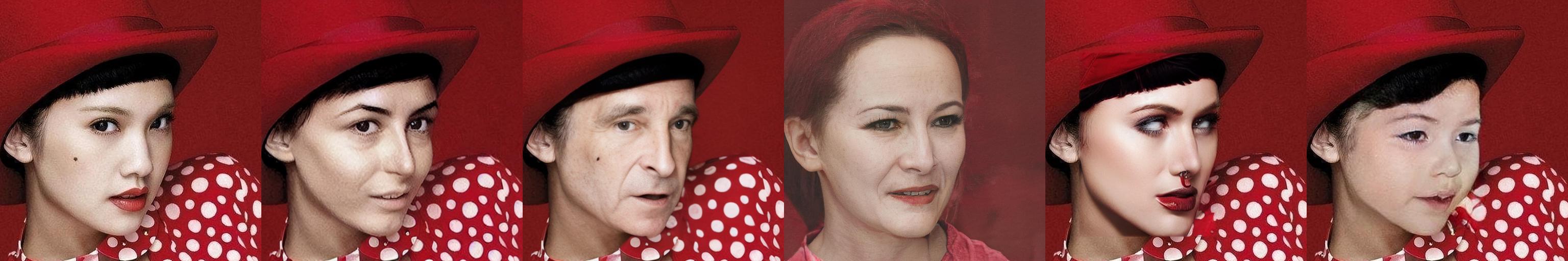}} \\
    \multicolumn{6}{@{}c@{}}{\includegraphics[width=\linewidth]{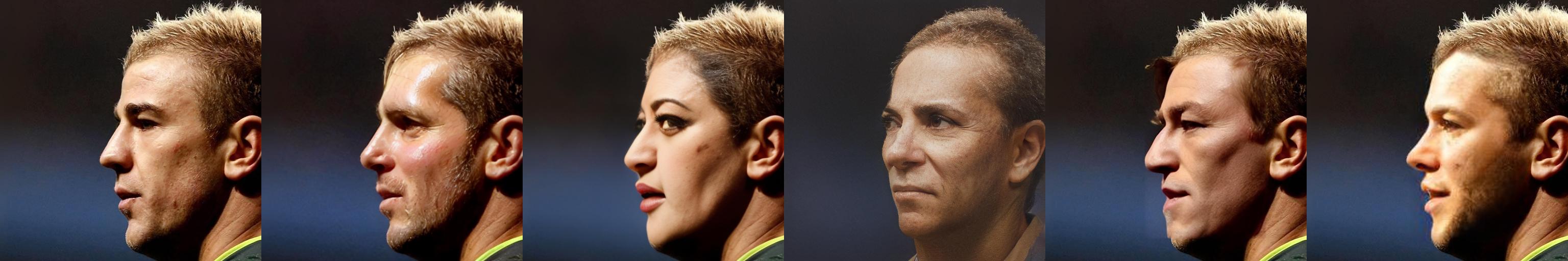}} \\
    \multicolumn{6}{@{}c@{}}{\includegraphics[width=\linewidth]{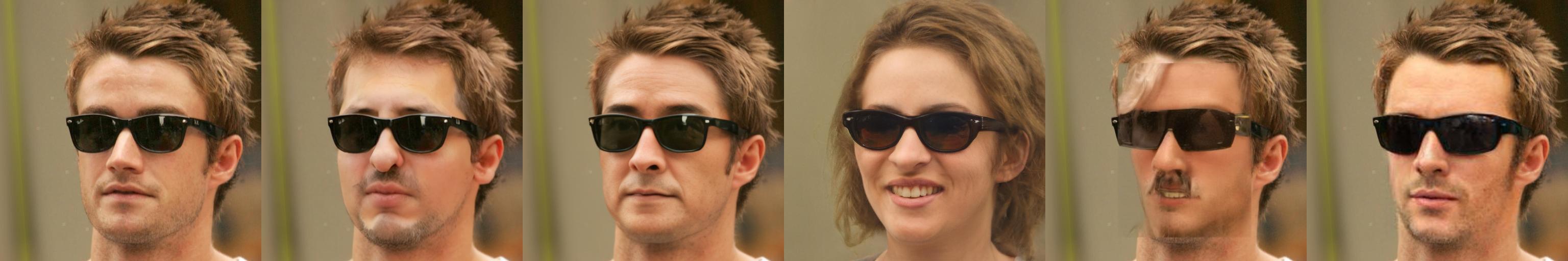}} \\
    \multicolumn{6}{@{}c@{}}{\includegraphics[width=\linewidth]{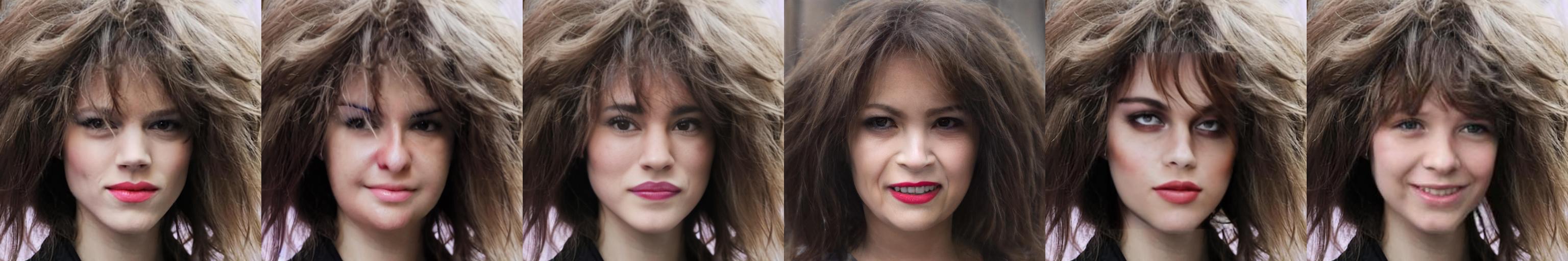}} \\
    \centering Original & \centering Ours & \centering FAMS~\cite{kung2025face} & \centering FALCO~\cite{barattin2023attribute} & \centering LDFA~\cite{klemp2023ldfa} & \centering DP2~\cite{hukkelaas2023deepprivacy2} \\
  \end{tabularx}
  \caption{Qualitative comparison of anonymization results on CelebA-HQ~\cite{karras2017progressive} images.}
  \label{fig:comp_cele_plus_1}
\end{figure*}

\begin{figure*}
  \centering
  \begin{tabularx}{\linewidth}{@{}X@{}X@{}X@{}X@{}X@{}X@{}}
    \multicolumn{6}{@{}c@{}}{\includegraphics[width=\linewidth]{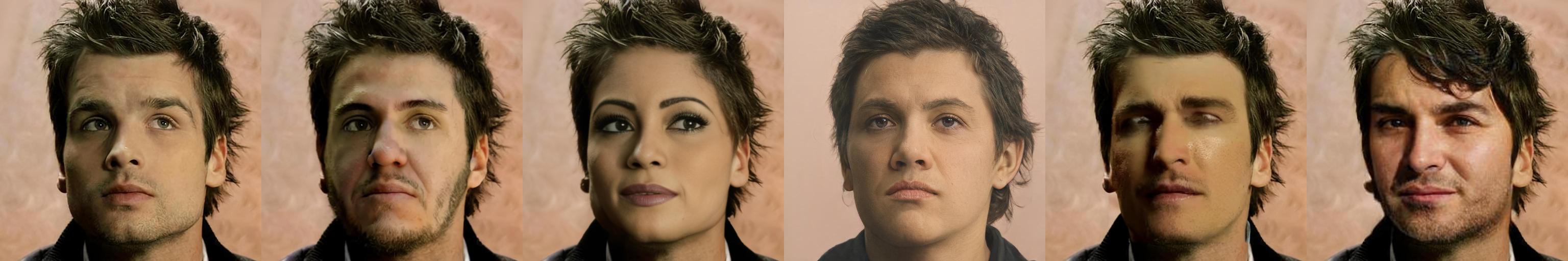}} \\
    \multicolumn{6}{@{}c@{}}{\includegraphics[width=\linewidth]{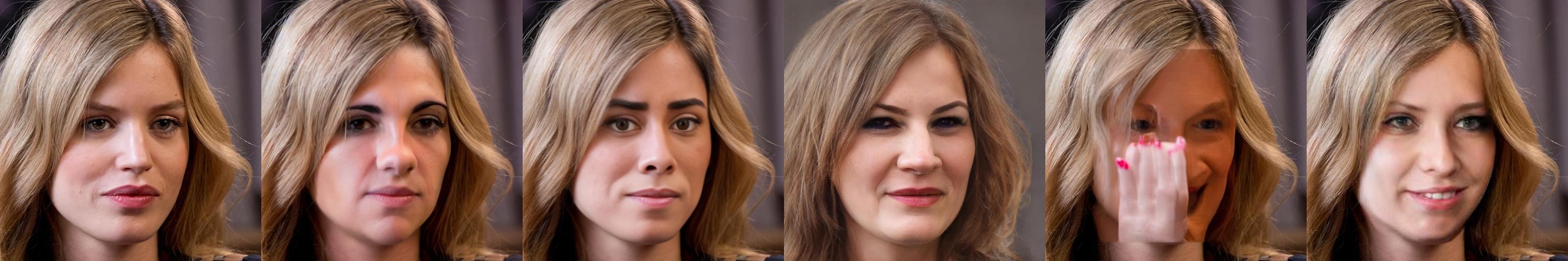}} \\
    \multicolumn{6}{@{}c@{}}{\includegraphics[width=\linewidth]{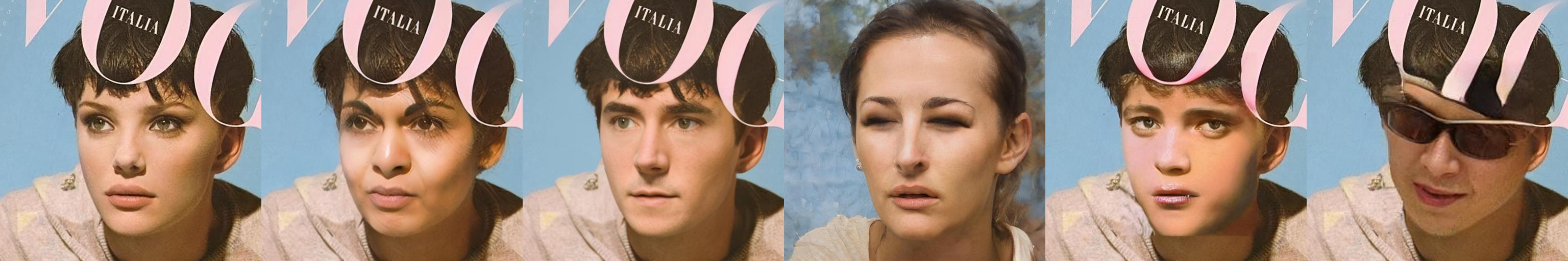}} \\
    \multicolumn{6}{@{}c@{}}{\includegraphics[width=\linewidth]{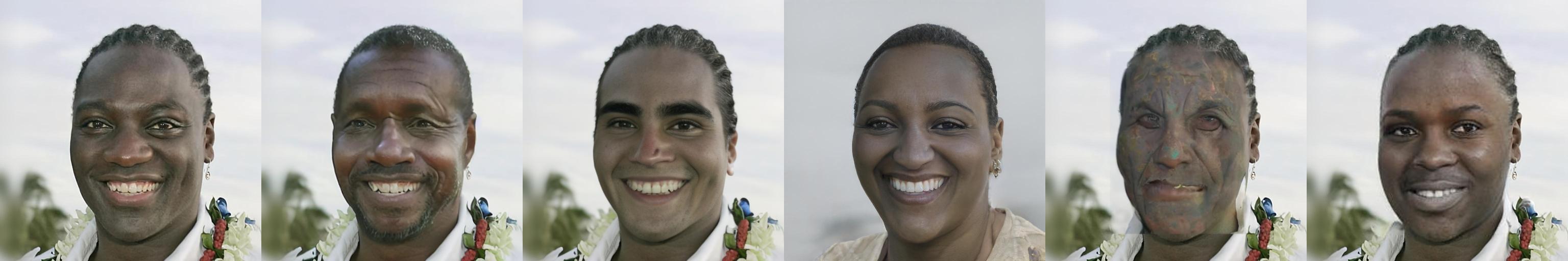}} \\
    \multicolumn{6}{@{}c@{}}{\includegraphics[width=\linewidth]{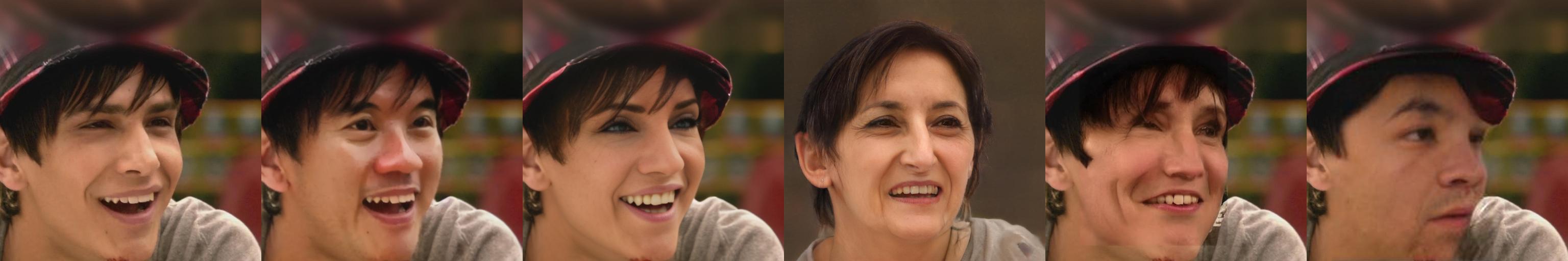}} \\
    \multicolumn{6}{@{}c@{}}{\includegraphics[width=\linewidth]{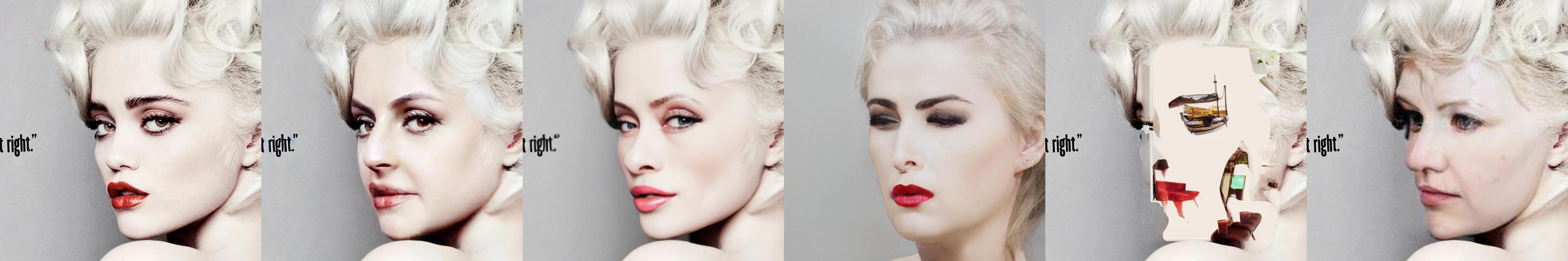}} \\
    \centering Original & \centering Ours & \centering FAMS~\cite{kung2025face} & \centering FALCO~\cite{barattin2023attribute} & \centering LDFA~\cite{klemp2023ldfa} & \centering DP2~\cite{hukkelaas2023deepprivacy2} \\
  \end{tabularx}
  \caption{Qualitative comparison of anonymization results on CelebA-HQ~\cite{karras2017progressive} images.}
  \label{fig:comp_cele_plus_2}
\end{figure*}

\begin{figure*}
  \centering
  \begin{tabularx}{\linewidth}{@{}X@{}X@{}X@{}X@{}X@{}X@{}}
    \multicolumn{6}{@{}c@{}}{\includegraphics[width=\linewidth]{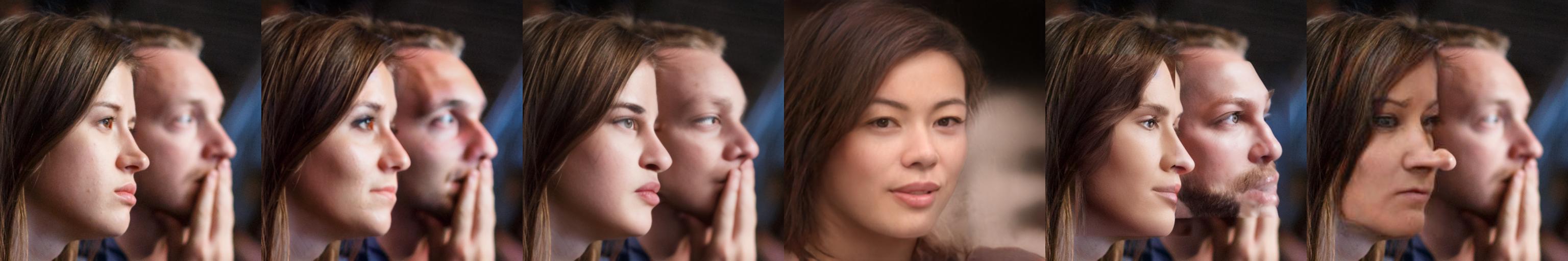}} \\
    \multicolumn{6}{@{}c@{}}{\includegraphics[width=\linewidth]{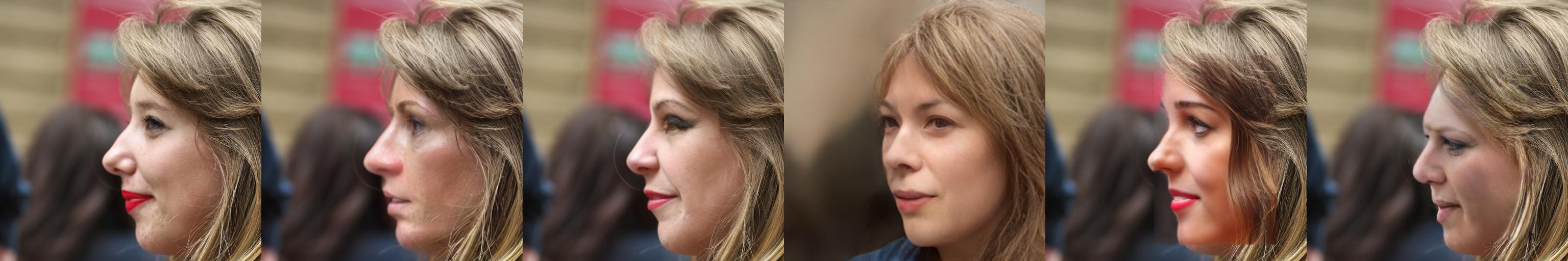}} \\
    \multicolumn{6}{@{}c@{}}{\includegraphics[width=\linewidth]{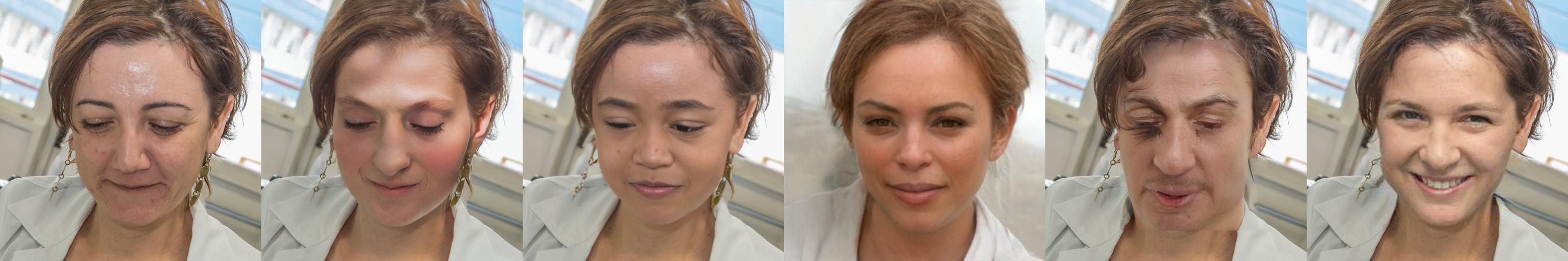}} \\
    \multicolumn{6}{@{}c@{}}{\includegraphics[width=\linewidth]{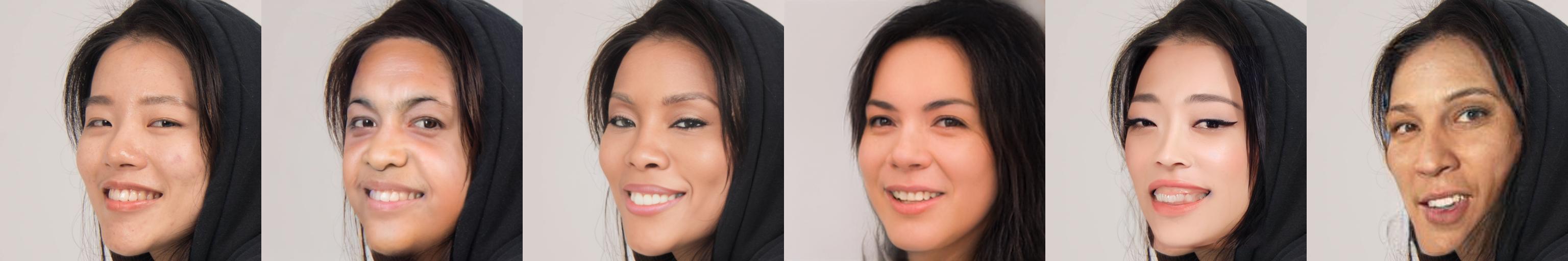}} \\
    \multicolumn{6}{@{}c@{}}{\includegraphics[width=\linewidth]{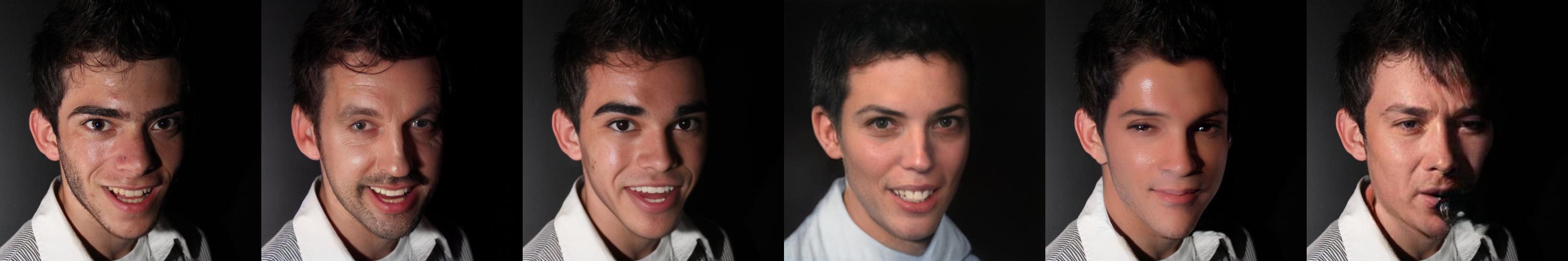}} \\
    \multicolumn{6}{@{}c@{}}{\includegraphics[width=\linewidth]{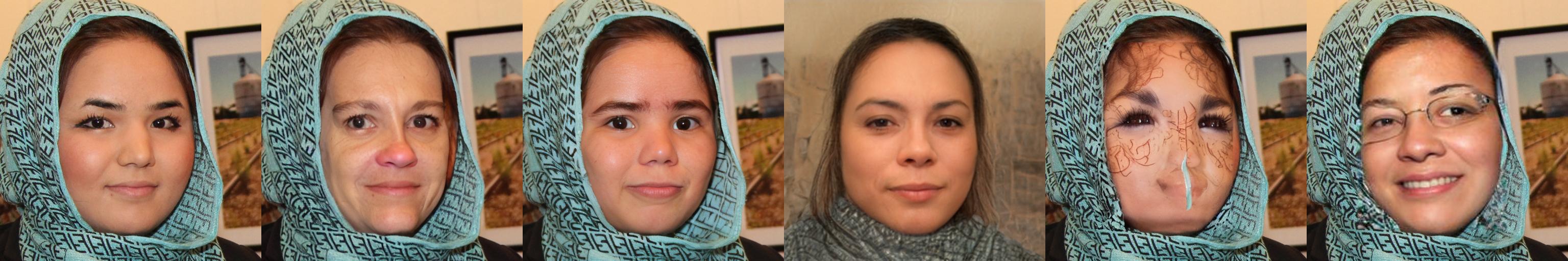}} \\
    \centering Original & \centering Ours & \centering FAMS~\cite{kung2025face} & \centering RiDDLE~\cite{li2023riddle} & \centering LDFA~\cite{klemp2023ldfa} & \centering DP2~\cite{hukkelaas2023deepprivacy2} \\
  \end{tabularx}
  \caption{Qualitative comparison of anonymization results on FFHQ~\cite{karras2019style} images.}
  \label{fig:comp_ffhq_plus_0}
\end{figure*}

\begin{figure*}
  \centering
  \begin{tabularx}{\linewidth}{@{}X@{}X@{}X@{}X@{}X@{}X@{}}
    \multicolumn{6}{@{}c@{}}{\includegraphics[width=\linewidth]{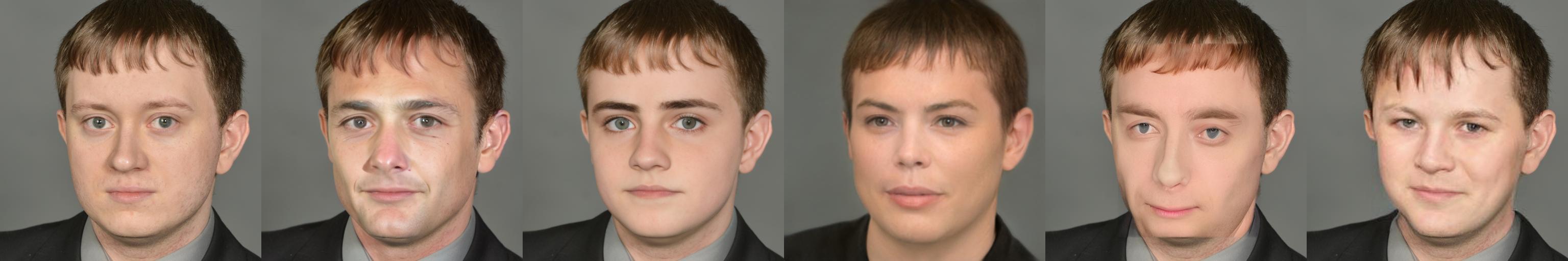}} \\
    \multicolumn{6}{@{}c@{}}{\includegraphics[width=\linewidth]{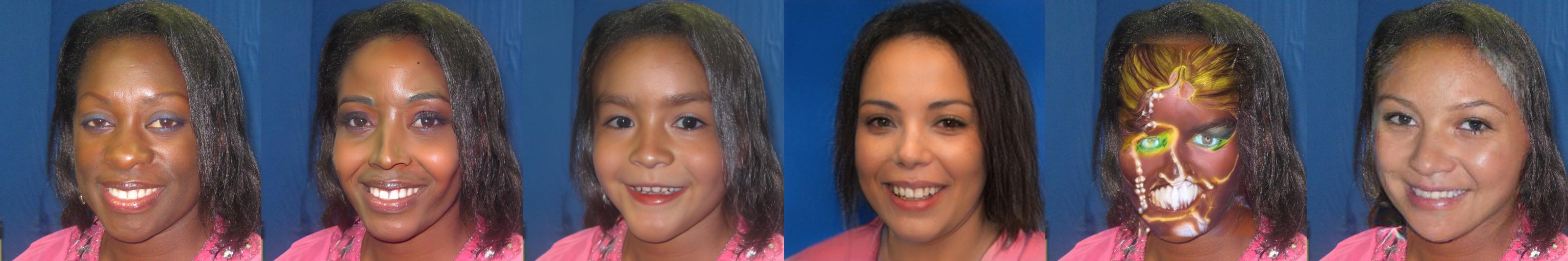}} \\
    \multicolumn{6}{@{}c@{}}{\includegraphics[width=\linewidth]{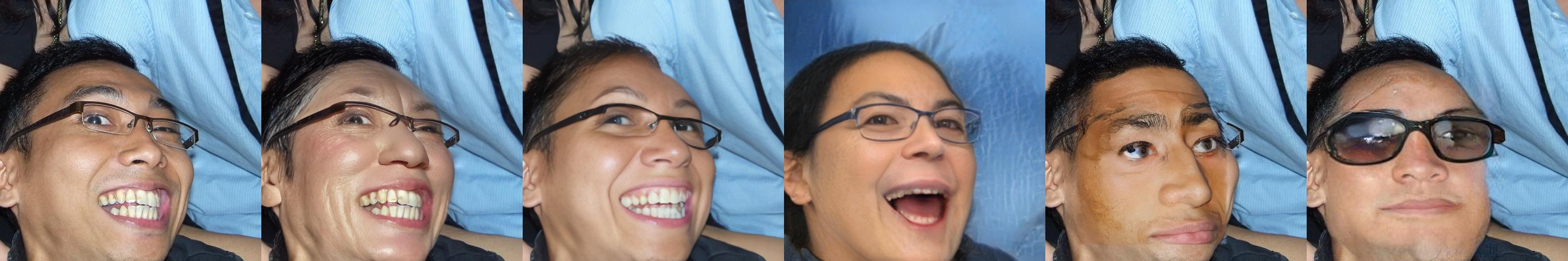}} \\
    \multicolumn{6}{@{}c@{}}{\includegraphics[width=\linewidth]{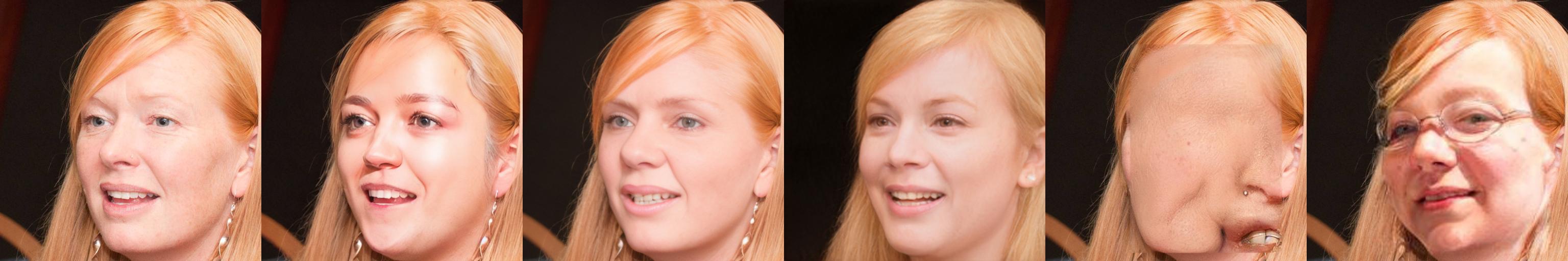}} \\
    \multicolumn{6}{@{}c@{}}{\includegraphics[width=\linewidth]{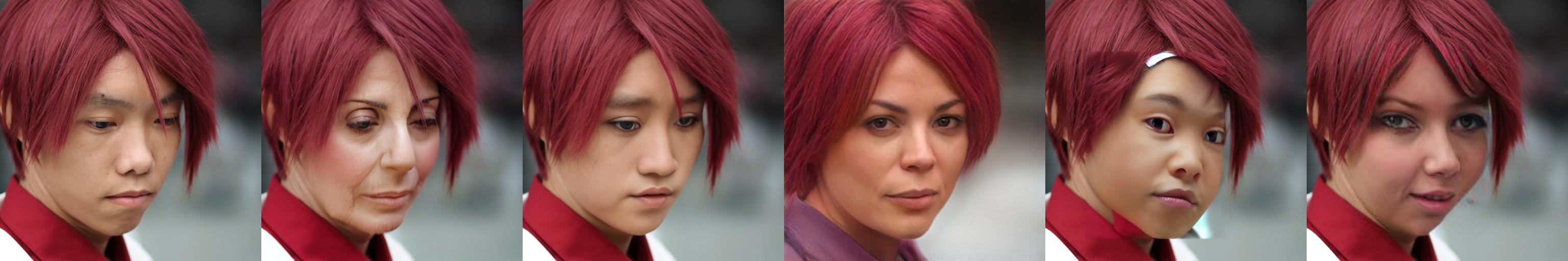}} \\
    \multicolumn{6}{@{}c@{}}{\includegraphics[width=\linewidth]{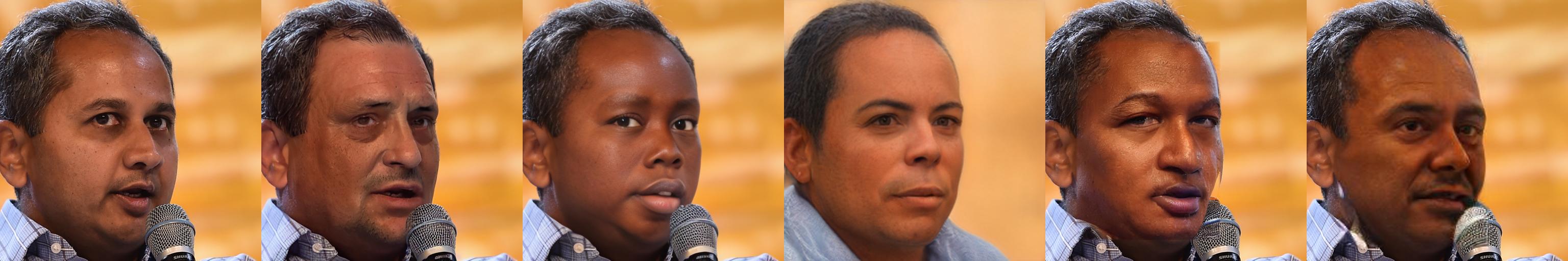}} \\
    \centering Original & \centering Ours & \centering FAMS~\cite{kung2025face} & \centering RiDDLE~\cite{li2023riddle} & \centering LDFA~\cite{klemp2023ldfa} & \centering DP2~\cite{hukkelaas2023deepprivacy2} \\
  \end{tabularx}
  \caption{Qualitative comparison of anonymization results on FFHQ~\cite{karras2019style} images.}
  \label{fig:comp_ffhq_plus_1}
\end{figure*}

\begin{figure*}
  \centering
  \begin{tabularx}{\linewidth}{@{}X@{}X@{}X@{}X@{}X@{}X@{}}
    \multicolumn{6}{@{}c@{}}{\includegraphics[width=\linewidth]{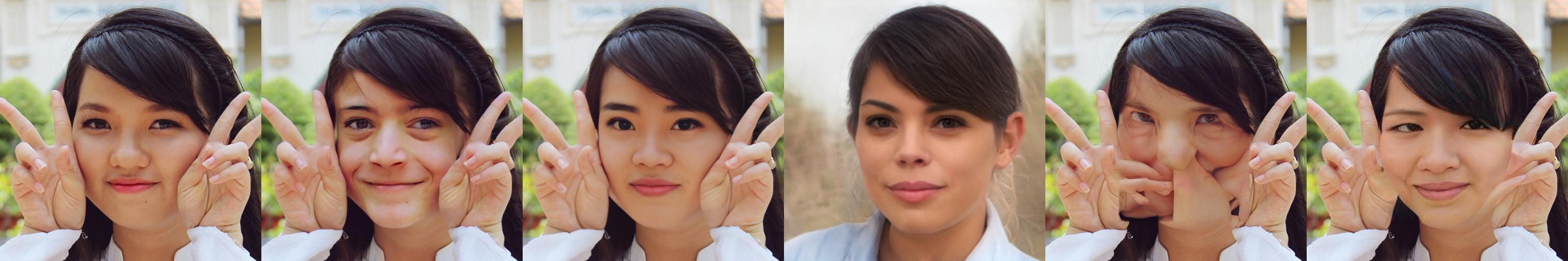}} \\
    \multicolumn{6}{@{}c@{}}{\includegraphics[width=\linewidth]{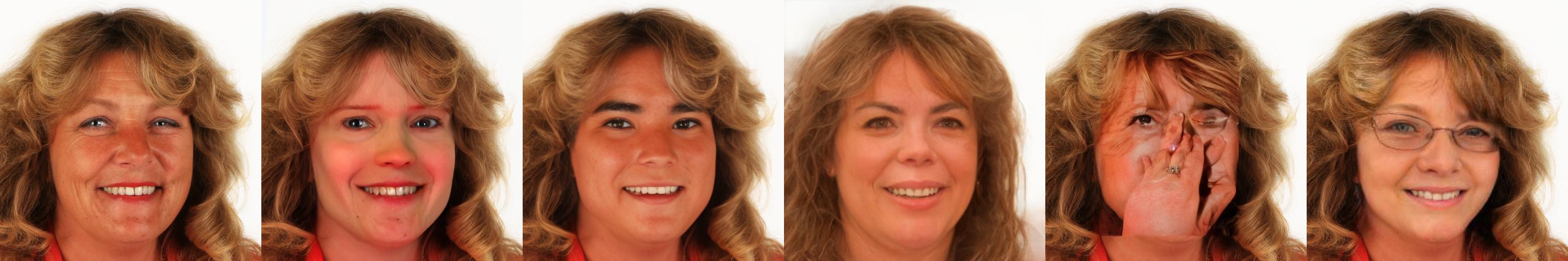}} \\
    \multicolumn{6}{@{}c@{}}{\includegraphics[width=\linewidth]{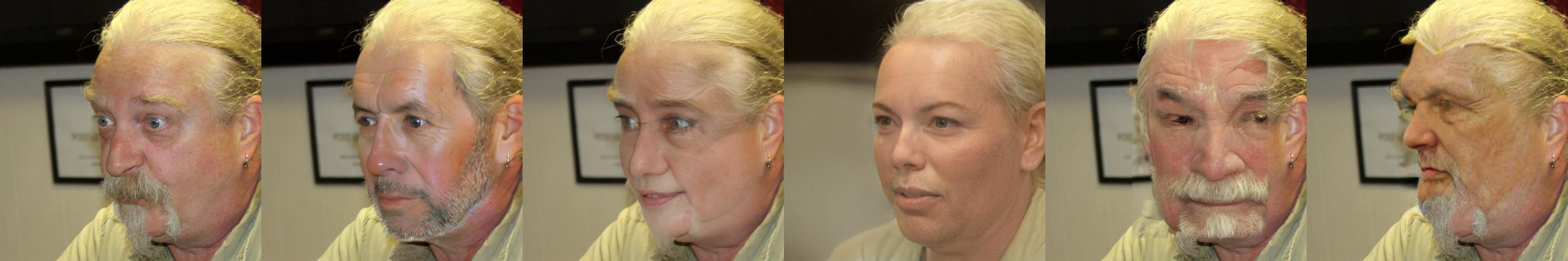}} \\
    \multicolumn{6}{@{}c@{}}{\includegraphics[width=\linewidth]{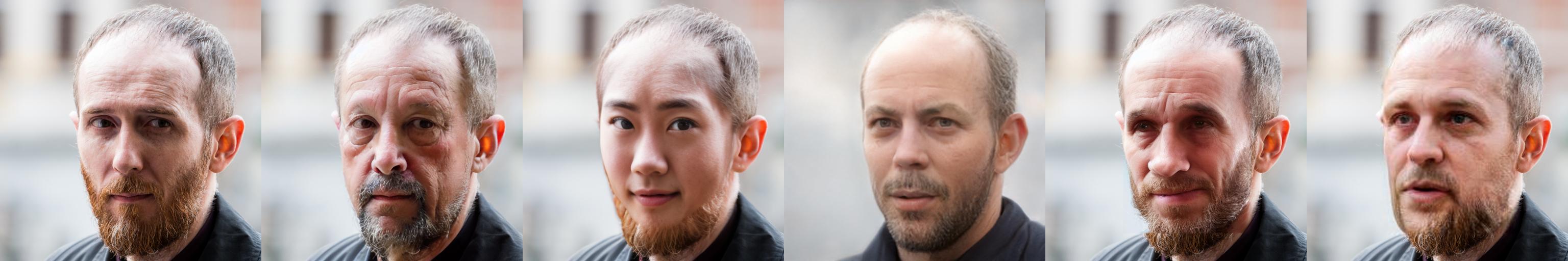}} \\
    \multicolumn{6}{@{}c@{}}{\includegraphics[width=\linewidth]{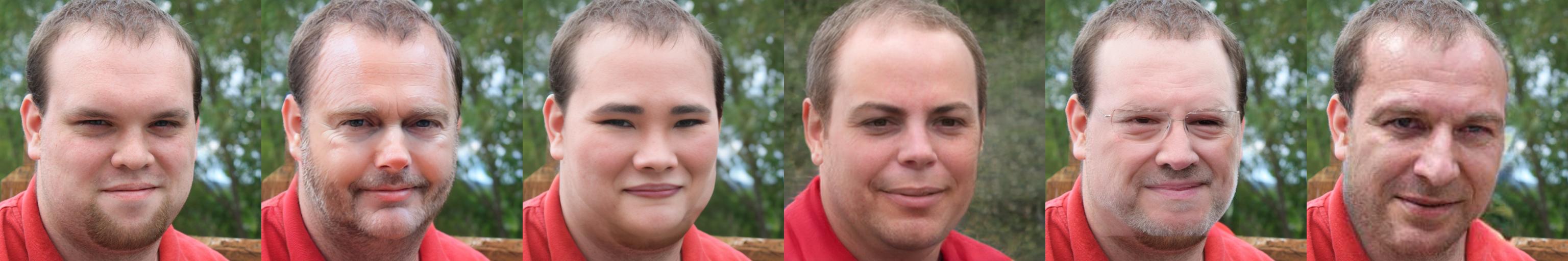}} \\
    \multicolumn{6}{@{}c@{}}{\includegraphics[width=\linewidth]{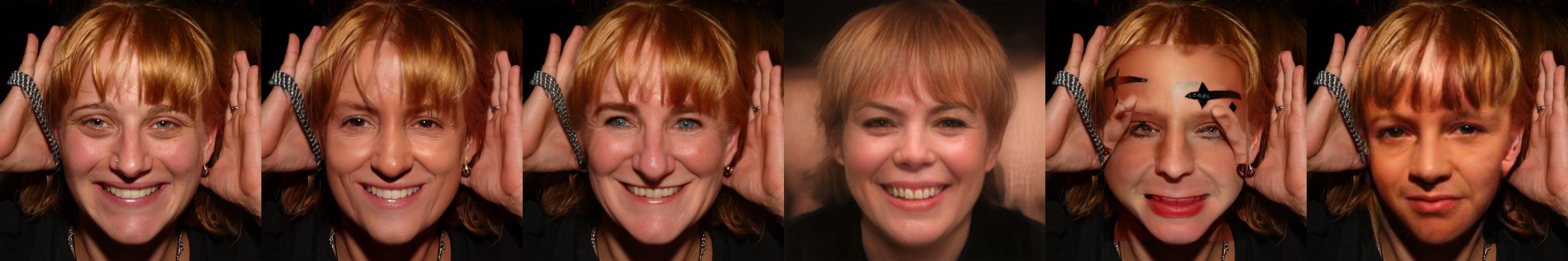}} \\
    \centering Original & \centering Ours & \centering FAMS~\cite{kung2025face} & \centering RiDDLE~\cite{li2023riddle} & \centering LDFA~\cite{klemp2023ldfa} & \centering DP2~\cite{hukkelaas2023deepprivacy2} \\
  \end{tabularx}
  \caption{Qualitative comparison of anonymization results on FFHQ~\cite{karras2019style} images.}
  \label{fig:comp_ffhq_plus_2}
\end{figure*}

We include additional qualitative comparisons in \cref{fig:comp_cele_plus_0,fig:comp_cele_plus_1,fig:comp_cele_plus_2,fig:comp_ffhq_plus_0,fig:comp_ffhq_plus_1,fig:comp_ffhq_plus_2}. These comparisons showcase the performance of our approach against state-of-the-art baselines, including FAMS~\cite{kung2025face}, FALCO~\cite{barattin2023attribute}, RiDDLE~\cite{li2023riddle}, LDFA~\cite{klemp2023ldfa}, and DP2~\cite{hukkelaas2023deepprivacy2}, using images from both the CelebA-HQ~\cite{karras2017progressive} and FFHQ~\cite{karras2019style} datasets.

\section{SOCIETAL IMPACTS}

The rise of AI-generated faces presents societal challenges. On one hand, technologies like facial anonymization offer privacy protections in an increasingly surveillance-driven world. On the other hand, the same capabilities can be weaponized for harmful purposes. For example, synthetic faces can be exploited to fabricate identities for scams, deepfake content, or online impersonation, eroding trust in digital interactions and media authenticity. These risks highlight the need for proactive measures to mitigate potential misuse. Technological solutions, such as deepfake detection tools and digital watermarking systems, can help differentiate real images from synthetic ones. By fostering collaboration across technology developers, we can harness the benefits of AI-generated faces while mitigating their potential to harm social trust and integrity.

